\documentclass[journal]{IEEEtran}
\ifCLASSINFOpdf
\else
\fi
\usepackage{amsmath}
\usepackage{amsthm}
\usepackage{algorithm}
\usepackage{algorithmic}
\usepackage{url}
\usepackage{color}
\usepackage{listings}
\usepackage{multirow}
\usepackage{amssymb}
\usepackage{url}
\usepackage{subfigure}
\usepackage{graphicx}
\newtheorem{thm}{Theorem}[section]

\newtheorem{Property}{Property}[section]

\usepackage{lineno}
\begin{document}
%
\title{
Multiple Flat Projections for Cross-manifold Clustering}
%
%

\author{Lan Bai,~Yuan-Hai~Shao,~Wei-Jie~Chen,~Zhen~Wang,~Nai-Yang~Deng
\IEEEcompsocitemizethanks{\IEEEcompsocthanksitem Lan Bai is with
School of Mathematical Sciences, Inner Mongolia University, Hohhot,
010021, P.R.China e-mail: imubailan@163.com.
\IEEEcompsocthanksitem Yuan-Hai Shao (*Corresponding author) is with School of Management, Hainan University, Haikou,
570228, P.R.China e-mail: shaoyuanhai21@163.com.
\IEEEcompsocthanksitem Wei-Jie Chen is with
Zhijiang College, Zhejiang University of Technology, HangZhou 310014, P.R.China e-mail: wjcper2008@126.com.
\IEEEcompsocthanksitem Zhen Wang is with
School of Mathematical Sciences, Inner Mongolia University, Hohhot,
010021, P.R.China e-mail: wangzhen@imu.edu.cn.
\IEEEcompsocthanksitem Nai-Yang Deng is
with College of Science, China Agriculture University, Beijing, 100083, P.R.China e-mail: dengnaiyang@cau.edu.cn.
}}

%
%

\markboth{IEEE TRANSACTIONS ON CYBERNETIC,~VOL.~X, NO.~X,XXXXXX}%
{Shell \MakeLowercase{\textit{et al.}}: Bare Demo of IEEEtran.cls
for Computer Society Journals}

%



\IEEEcompsoctitleabstractindextext{
\begin{abstract}
Cross-manifold clustering is a hard topic and many traditional clustering methods fail because of the cross-manifold structures. In this paper, we propose a Multiple Flat Projections Clustering (MFPC) to deal with cross-manifold clustering problems. In our MFPC, the given samples are projected into multiple subspaces to discover the global structures of the implicit manifolds. Thus, the cross-manifold clusters are distinguished from the various projections. Further, our MFPC is extended to nonlinear manifold clustering via kernel tricks to deal with more complex cross-manifold clustering. A series of non-convex matrix optimization problems in MFPC are solved by a proposed recursive algorithm. The synthetic tests show that our MFPC works on the cross-manifold structures well. Moreover, experimental results on the benchmark datasets show the excellent performance of our MFPC compared with some state-of-the-art clustering methods.
\end{abstract}

}

%

\maketitle

\IEEEdisplaynotcompsoctitleabstractindextext

\IEEEpeerreviewmaketitle

\section{Introduction}
\IEEEPARstart{C}{lustering} is the process of grouping data samples into clusters \cite{ClusterBook4,ClusterBook5}, with similarity of within-cluster and dissimilarity of between-cluster. It has been applied in many real world applications, e.g., image processing \cite{TIPH,DabovImage}, object tracking \cite{Yi2013Online,Hongwei2018Manifold} and object detection \cite{ClusterA1,ClusterA2}. A large number of studies \cite{Kplane,Kflat,SSC,Pengfei2019Dual,Canyi2019Subspace} have shown that the meaningful structures of data possibly reside on several low-dimensional manifolds. Based on this observation, the objective of clustering is convert to cluster the samples from the implicit low-dimensional manifolds, called manifold clustering \cite{MS,Nengwen2017Robust}. Manifold clustering has been applied in many applications, e.g., manifold learning, \cite{ISOMAP,LLE,Belkin2006Manifold}, interpretation of video \cite{LaveeUnderstanding}, motion capture \cite{Moeslund2006A} and hand writing recognition \cite{PLAMONDON2000On}.

\begin{figure*}[htbp]
\centering
   \subfigure[Data]{\includegraphics[width=0.14\textheight]{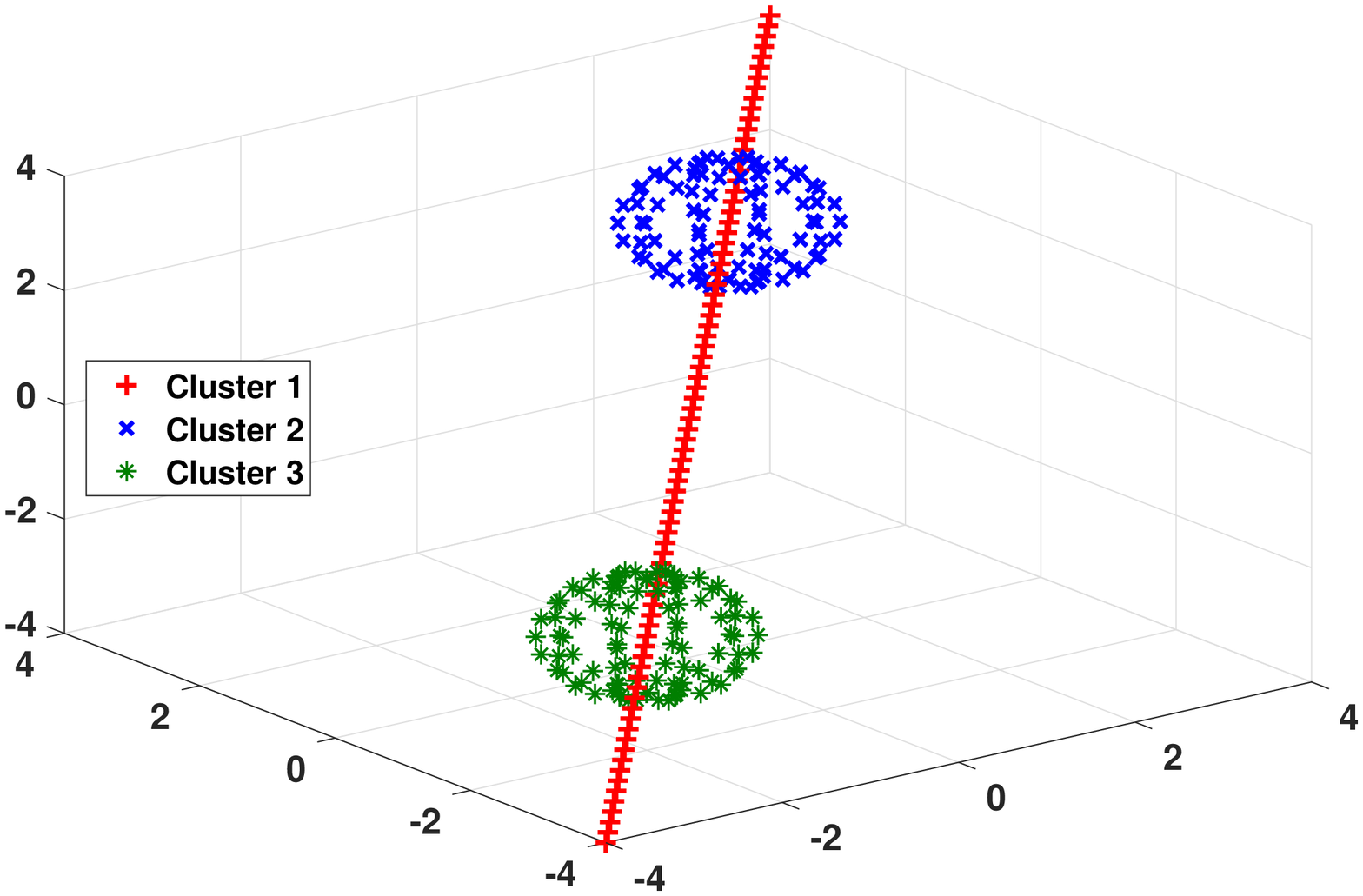}}
   \subfigure[SMMC]{\includegraphics[width=0.14\textheight]{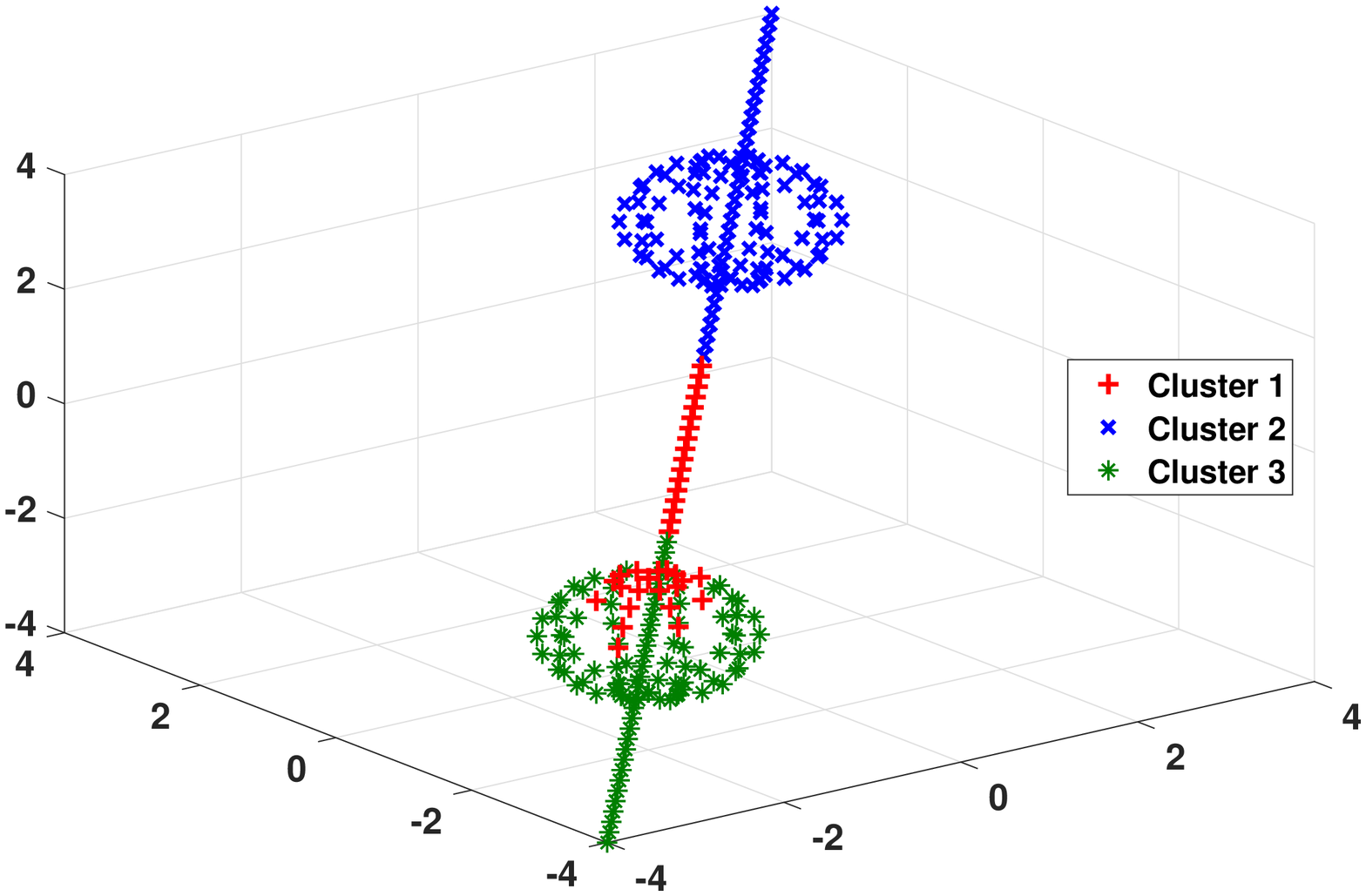}}
   \subfigure[$k$means]{\includegraphics[width=0.14\textheight]{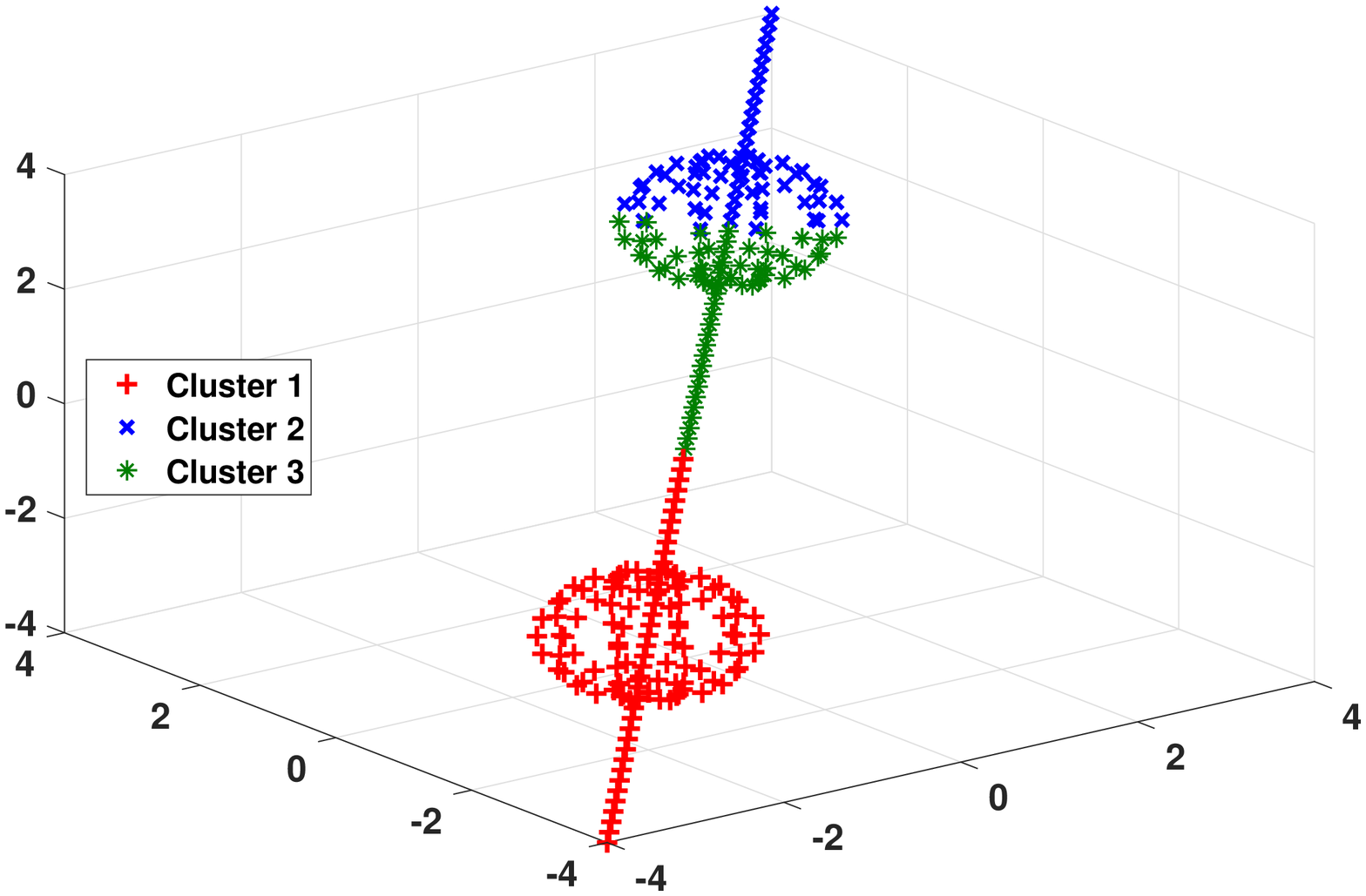}}
   \subfigure[$k$PC]{\includegraphics[width=0.14\textheight]{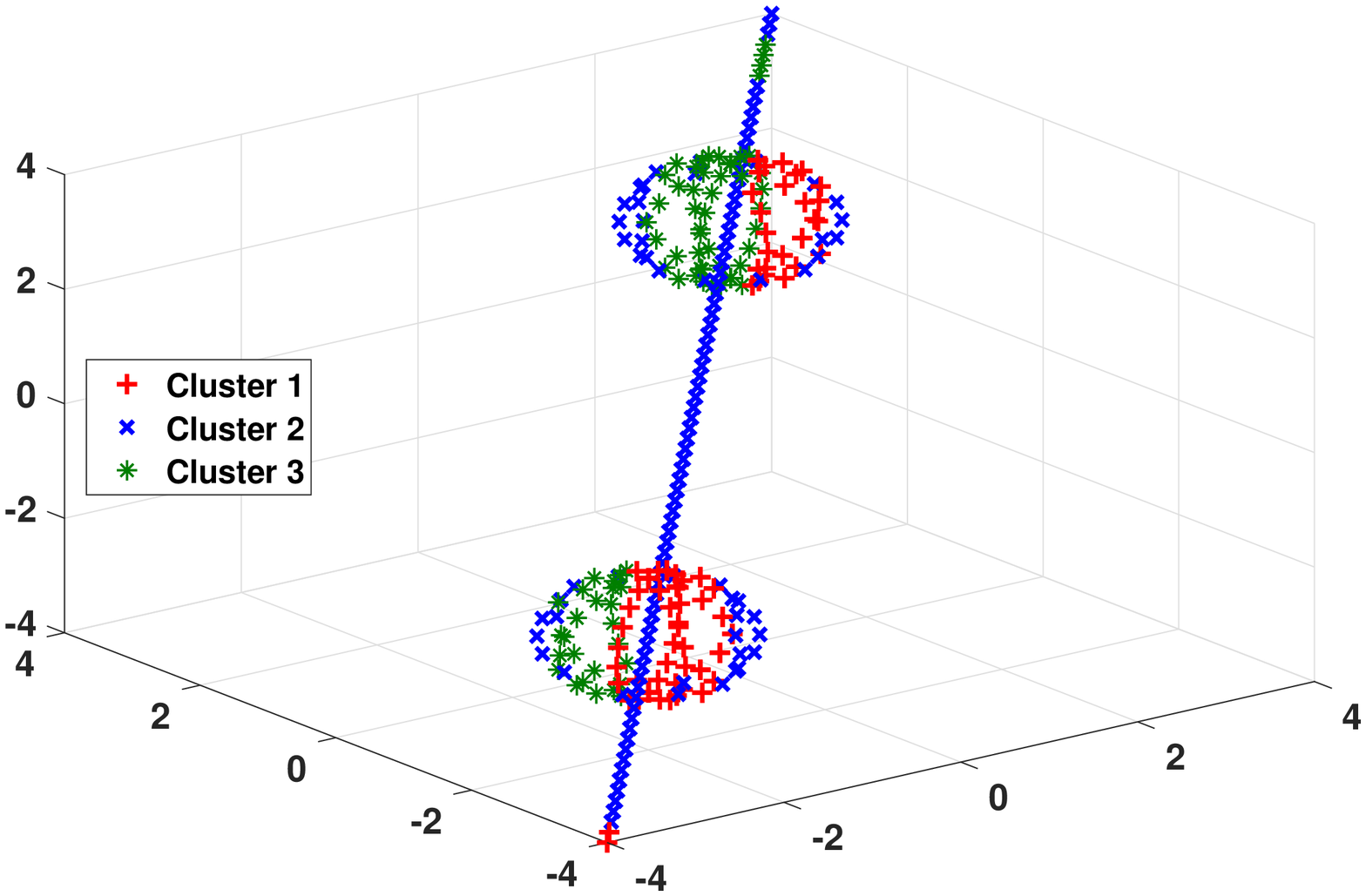}}
   \subfigure[$k$PPC]{\includegraphics[width=0.14\textheight]{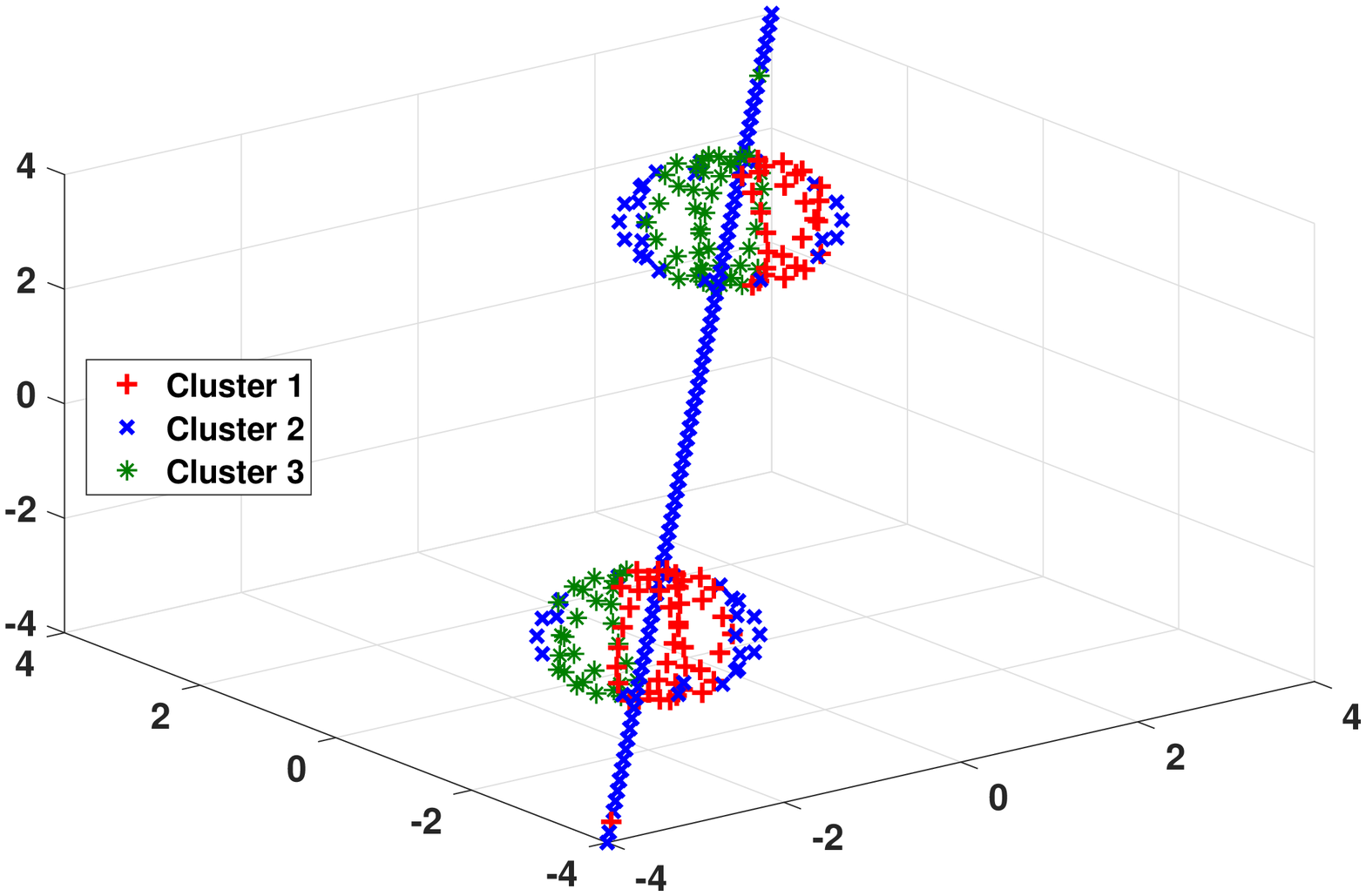}}
   \subfigure[L$k$PPC]{\includegraphics[width=0.14\textheight]{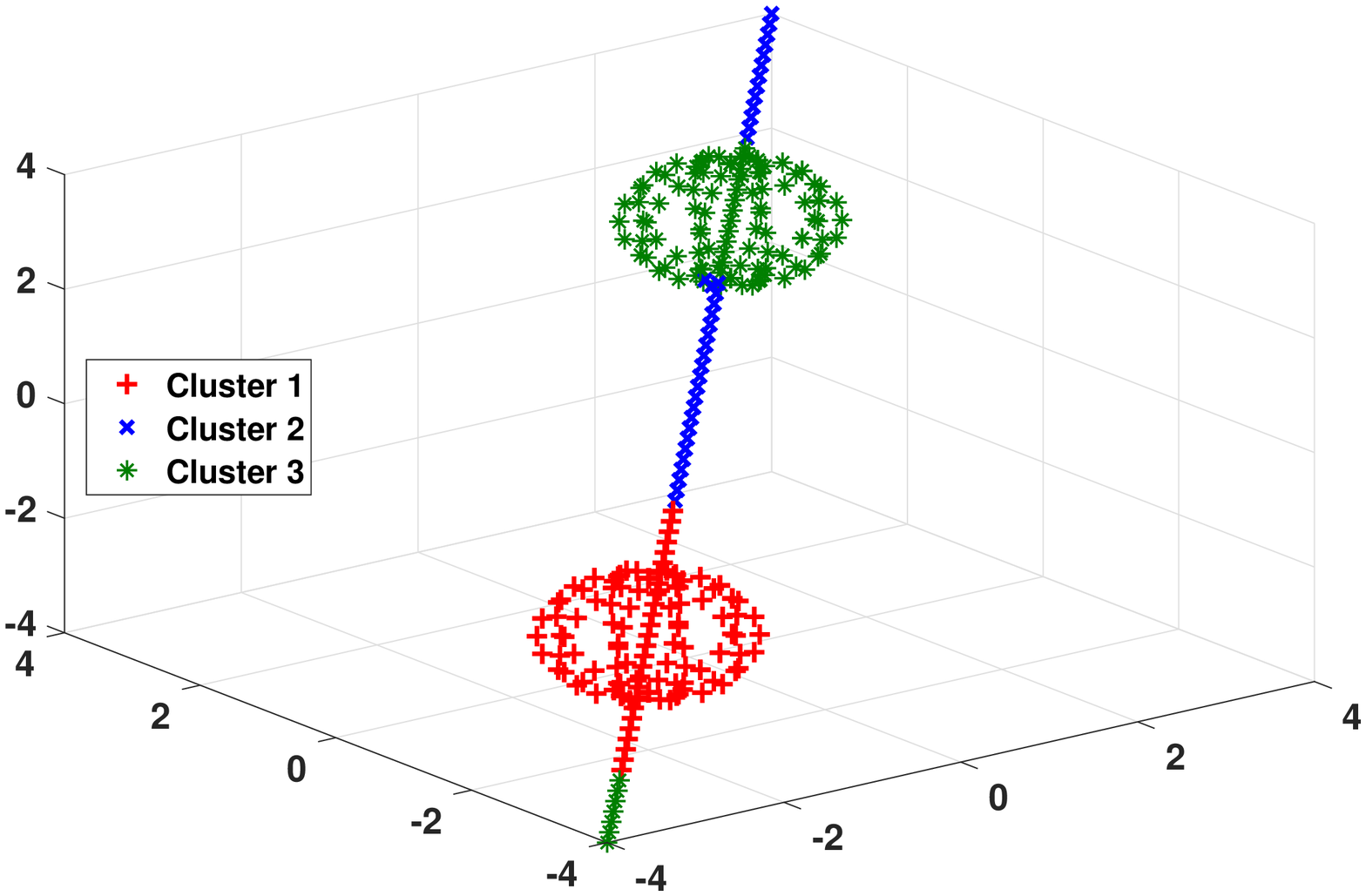}}
   \subfigure[TWSVC]{\includegraphics[width=0.14\textheight]{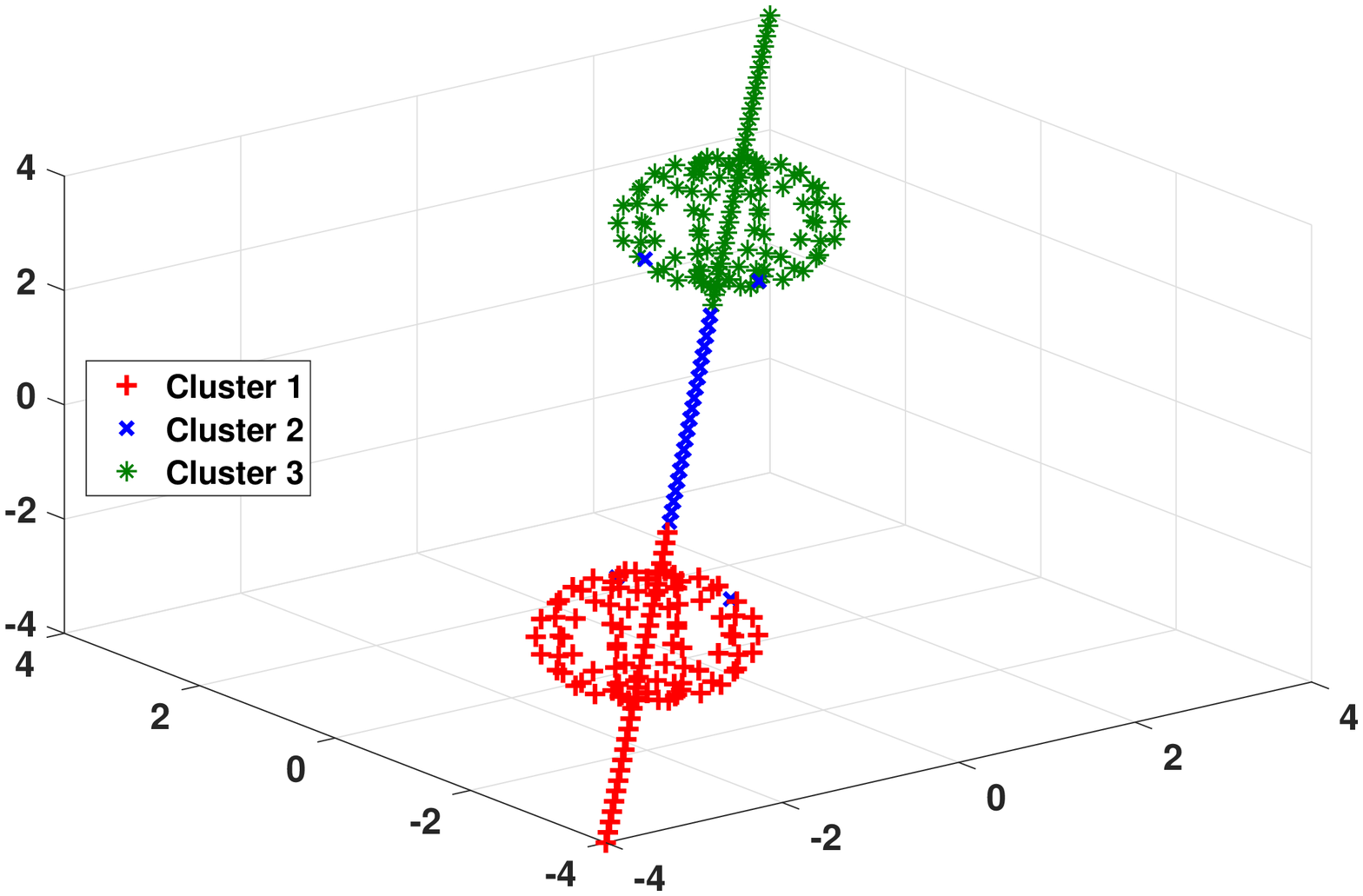}}
   \subfigure[$k$FC]{\includegraphics[width=0.14\textheight]{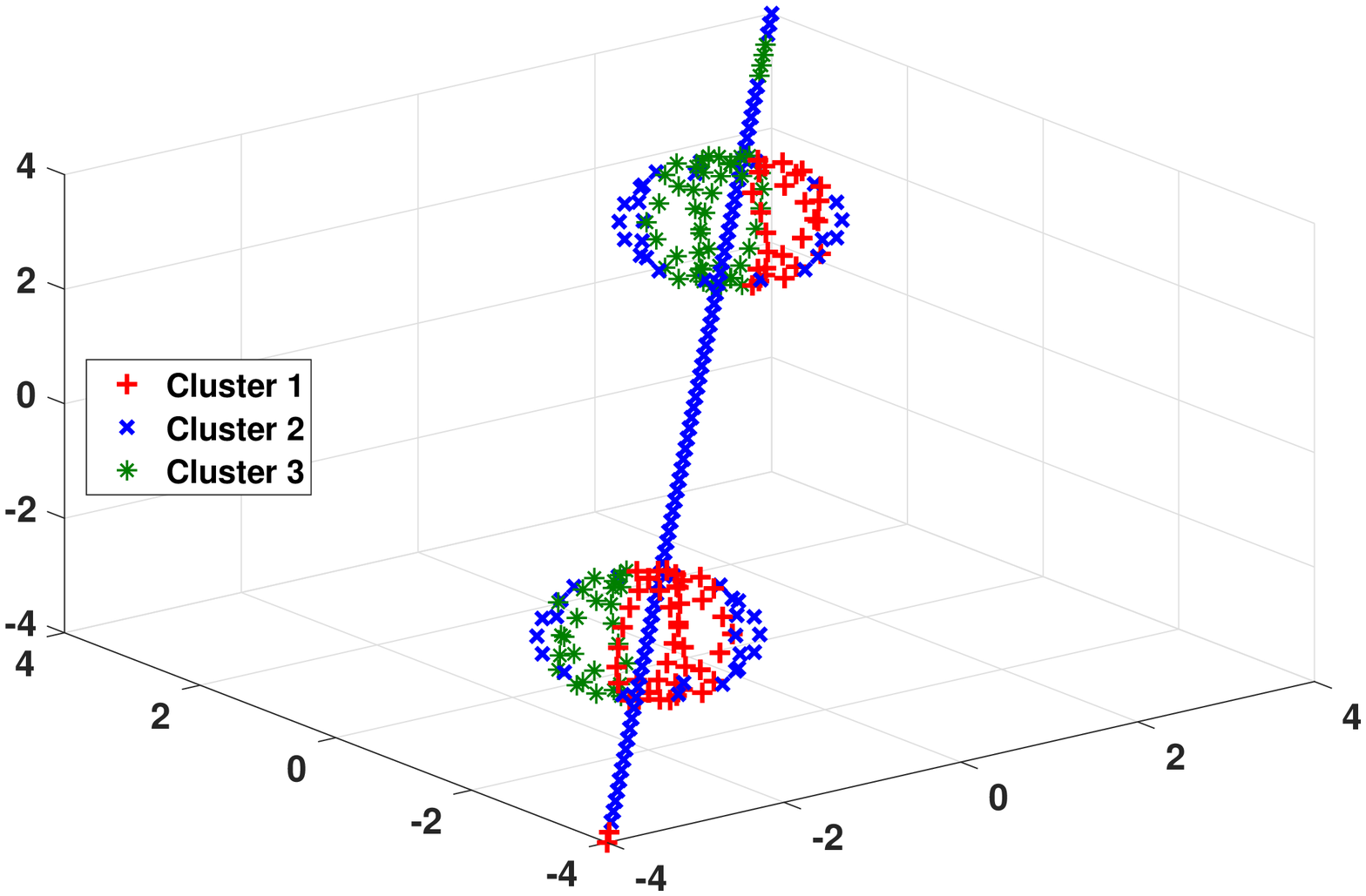}}
   \subfigure[L$k$FC]{\includegraphics[width=0.14\textheight]{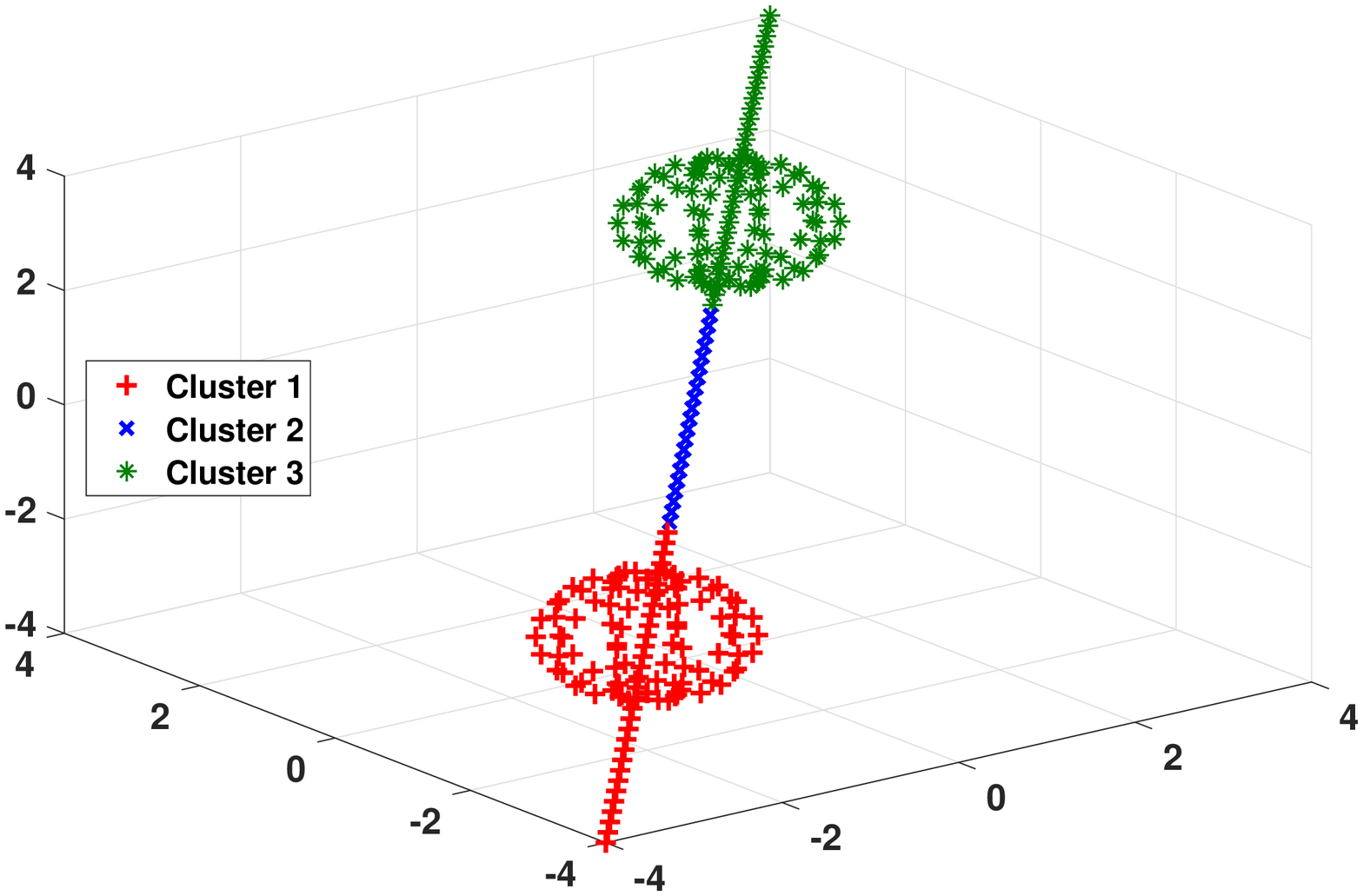}}
   \subfigure[MFPC]{\includegraphics[width=0.14\textheight]{TangMSPC.eps}}
\caption{Clustering results of some state-of-the-art methods on a toy example with cross-manifold structures, where the two spheres intersect with the line in $\mathbb{R}^3$.}\label{Tanghulu1}
\end{figure*}

For manifold clustering, the data generally includes well-separated and cross structures \cite{MMC}. The former are easy to recognize due to its independence, but not for the latter. On the one hand, the attribution of the samples near the intersection of cross manifolds are ambiguous.
On the other hand, the cross structure severs the connection of the samples on the same manifold, results in different clusters from this manifold. Fig. \ref{Tanghulu1}(a) is a toy example which has one class on a line and the other two classes on two spheres, respectively. It looks like the candied haws on a stick. The samples on the line may be misclassified into other clusters, because their links are severed by the spheres.

At present, the above cross-manifold clustering is still a hard topic \cite{MMCA}, though there have been two types of manifold clustering methods: spectral-type clustering \cite{ATSC,OSC,Li2018Fast,Panda2018Nystr} and flat-type clustering \cite{Kplane,Kflat}. Spectral-type clustering assigns the samples into clusters by the similarity graph, which is the local neighborhood relationship. Several spectral-type methods tried to propose a delicate similarity graph to handle the cross-manifold structure, e.g, Spectral Clustering on Multiple Manifolds (SMMC) \cite{SCMM} and Local and Structural Consistency for Multi-Manifold Clustering (LSC) \cite{MSMMC}. However, these methods have difficulties in dealing with the samples near intersections, because the neighborhood of a sample can contain samples from different manifolds and the similarity graph often is fragile. In these methods, some subtle techniques were used to distinguish the different manifolds from the intersections \cite{SSC}. In contrast, flat-type clustering \cite{Kplane} assigns the samples into clusters from global perspective of view. To determine the formation of linear manifolds, Mangasarian et al. proposed $k$-Plane Clustering ($k$PC) \cite{Kplane} by hiring planes/hyperplanes to represent the samples from different manifolds. Subsequently, to find appropriate planes/hyperplanes, many other flat-type clustering methods were proposed based on $k$PC, e.g., $k$-Proximal Planes Clustering ($k$PPC) \cite{kPPC} and Twin Support Vector Machine for Clustering (TWSVC) \cite{TWSVC} with discriminative information, Local $k$-Proximal Plane Clustering (L$k$PPC) \cite{LKPPC} with localization techniques to avoid the infinite extension of the linear models, L1-TWSVC \cite{L1TWSVM} and Twin Bound Vector Machine for Clustering (TBSVC) \cite{TBSVC} to deal with noises. However, the planes/hyperplanes used in the above methods cannot deal with complicated flats apparently \cite{Kflat}. Thus, the unitary planes/hyperplanes were extended to the general flats to suit for more complicated manifolds, e.g., $k$-Flats Clustering ($k$FC) \cite{Kflat} and Local $k$-Flats Clustering (L$k$FC) \cite{Lkflat}.
Many linear manifolds, e.g., lines, planes/hyperplanes and flats, were recognized by the corresponding flat-type methods. However, the flats were obtained without discriminative information in these methods, and thus they cannot recognize the implicit manifolds from cross-manifold structures well. As the toy example, three clusters in $\mathbb{R}^3$ are given in Fig. \ref{Tanghulu1}(a) by different colors. More precisely, the red samples lie on a straight line, both blue and green ones lie respectively on two spheres. Fig. \ref{Tanghulu1}(c)-(h) show the clusters obtained by some state-of-the-art flat-type clustering methods. The results obviously are not satisfactory and reveal their shortcomings. In seeking a flat for an implicit manifold, merely keeping the current samples close to the flat is insufficient, because other cluster samples (especially near the intersections) may close to this flat too. Hence, the discriminative information should be employed. Additionally, the normalization for the flats should be considered at the same time.

In this paper, we proposed a novel flat-type method named Multiple Flat Projections Clustering (MFPC) for cross-manifold problems. For the $k$ implicit flats, our MFPC seeks $k$ corresponding projection subspaces such that the samples projected into each subspace are partially close to the subspace center and the rest are far away from it. When our MFPC considers the global manifold structures in the projection subspaces, the cross-manifold structures would be distinguished by different subspaces, avoiding their local analysis. Fig. \ref{Tanghulu1}(j) is the clustering result of our MFPC, which is the same as the real data obviously. Furthermore, our MFPC is extended for more complicated manifolds via kernel tricks.



The contributions of this paper includes:

(i) A flat-type clustering method is proposed with strong adaptability to cross-manifold structures;

(ii) For each projection subspace, all the samples are projected into a unit sphere to unify the normalization for the subspaces in some sense.

(iii) The non-convex matrix optimization problems in our MFPC are decomposed into several non-convex vector optimization problems by a recursive algorithm, and the latter problems are solved by a proposed iterative algorithm of which the convergence is also given;

(iv) Experiments on some synthetic and benchmark datasets show the amazing performance of our MFPC compared with some state-of-the art clustering methods.

The rest of this paper is organized as follows.
In section 2, some related works, including $k$PC, $k$PPC, L$k$PPC, $k$FC and L$k$FC are reviewed. Section 3 elaborates our MFPC as well as its solution. Experiments are arranged in Section 4, and conclusions are given in Section 5. The appendix gives the proofs of the relevant theorems in this paper.

%
%

\section{background}

Given $m$ samples $X=(x_1,x_2,\ldots,x_m)\in \mathbb{R}^{n\times m}$, consider to cluster the $m$ samples into $k$ clusters with their corresponding labels $Y=(y_1,y_2,\ldots,y_m)$ from 1 to $k$.
Let $N=\{1,\ldots,m\}$ to represent the index set of $X$. $N_i$ and $N\backslash N_i$ represent the index sets of sample belongs to the $i$-th ($i=1,\ldots,k$) cluster and the rest, respectively. $m_i$ denotes the number of the elements in the $i$-th cluster. Thus, $\bar{x}_i=\frac{1}{m_i}\underset{j\in N_i}\sum x_j$ is the mean of the $i$-th cluster. The $L_2$ norm and Frobenius norm are respectively denoted by $||\cdot||$ and $||\cdot||_F$, $|\cdot|$ denotes the absolute value, and $e$ denotes a vector of ones with an appropriate dimension. Let us remind some related works on clustering.

\subsection{$k$PC}
$k$PC \cite{Kplane} wishes to cluster the given samples into $k$ clusters such that the cluster samples are respectively close to the $k$ cluster center planes, which are defined as
\begin{equation}\label{kPCPlane}
\begin{array}{l}
w_i^\top x+b_i=0,~i=1,\ldots,k,
\end{array}
\end{equation}
where $w_i\in \mathbb{R}^n$ and $b_i\in \mathbb{R}$. The required $k$ cluster centers are obtained iteratively. Start from an stochastic initialization $(w_i, b_i)$ with $i=1,\ldots, k$, then the labels are updated by
\begin{equation}\label{PredictkPC}
\begin{array}{l}
y=\underset{i=1,\ldots,k}{\arg\min}~|w_i^\top x+b_i|.
\end{array}
\end{equation}
The cluster center planes are updated by solving the following problem with $i=1,\ldots,k$,
\begin{equation}\label{kPC}
\begin{array}{l}
\underset{w_i,b_i}{\min}~~
\sum\limits_{j\in N_i}(w_i^{\top}x_j+b_i)^2\\
\hbox{s.t.\ }~~~||w_i||^2=1,
\end{array}
\end{equation}
which is equivalent to an eigenvalue problem. The $k$ cluster center planes \eqref{kPCPlane} and the samples' labels are updated alternately until there is a repeated overall assignment of samples to clusters or a non-decrease in the overall objective.

\subsection{$k$PPC}
$k$PPC \cite{kPPC} requires the cluster center plane not only close to the samples from this cluster but also far away from the samples from other clusters. Instead of solving problems \eqref{kPC} in $k$PC, $k$PPC updates the $i$-th ($i=1,\ldots,k$) cluster center planes \eqref{kPCPlane} by
\begin{equation}\label{kPPC}
\begin{array}{l}
\underset{w_i,b_i}{\min}\sum\limits_{j\in N_i}
(w_i^{\top}x_j+b_i)^2-c\sum\limits_{j\in N\backslash N_i}(w_i^{\top}x_j+b_i)^2\\
\hbox{s.t.\ }~~||w_i||^2=1,
\end{array}
\end{equation}
where $c>0$ is a parameter. The solution to the above problem can also be obtained by solving an eigenvalue problem. Since $k$PC performs unstable from its stochastic initialization, a Laplacian graph-based initialization is used in $k$PPC to obtain stable results.

Due to the planes used in $k$PC and $k$PPC extend infinitely, the following method localizes the cluster center planes with center points.

\subsection{L$k$PPC}
By hiring the cluster centers from $k$means \cite{Kmeans2}, L$k$PPC \cite{LKPPC} supposes a cluster has an extra center point. This yields the following problem for the $i$-th cluster with $i=1,\ldots,k$,
\begin{equation}\label{LkPPC}
\begin{array}{l}
\underset{w_i,b_i,\nu_i}{\min}\sum\limits_{j\in N_i}
(w_i^{\top}x_j+b_i)^2-c_1\sum\limits_{j\in N\backslash N_i}(w_i^{\top}x_j+b_i)^2\\
~~~~~~~~~+c_2\sum\limits_{j\in N_i}||x_j-\nu_i||^2\\
\hbox{s.t.\ }~~~||w_i||^2=1,
\end{array}
\end{equation}
where $\nu_i$ is the center point, and $c_1$ and $c_2$ are the trade-off parameters. The solution to problem \eqref{LkPPC} can be obtained similar to $k$PPC. Once the $k$ cluster center points and planes are obtained, a sample $x$ is assigned into a cluster by
\begin{equation}\label{PredictLkPPC}
\begin{array}{l}
y=\underset{i=1,\ldots,k}{\arg\min}~|w_i^{\top}x+b_i|^2+c_2||x-\nu_i||^2.\\
\end{array}
\end{equation}

\subsection{$k$FC}
$k$FC \cite{Kflat} generalizes the planes in $k$PC by flats, which are defined as
\begin{equation}\label{Flat}
\begin{array}{l}
W_i^\top x-\gamma_i=0,~i=1,\ldots,k,
\end{array}
\end{equation}
where $W_i\in \mathbb{R}^{n\times p},\gamma_i\in \mathbb{R}^{p}$, $1\leq p<n$ is a parameter to control the dimension of flat.

Similar to $k$PC, the cluster center flats and the labels in $k$FC are updated alternately. Thereinto, the cluster center flats are close to their corresponding samples by considering $k$ matrix optimization problems with $i=1,\ldots,k$,
\begin{equation}\label{qflat}
\begin{array}{l}
\underset{W_i,\gamma_i}\min\sum\limits_{j\in N_i}||W_i^{\top}x_j-\gamma_i||^{2}\\
\hbox{s.t.\ }~~ W_i^{\top}W_i=I,
\end{array}
\end{equation}
where $I$ is an identity matrix. The solution to problem \eqref{qflat} can be obtained by solving an eigenvalue problem, and the labels are computed by
\begin{equation}\label{Predictkflat}
\begin{array}{l}
y=\underset{i=1,\ldots,k}{\arg\min}~||W_i^{\top}x-\gamma_i||.
\end{array}
\end{equation}

Apparently, $k$FC is $k$PC if $p=1$. However, $k$FC may suit for more complicated manifolds than $k$PC when $p>1$.

\subsection{L$k$FC}
Similar to L$k$PPC, L$k$FC \cite{Lkflat} introduces the center point into $k$FC, and yields the problem with $i=1,\ldots,k,$
\begin{equation}\label{lqflat}
\begin{array}{l}
\underset{W_i,\gamma_i}{\min}\sum\limits_{j\in N_i}
||W_i^{\top}(x_j-\gamma_i)||^{2}+c\sum\limits_{j\in N_i}||x_j-\gamma_i||^2\\
\hbox{s.t.\ }~~~~ W_i^{\top}W_i=I.
\end{array}
\end{equation}
The above problem can also be convert to an eigenvalue problem, and the labels are updated by
\begin{equation}\label{PredictLqflat}
\begin{array}{l}
y=\underset{i=1,\ldots,k}{\arg\min}~||W_i^{\top}(x-\gamma_i)||^2+c||x-\gamma_i||^2.
\end{array}
\end{equation}
Once the loop between cluster centers and labels terminates, an undirected graph on the current clusters with the affinity matrix is constructed and the samples are clustered into $k$ clusters by some spectral-type clustering methods \cite{Andrew2001On}.


\section{MFPC}

\subsection{Linear Formation}
Recently, a general model of the plane-based clustering has been given in \cite{wang2019general}. As its extension to flat-type clustering, for each cluster we find a $q$-dimensional flat
\begin{eqnarray}\label{GENflat}
\begin{array}{l}
W_i^\top(x_j-\bar{x}_i)=0
\end{array}
\end{eqnarray}
by the following general model with variables $W_i\in \mathbb{R}^{n\times p}$ $(i=1,\ldots,k)$ and labels $y_j$ $(j=1,\ldots,m)$ as
\begin{eqnarray}\label{GENSUB}
\begin{array}{l}
\underset{W_i,y_\cdot}{\min}\sum\limits_{i=1}^{k}||W_i||_\mathcal{F}+\sum\limits_{j=1}^{m} L(y_j,x_j,W_1,\ldots,W_k),\\
\end{array}
\end{eqnarray}
where $y_\cdot$ denotes $\{y_j|j=1,\ldots,m\}$, $||W_i||_\mathcal{F}$ is the regularization in the functional space $\mathcal{F}$ to control the complexity of the model, and $L(\cdot)$ is the loss of a sample assigning to a cluster.

Following the general model \eqref{GENSUB} and corresponding to $q$-dimensional flat for each cluster, we seek $k$ matrices $W_i=(w_{i,1},\ldots,w_{i,p})\in \mathbb{R}^{n\times p}$ with $i=1,\ldots,k$, where $W_i$ yields the $i$-th projection subspace spanned by its column vectors $w_{i,1},\ldots,w_{i,p}$ and $p=n-q$ is parameter.
Specifically, by using the symmetric hinge loss function \cite{TWSVC,wang2019general},
our linear MFPC solves $k$ matrix optimization subproblems with $i=1,\ldots,k$ as
\begin{eqnarray}\label{MSSVCmain}
\begin{array}{l}
\underset{W_{i},\xi_{i,\cdot}}{\min}~~\frac{1}{2}||W_i||_F^2+
\frac{c_{1}}{2}\underset{j\in N_i}\sum||W_{i}^{\top}(x_j-\bar{x}_i)||^2+c_{2}\underset{j\in N\backslash N_i}\sum\xi_{i,j}\\
\hbox{s.t.\ }
~~~~||W_{i}^{\top}(x_j-\bar{x}_i)||\geq 1-\xi_{i,j},~\xi_{i,j}\geq 0,~j\in N\backslash N_i,\\
~~~~~~~~\underset{j\in N}\sum||W_i^{\top}x_j||^2=1,\\
~~~~~~~~W_i^{\top}W_i~~ \text{is a diagonal matrix},
\end{array}
\end{eqnarray}
where $\bar{x}_i$ is the center of the $i$-th cluster, $c_1$ and $c_2$ are positive parameters, and $\xi_{i,\cdot}=\{\xi_{i,j}\in\mathbb{R}|j\in N\backslash N_i\}$ is the set of slack variables.

\begin{figure*}[htbp]
\centering
   \subfigure[Samples projected by $W_1$]{\includegraphics[width=0.18\textheight]{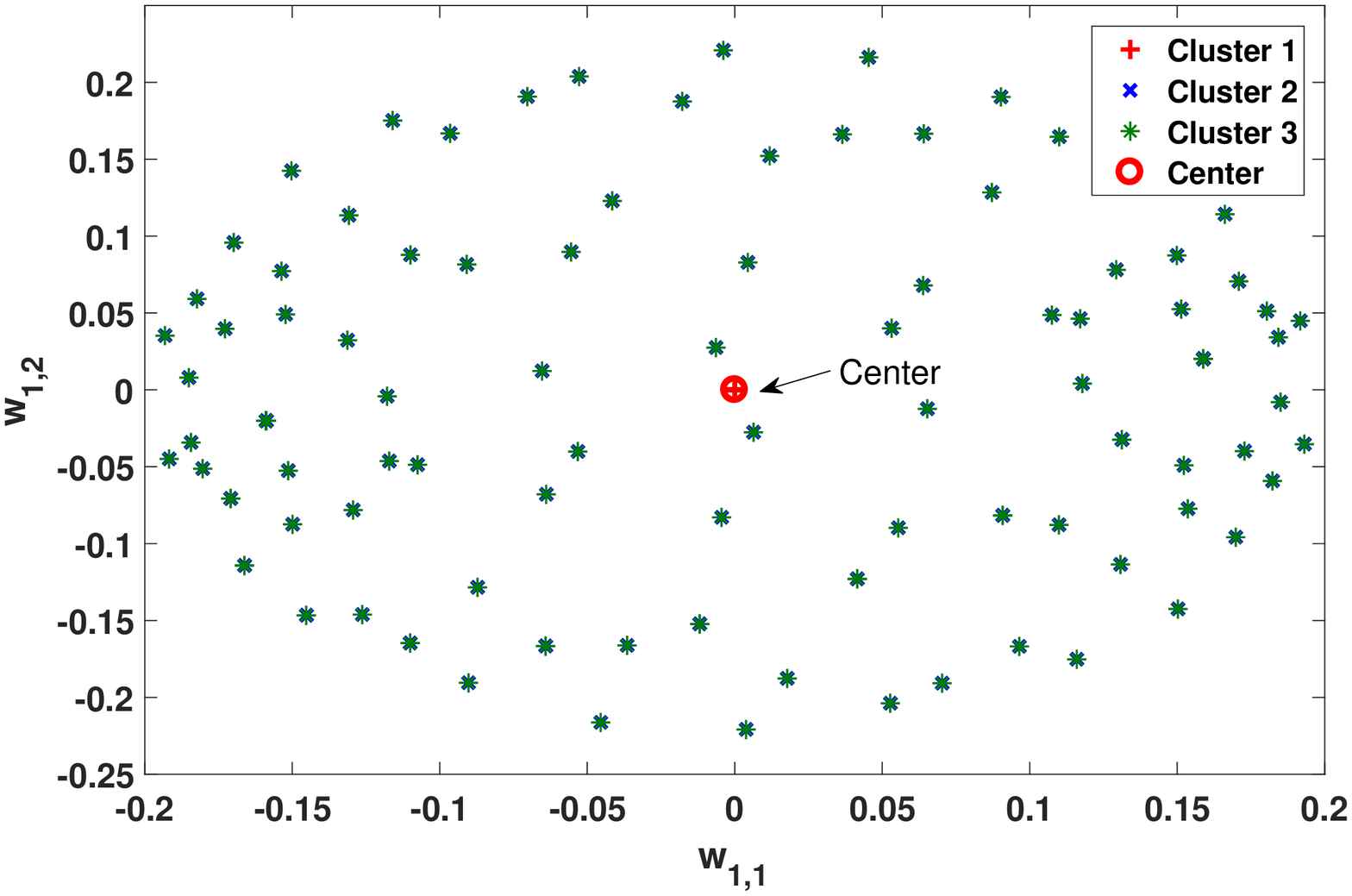}}
   \subfigure[Samples projected by $W_2$]{\includegraphics[width=0.18\textheight]{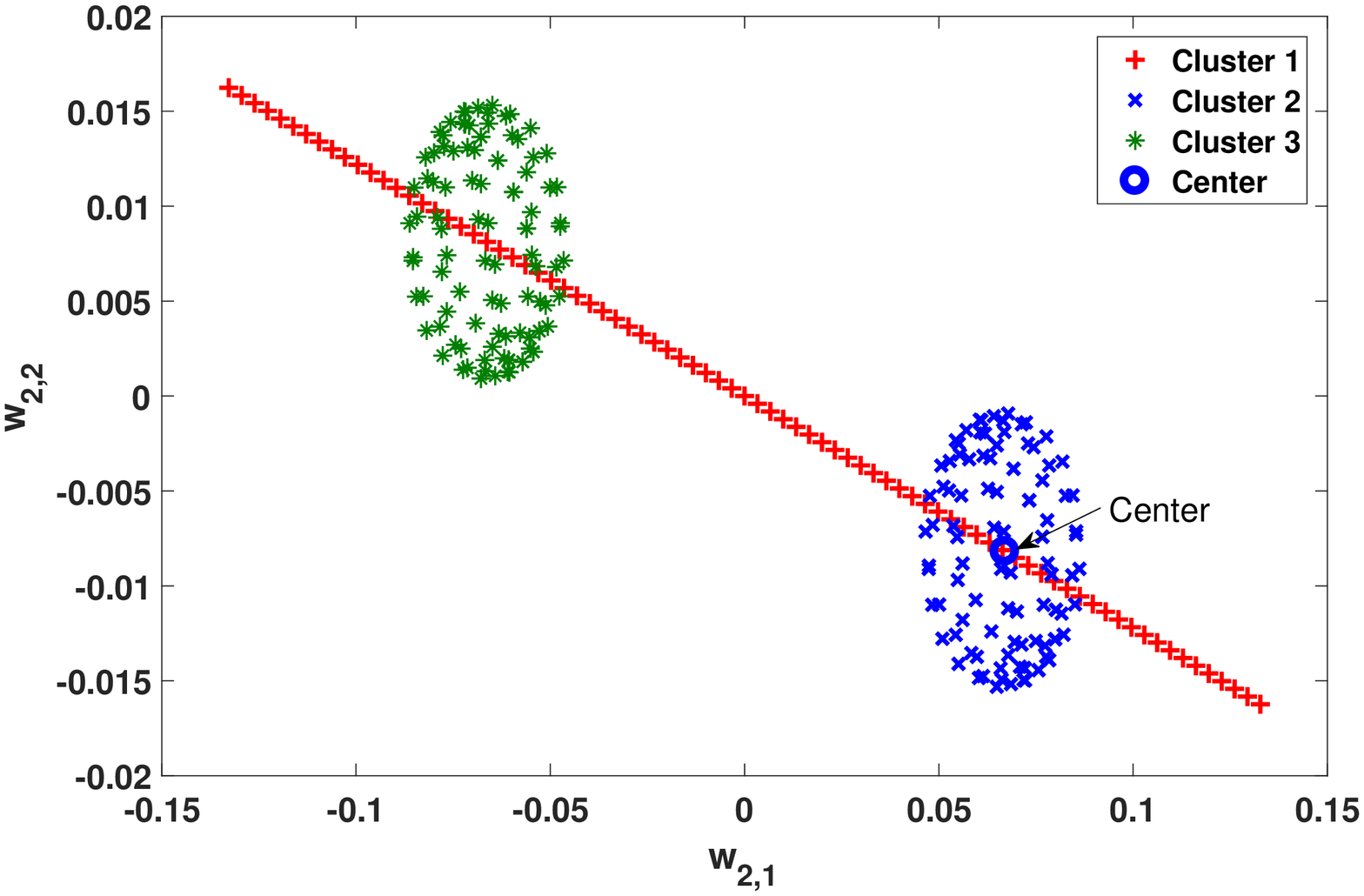}}
   \subfigure[Samples projected by $W_3$]{\includegraphics[width=0.18\textheight]{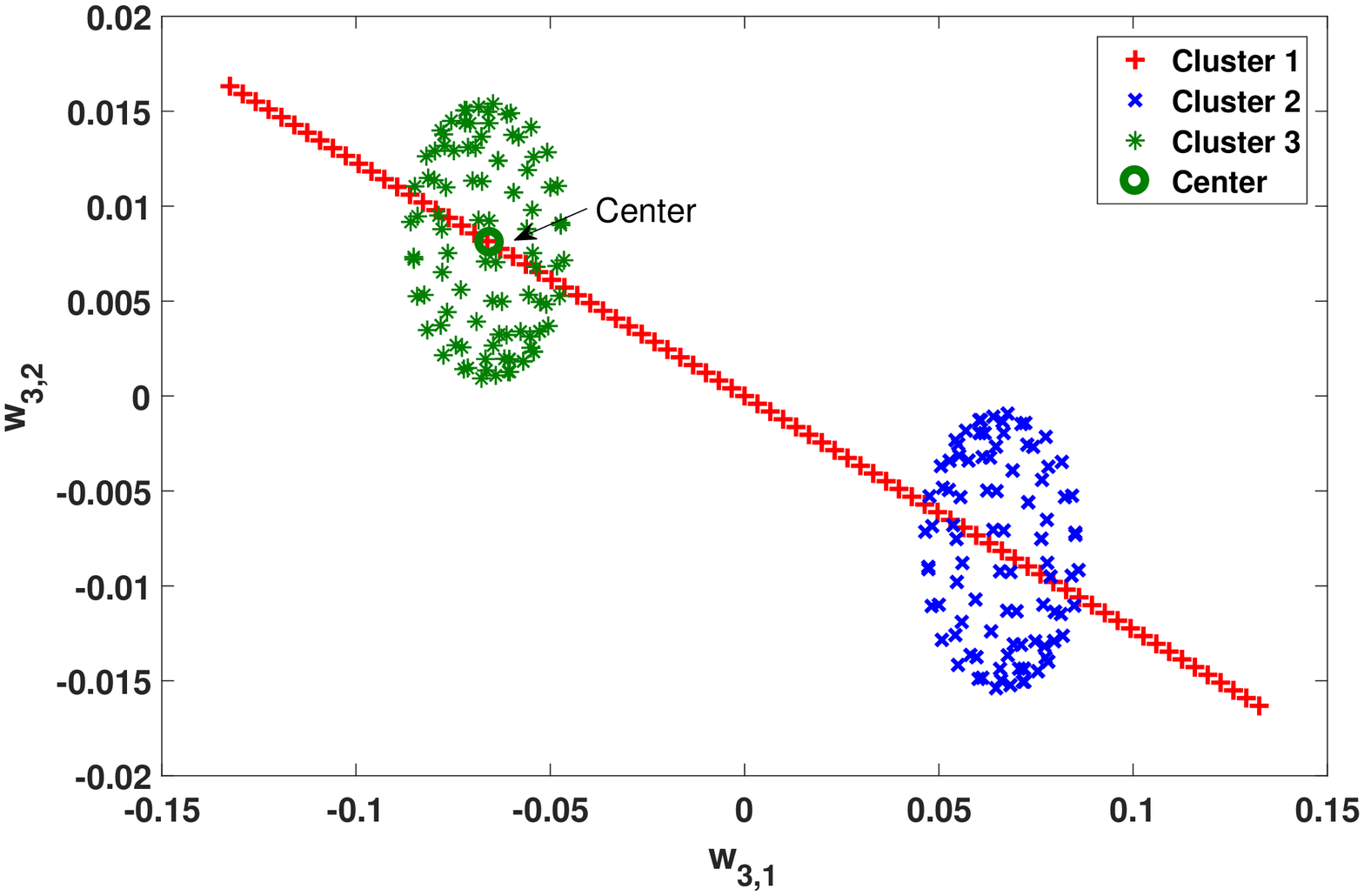}}
   \subfigure[Decision values]{\includegraphics[width=0.18\textheight]{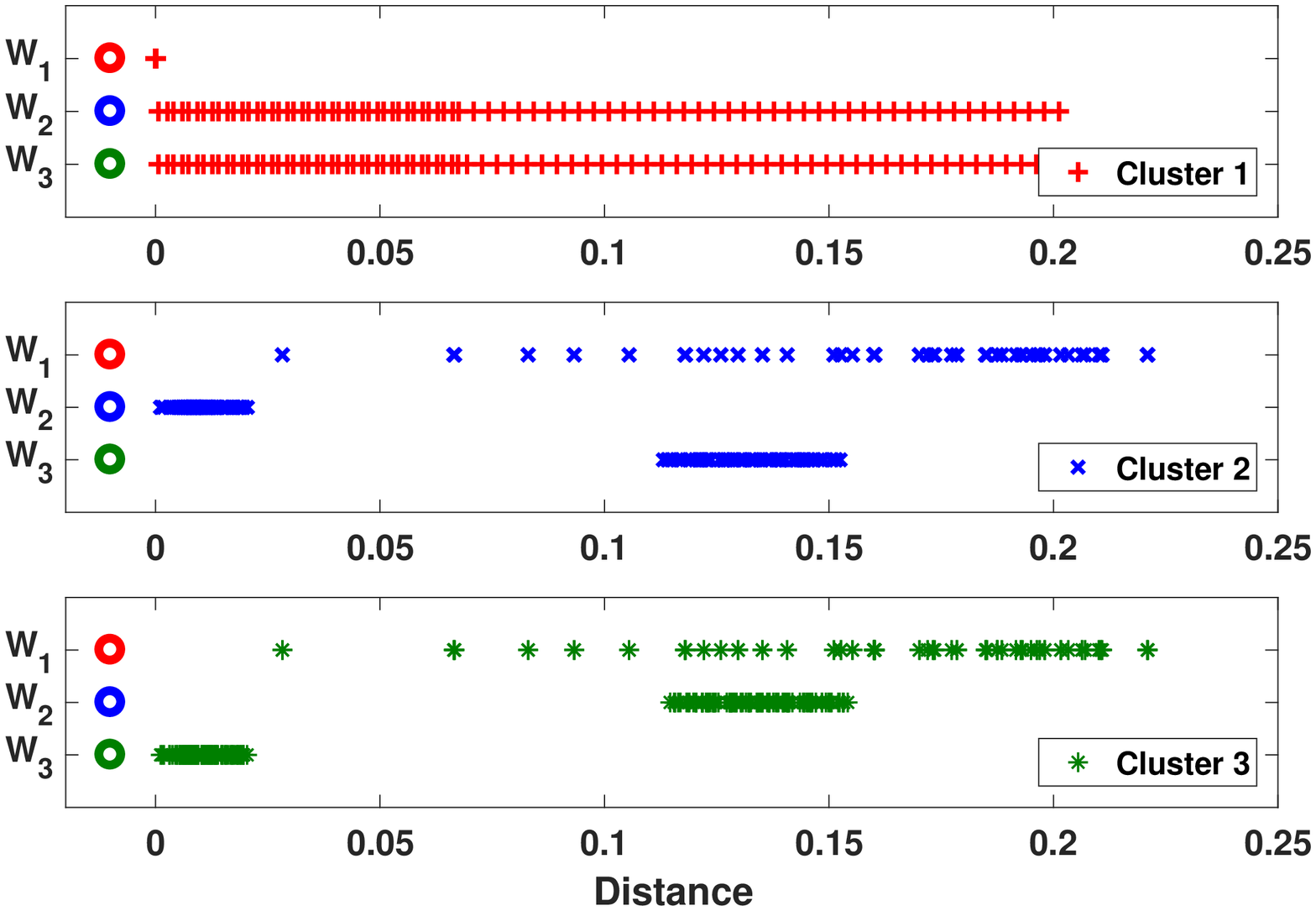}}
\centering\caption{Illustrations of the projected samples in three subspaces and their decision values from the subspaces' centers by MFPC, where two clusters overlap after projection in $W_1$, and the sample projection by $W_2$ and $W_3$ are the same except for the projection center.}\label{Tanghuluprojection}
\end{figure*}

The geometric interpretation of problem \eqref{MSSVCmain} is clear. The second term in the objective function shows that a sample $x$ belonging to the $i$-th cluster would be projected by $W_i$ (i.e., $W_i^{\top}x$) as close as possible to the projected cluster center $W_i^\top \bar{x}_i$. The first constraint requires that for a sample $x$ belonging to other clusters, the projection $W_i^{\top}x$ would be far away from the projected cluster center $W_i^{\top}\bar{x}_i$ to some extent.
In addition, the matrices $W_i$ $(i=1,\ldots, k)$ are normalized by the second constraint, which keeps the manifolds in the subspace with uniform measurement. The third constraint guarantees the column orthogonality of the matrices $W_i,~i=1,\ldots,k$.
The following theorem guarantees the maximum scatter of between-clusters (see the proof in Appendix A).
\begin{thm}
Under the condition that the first constraint strict holds in \eqref{MSSVCmain}, minimizing the regularization term in the objective is equivalent to maximizing the smallest distance between the samples of other clusters and the center of the current cluster in the projection subspace.
\end{thm}
It is easy to prove that the equality constraint $\underset{j\in N}\sum||W_i^{\top}x_j||^2=1$ provides the following property.
\begin{Property}
All samples are projected in a unit ball in each projection subspace.
\end{Property}
%

Starting from an initial sample assignment, our MFPC solves $k$ subproblems \eqref{MSSVCmain} to obtain $k$ projections $W_i$ with $i=1,\ldots,k$. Then, the samples are reassigned into the clusters by their decision values (i.e., the distances of the sample projection from each center projection) as
\begin{equation}\label{FMDlabel}
\begin{array}{l}
y=\underset{i=1,\ldots,k}{\arg}{\min}||W_i^{\top}x-W_i^{\top}\bar{x}_i||.
\end{array}
\end{equation}
The projection matrix and assignment are updated alternately until a repeated overall assignment and a non-decrease in the overall objective \eqref{GENSUB} appear simultaneously.

Now, let us explain the behavior of the projection subspaces generated by our MFPC shown in Fig. \ref{Tanghulu1}(j). Fig. \ref{Tanghuluprojection} plots the three projection subspaces denoted by $W_1=(w_{1,1},w_{1,2})$, $W_2=(w_{2,1},w_{2,2})$ and $W_3=(w_{3,1},w_{3,2})$, where Fig. \ref{Tanghuluprojection}(a-c) show the projected samples in the corresponding subspaces and Fig. \ref{Tanghuluprojection}(d) shows the distances between the sample projections and the subspaces' centers (i.e., the centers' projections). It can be see that the samples of cluster 1 are projected onto a point around and other samples overlap and are far away from it in Fig. \ref{Tanghuluprojection}(a). The projected samples in Fig. \ref{Tanghuluprojection}(b) are the same as Fig. \ref{Tanghuluprojection}(c) but with different center projection. Obviously, the projected samples of cluster 2 are close to the center in Fig. \ref{Tanghuluprojection}(b), and the projected samples of cluster 3 are close to the center in Fig. \ref{Tanghuluprojection}(c). Hence, the samples on the three manifolds are clustered into three correct clusters according to \eqref{FMDlabel} together with Fig. \ref{Tanghuluprojection}(d).
%


\subsection{Solution of MFPC}
In this subsection, we discuss the solution to problem \eqref{MSSVCmain}, which is decomposed into $p$ subproblems recursively. Suppose $w_{i,l}~(l=1,\ldots,p)$ is the $l$-th column of $W_i$ and define the within-cluster scatter matrix \cite{RDA} as
\begin{eqnarray}\label{MSSVCSi}
S_{i}=\underset{j\in N_i}\sum(x_j-\bar{x}_i)(x_j-\bar{x}_i)^{\top}.
\end{eqnarray}
The first subproblem (i.e., $l=1$) is
\begin{eqnarray}\label{MSSVC1}
\begin{array}{l}
\underset{w_{i,l},\xi_{i,\cdot}}{\min}~~\frac{1}{2}||w_{i,l}||^{2}+
\frac{c_{1}}{2}w_{i,l}^{\top}S_{i}w_{i,l}+c_{2}\underset{j\in N \backslash N_i}\sum\xi_{i,j}\\
\hbox{s.t.}
~~||w_{i,l}^{\top}(x_j-\bar{x}_i)||\geq 1-\xi_{i,j},~\xi_{i,j}\geq 0,~j\in N \backslash N_i,\\
~~~~~\underset{j\in N}\sum(w_{i,l}^{\top}x_j)^{2}=1,
\end{array}
\end{eqnarray}
which is a non-convex problem evidently.

In the following, we solve problem \eqref{MSSVC1} by combining the penalty function algorithm and concave-convex procedure (CCCP) \cite{CCCP}. Consider the unconstraint penalty formation of problem \eqref{MSSVC1}:
\begin{eqnarray}\label{MSSVCW}
\begin{array}{l}
\underset{w_{i,l}}{\min}~~\frac{1}{2}||w_{i,l}||^{2}+
\frac{c_{1}}{2}w_{i,l}^{\top}S_{i}w_{i,l}+c_{2}\underset{j\in N\backslash N_i}\sum (1- \\~~~~~~~~||w_{i,l}^{\top}(x_j-\bar{x}_i)||)_++\frac{1}{2}\sigma|\underset{j\in N}\sum(w_{i,l}^{\top}x_j)^{2}-1|,\\
\end{array}
\end{eqnarray}
where $(\cdot)_+$ replaces the negative value by zero, and $\sigma>0$ is the penalty parameter. Note that
\begin{eqnarray}\label{ConvexConcave1}
\begin{array}{ll}
&(1-||w_{i,l}^{\top}(x_j-\bar{x}_i)||)_+\\
=&1-|w_{i,l}^{\top}(x_j-\bar{x}_i)|+(|w_{i,l}^{\top}(x_j-\bar{x}_i)|-1)_+,
\end{array}
\end{eqnarray}
and
\begin{eqnarray}\label{ConvexConcave2}
\begin{array}{ll}
&|\underset{j\in N}\sum(w_{i,l}^{\top}x_j)^{2}-1|\\
=&1-\underset{j\in N}\sum(w_{i,l}^{\top}x_j)^{2}+2(\underset{j\in N}\sum(w_{i,l}^{\top}x_j)^{2}-1)_+.
\end{array}
\end{eqnarray}
Substitute \eqref{ConvexConcave1} and \eqref{ConvexConcave2} into \eqref{MSSVCW} and we have its equivalent as
\begin{eqnarray}\label{MSSVC2}
\begin{array}{l}
\underset{w_{i,l}}{\min}~~F_{vex}(w_{i,l})+F_{cav}(w_{i,l}),
\end{array}
\end{eqnarray}
where $F_{vex}(w_{i,l})=\frac{1}{2}||w_{i,l}||^{2}+
\frac{c_{1}}{2}w_{i,l}^{\top}S_{i}w_{i,l}+c_{2}\underset{j\in N\backslash N_i}\sum(|w_{i,l}^{\top}(x_j-\bar{x}_i)|-1)_++\sigma(\underset{j\in N}\sum(w_{i,l}^{\top}x_j)^{2}-1)_+$ and $F_{cav}(w_{i,l})=-c_{2}\underset{j\in N\backslash N_i}\sum|w_{i,l}^{\top}(x_j-\bar{x}_i)|-\frac{1}{2}\sigma\underset{j\in N}\sum(w_{i,l}^{\top}x_j)^{2}$. It is easy to conclude that $F_{vex}(w_{i,l})$ is convex and $F_{cav}(w_{i,l})$ is concave w.r.t. $w_{i,l}$. Thus, problem \eqref{MSSVC2} is also called difference of convex functions (DC) problem \cite{DC}. Here, we construct a series of problems with $t=0,1,2,\ldots$ as
\begin{eqnarray}\label{MSSVCtmp1}
\begin{array}{l}
\underset{w_{i,l}^{(t+1)}}{\min}~~F_{vex}(w_{i,l}^{(t+1)})+\nabla F_{cav}(w_{i,l}^{(t)})^\top w_{i,l}^{(t+1)},
\end{array}
\end{eqnarray}
where
\begin{eqnarray}\label{Subgradient}
\begin{array}{l}
\nabla F_{cav}(w_{i,l}^{(t)})
=-c_{2}\underset{j\in N\backslash N_i}\sum\text{sign}(w_{i,l}^{(t)\top}(x_j-\bar{x}_i))(x_j-\bar{x}_i)\\
~~~~~~~~~~~~~~~~~~-\sigma\underset{j\in N}\sum(w_{i,l}^{(t)\top}x_j)x_j
\end{array}
\end{eqnarray}
is the sub-gradient of $F_{cav}(w_{i,l})$ at $w_{i,l}^{(t)}$.
The above problem \eqref{MSSVCtmp1} is a convex quadratic programming problem (CQPP) and can be solved by many efficient algorithms, e.g., Newton algorithms and coordinate descent \cite{NesterovEfficiency} approaches. The series of problems \eqref{MSSVCtmp1} are solved in sequence until the difference of $w_{i,l}^{(t+1)}$ in the adjacent two steps is smaller than a tolerance, and the final $w_{i,l}^{(t+1)}$ is regarded as the solution of \eqref{MSSVC1}. The above procedures are summarized in Algorithm 1.

\begin{algorithm}[H]\label{alg:cd1}
\caption{Iterative algorithm to solve problem \eqref{MSSVC1}}
\textbf{Input:} Dataset $X$, index set $N_i$ for the $i$-th cluster, positive parameters $c_1,c_2,\sigma$ and a tolerance $tol$ (typically, $tol=1e-3$).\\
1. compute $S_i$ by \eqref{MSSVCSi};\\
2. set $t=0$ and $w_{i,l}^{(0)}$ be the eigenvector of the smallest eigenvalue of $S_i$;\\
3. \textbf{do}
\par\setlength\parindent{1em}(a) compute $F_{cav}(w_{i,l}^{(t)})$ by \eqref{Subgradient};
\par\setlength\parindent{1em}(b) compute $w_{i,l}^{(t+1)}$ by solving CQPP \eqref{MSSVCtmp1};
\par\setlength\parindent{1em}(c) $t=t+1$.\\
\textbf{while} $||w_{i,l}^{(t)}-w_{i,l}^{(t-1)}||>tol$.\\
\textbf{Output:} $w_{i,l}=w_{i,l}^{(t)}$.
\end{algorithm}
In Algorithm 1, $w_{i,l}^{(0)}$ is initialized as the eigenvector of the smallest eigenvalue of $S_i$. In fact, it is the solution to
\begin{equation}\label{IMD1}
\begin{array}{l}
\underset{w_{i,l}}{\min}\underset{{j\in N_i}}\sum(w_{i,l}^{\top}(x_j-\bar{x}_i))^2\\
\hbox{s.t.}~~||w_{i,l}||=1.
\end{array}
\end{equation}
In other words, the initial $w_{i,l}^{(0)}$ keeps the projected cluster samples close to their center.

In addition, we have following convergence theorem from the CCCP convergence theorem immediately (see Theorem 2 in ref. \cite{CCCP}).
\begin{thm}
The sequence $\{w_{i,l}^{(0)},w_{i,l}^{(1)},\ldots\}$ obtained by algorithm 1 converges to a minimum or saddle point to problem \eqref{MSSVCW}.
\end{thm}

Once we obtain the first column $w_{i,1}$ of $W_i$  by solving the first subproblem \eqref{MSSVC1}, other columns of $W_i$ would be obtained recursively as follow: (i) Determine a projection vector $w_{i,l}$;
(ii) Generate the orthocomplement of the given data by $w_{i,l}$ to determine the next projection vector $w_{i,l+1}$.
The recursive algorithm to solve problem \eqref{MSSVCmain} is summarized in Algorithm 2.
\begin{algorithm}[H]\label{alg:cd2}
\caption{Recursive algorithm to solve problem \eqref{MSSVCmain}}
\textbf{Input:} Dataset $X$, index set $N_i$ for the $i$-th cluster, positive parameters $c_1,c_2,\sigma$, an integer $1\leq p<n$ and a tolerance $tol$ (typically, $tol=1e-3$).\\
1. set $l=1$, and computer $w_{i,1}$ by Algorithm 1;\\
2. set $X_l=\{x_{j,l}|x_{j,l}=x_j,j=1,\ldots,m\}$;\\
2. \textbf{for} $l=1,\ldots,p-1$
\par\setlength\parindent{1em}(a) set $\tilde{w}_{i,l}=w_{i,l}/||w_{i,l}||$;
\par\setlength\parindent{1em}(b) compute $X_{l+1}=\{x_{j,l+1}|x_{j,l+1}=x_{j,{l}}-\tilde{w}_{i,l}^{\top}x_{j,{l}}\tilde{w}_{i,l},j=1,\ldots,m\}$;
\par\setlength\parindent{1em}(c) replace $X$ with $X_{l+1}$ in Algorithm 1, and then implement Algorithm 1 to obtain $w_{i,l+1}$;\\
\textbf{Output:} $W_i=(w_{i,1},w_{i,2},\ldots,w_{i,p})$.\\
\end{algorithm}

Specifically, the following theorem guarantees that the solution obtained by Algorithm 2 satisfies the constraint ``$W_i^{\top}W_i$ is a diagonal matrix'' in \eqref{MSSVCmain}.
\begin{thm}
The $p$ projection vectors
$(w_{i,1},w_{i,2},\ldots,w_{i,p})$ obtained by Algorithm 2 are orthogonal to each other.
\end{thm}
See the proof in Appendix B.

\subsection{Nonlinear Formation}
Now, we extend MFPC to the nonlinear case. Suppose $\phi(\cdot)$ is a nonlinear mapping from $\mathbb{R}^n$ to $\mathbb{H}$, where $\mathbb{H}$ is a high dimensional feature space. Our nonlinear MFPC seeks $k$ cluster projections $W_i$ with $i=1,\ldots,k$ in $\mathbb{H}$. The kernel tricks \cite{TWSVC,TBSVC} help us to select an appropriate feature space $\mathbb{H}$ without giving the nonlinear mapping $\phi(\cdot)$. By selecting a kernel function $K(\cdot,\cdot)$ as the inner product in $\mathbb{H}$, the $i$-th ($i=1,\ldots,k$) projection $W_i$ in nonlinear MFPC is obtained by considering the following problem
\begin{eqnarray}\label{NLMSSVC}
\begin{array}{l}
\underset{W_{i},\xi_{i,\cdot}}{\min}~~\frac{1}{2}||W_i||_F^2+
\frac{c_{1}}{2}\underset{j\in N_i}\sum||W_{i}^{\top}(K(x_j,X)-K(\bar{x}_i,X))||^2\\
~~~~~~~~+c_{2}\underset{j\in N\backslash{N}_i}\sum\xi_{i,j}\\
\hbox{s.t.\ }
||W_{i}^{\top}(K(x_j,X)-K(\bar{x}_i,X))||\geq 1-\xi_{i,j},j\in N\backslash{N}_i,\\
~~~~~~~\xi_{i,j}\geq 0,~j\in N\backslash{N}_i,\\
~~~~~~~\underset{j\in N}\sum||W_i^\top K(x_j,X)||^2=1,\\
~~~~~~~~W_i^{\top}W_i~~ \text{is a diagonal matrix},
\end{array}
\end{eqnarray}

The above problem can also be solved by Algorithm 2. The problem corresponding to \eqref{MSSVC1} is
\begin{eqnarray}\label{NLMSSVCW}
\begin{array}{l}
\underset{w_{i,l},\xi_{i,\cdot}}{\min}~~\frac{1}{2}||w_{i,l}||^{2}+
\frac{c_{1}}{2}w_{i,l}^{\top}S_{i}^{\phi}w_{i,l}+c_{2}\underset{j\in N\backslash{N}_i}\sum\xi_{i,j}\\
\hbox{s.t.}
|w_{i,l}^{\top}(K(x_j,X)-K(\bar{x}_i,X)|\geq 1-\xi_{i,j},j\in N\backslash{N}_i,\\
~~~~~\xi_{i,j}\geq 0,~j\in N\backslash{N}_i,\\
~~~~~\underset{j\in N}\sum(w_{i,l}^{\top}K(x_j,X))^{2}=1.
\end{array}
\end{eqnarray}
where
$S_{i}^{\phi}=\underset{j\in N_i}\sum(K(x_j,X)-K(\bar{x}_i,X))(K(x_j,X)-K(\bar{x}_i,X))^{\top}$.

Once we obtain $k$ projections $W_i~(i=1,\ldots,k)$, a sample $x$ is relabeled by
\begin{eqnarray}\label{decisionk}
y=\underset{i=1,\ldots,k}{\arg\min}
||W_i^{\top}{K(x,X)}-W_i^{\top}K(\bar{x}_i,X)||.
\end{eqnarray}

For a large scale dataset $X$, the kernel function $K(\cdot,X)$ transforms the samples into a space with a much higher dimension than linear formation, resulting in a large amount of computations. However, the reduced kernel tricks \cite{RSVM,SGTSVM}, which replaces $K(\cdot,X)$ with $K(\cdot,\tilde{X})$, can reduce the computation efficiently, where $\tilde{X}$ is selected from $X$ randomly and its size is much smaller than $X$.
\subsection{Computational Complexity}
For our MFPC, the main computational cost is in solving the optimization problem \eqref{MSSVCW}. In Algorithm 1, the main computational cost is dominated in solving the CQPP. The time complexity of solving this QPP is generally no more than O($m^3/4+n^3$). Thus, the total complexity of Algorithm 2 is about O($pt(m^3/4+n^3)$), where $t$ is the iterative number and $p$ is the recursive number.
In contrast, other flat-type methods, e.g., $k$PC, $k$PPC and L$k$PPC, which solve eigenvalue problems with the complexity O($n^3$).
\begin{figure*}[htbp]
\centering
\subfigure[Data]{\includegraphics[width=0.14\textheight]{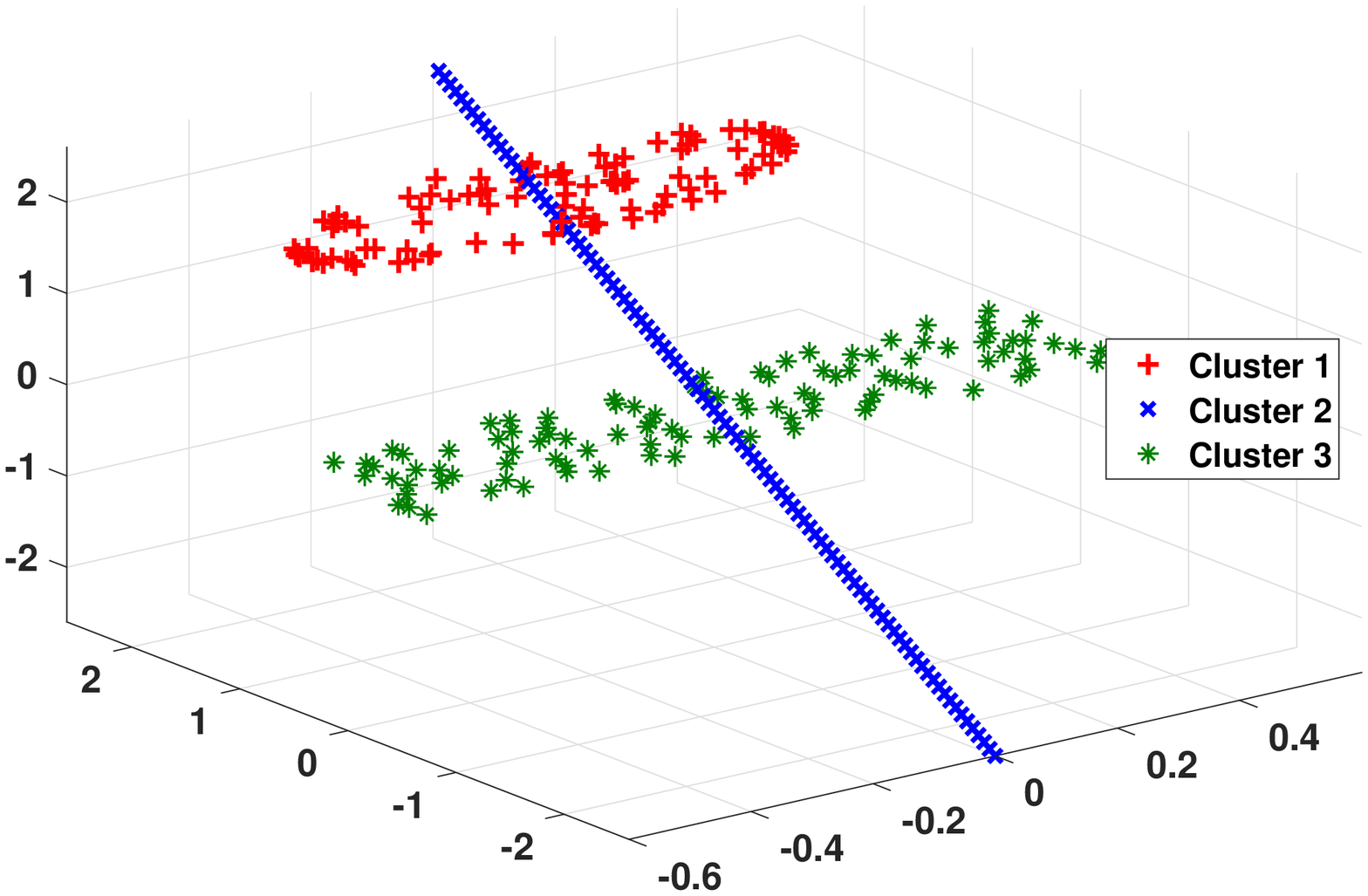}}
\subfigure[SMMC]{\includegraphics[width=0.14\textheight]{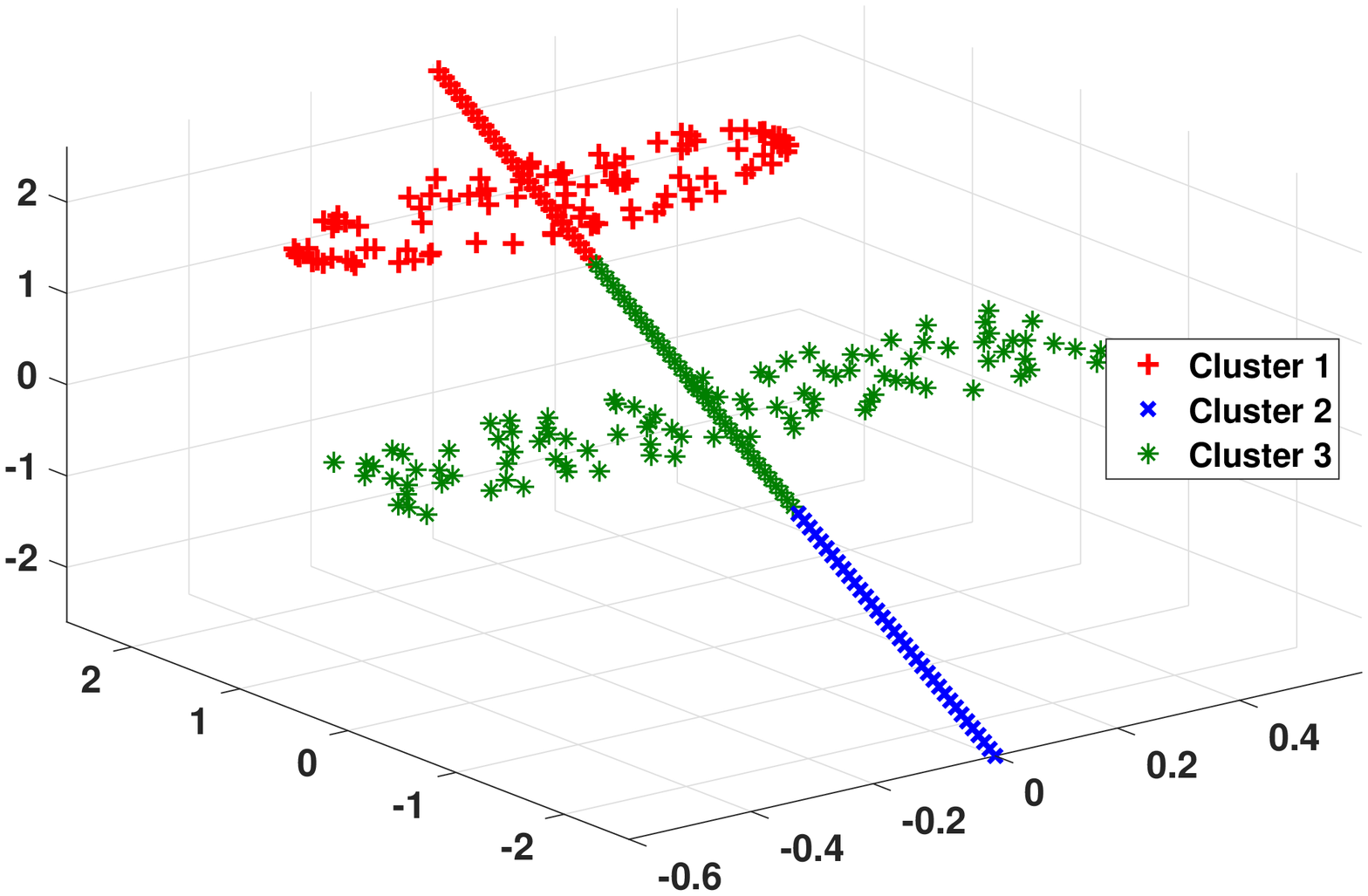}}
\subfigure[$k$means]{\includegraphics[width=0.14\textheight]{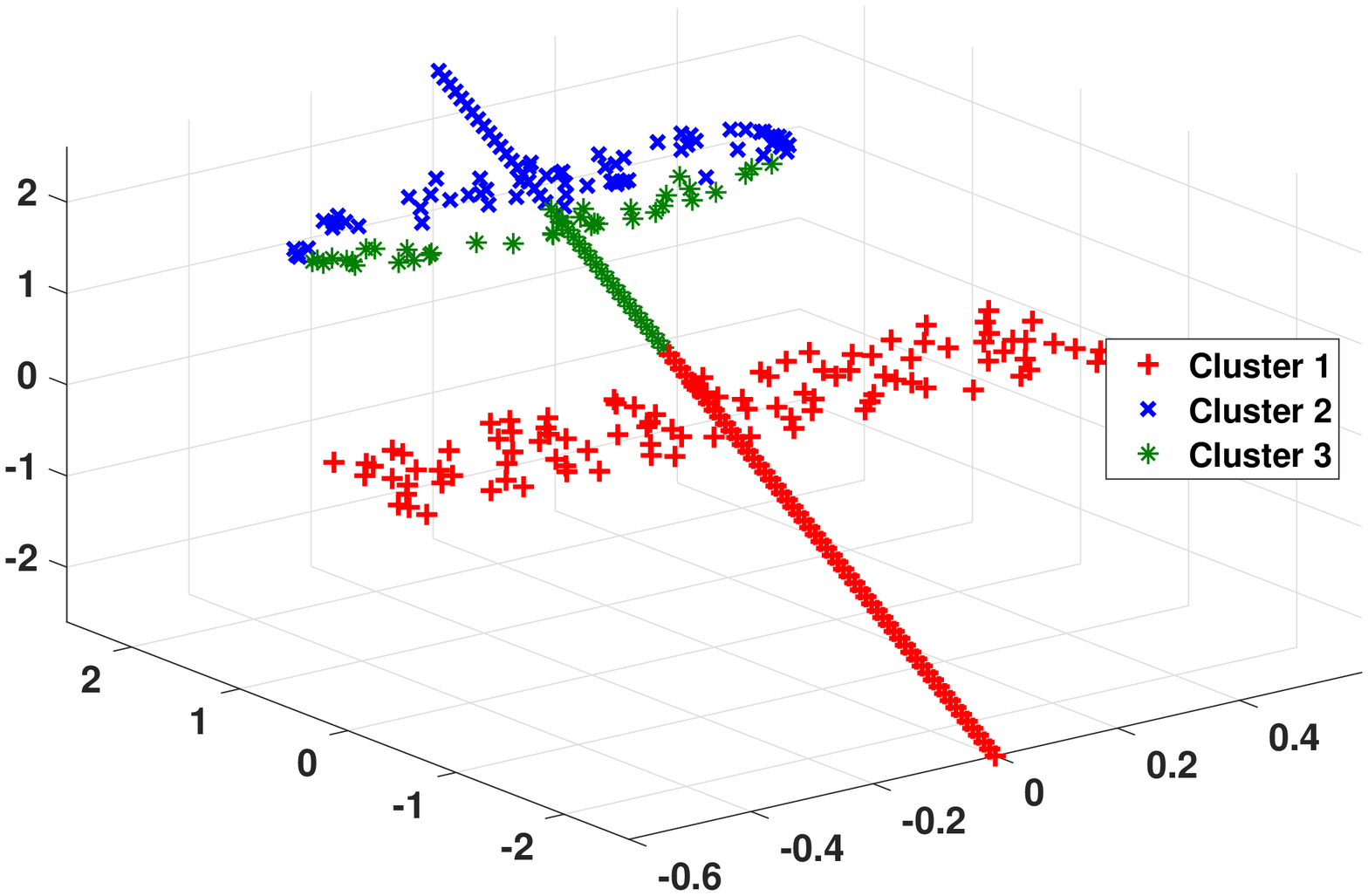}}
\subfigure[$k$PC]{\includegraphics[width=0.14\textheight]{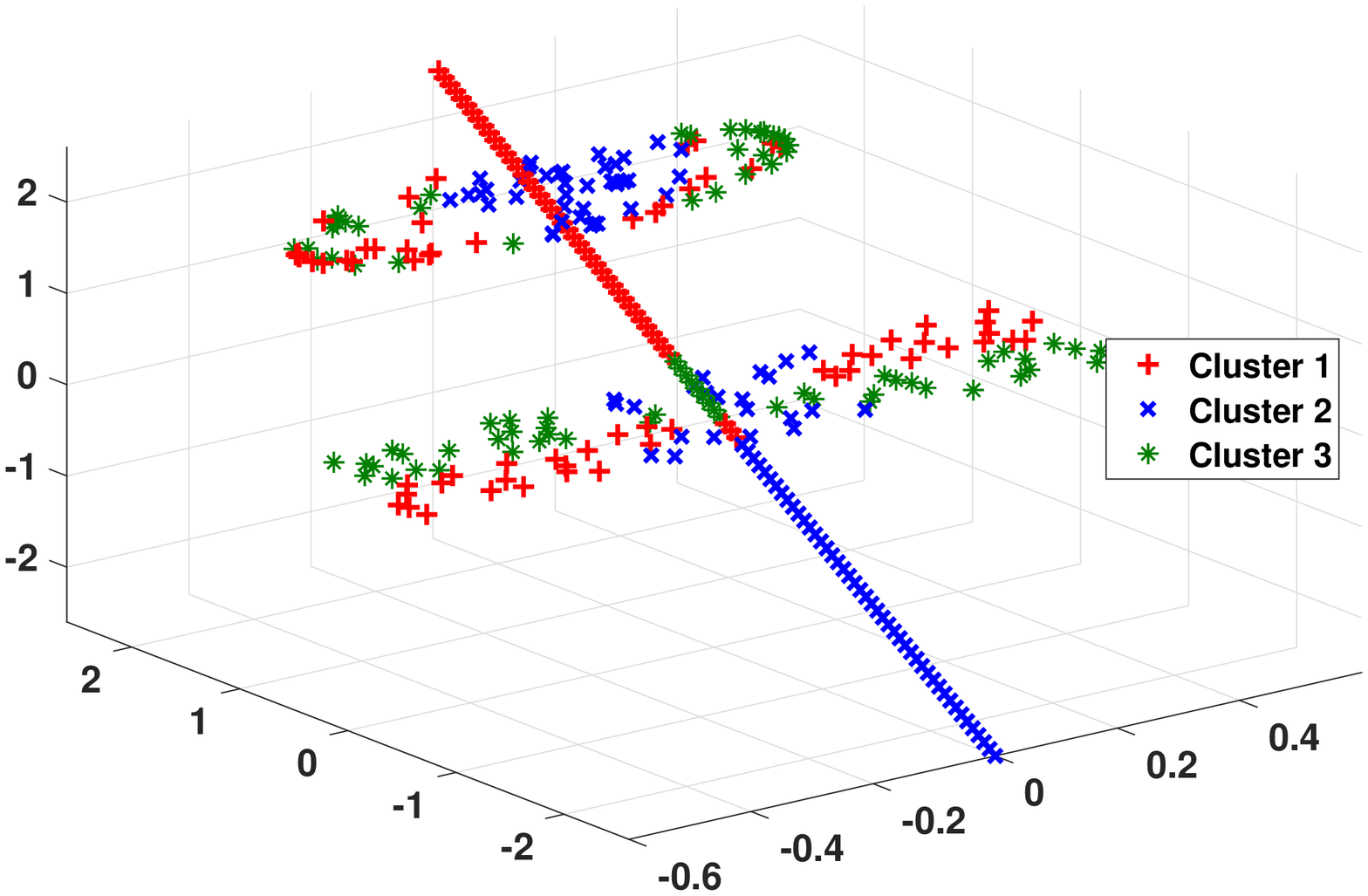}}
\subfigure[$k$PPC]{\includegraphics[width=0.14\textheight]{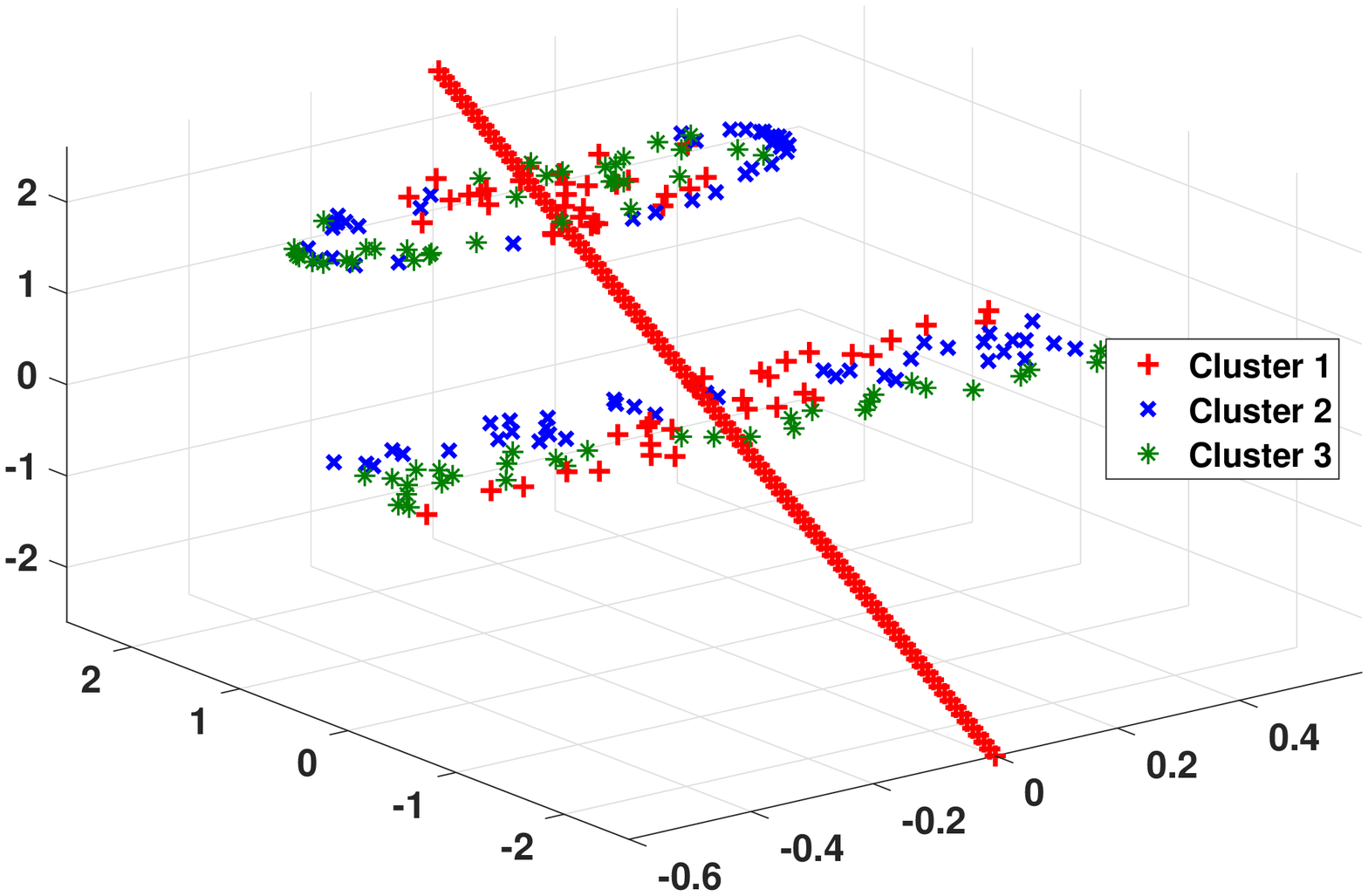}}
\subfigure[L$k$PPC]{\includegraphics[width=0.14\textheight]{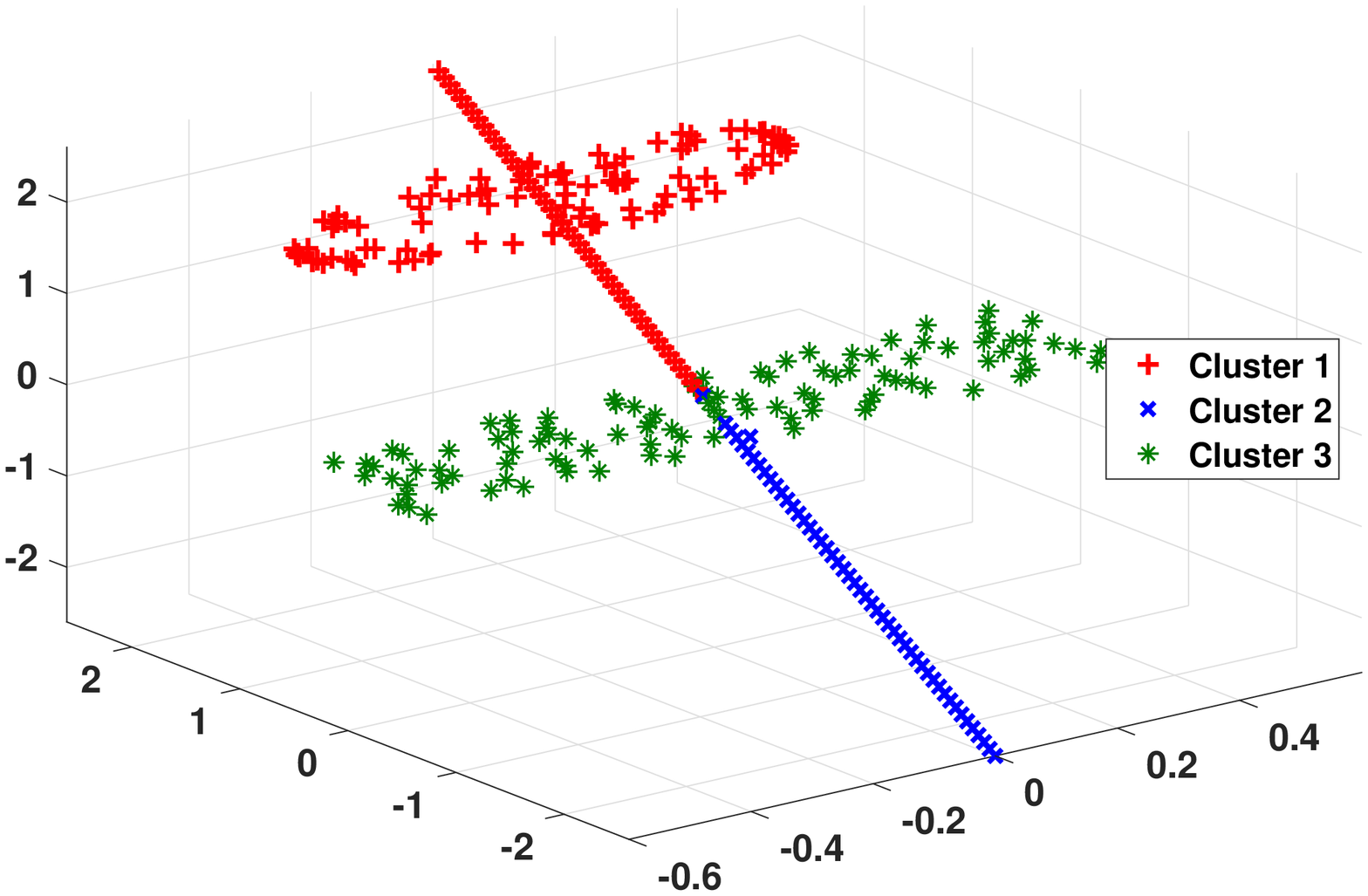}}
\subfigure[TWSVC]{\includegraphics[width=0.14\textheight]{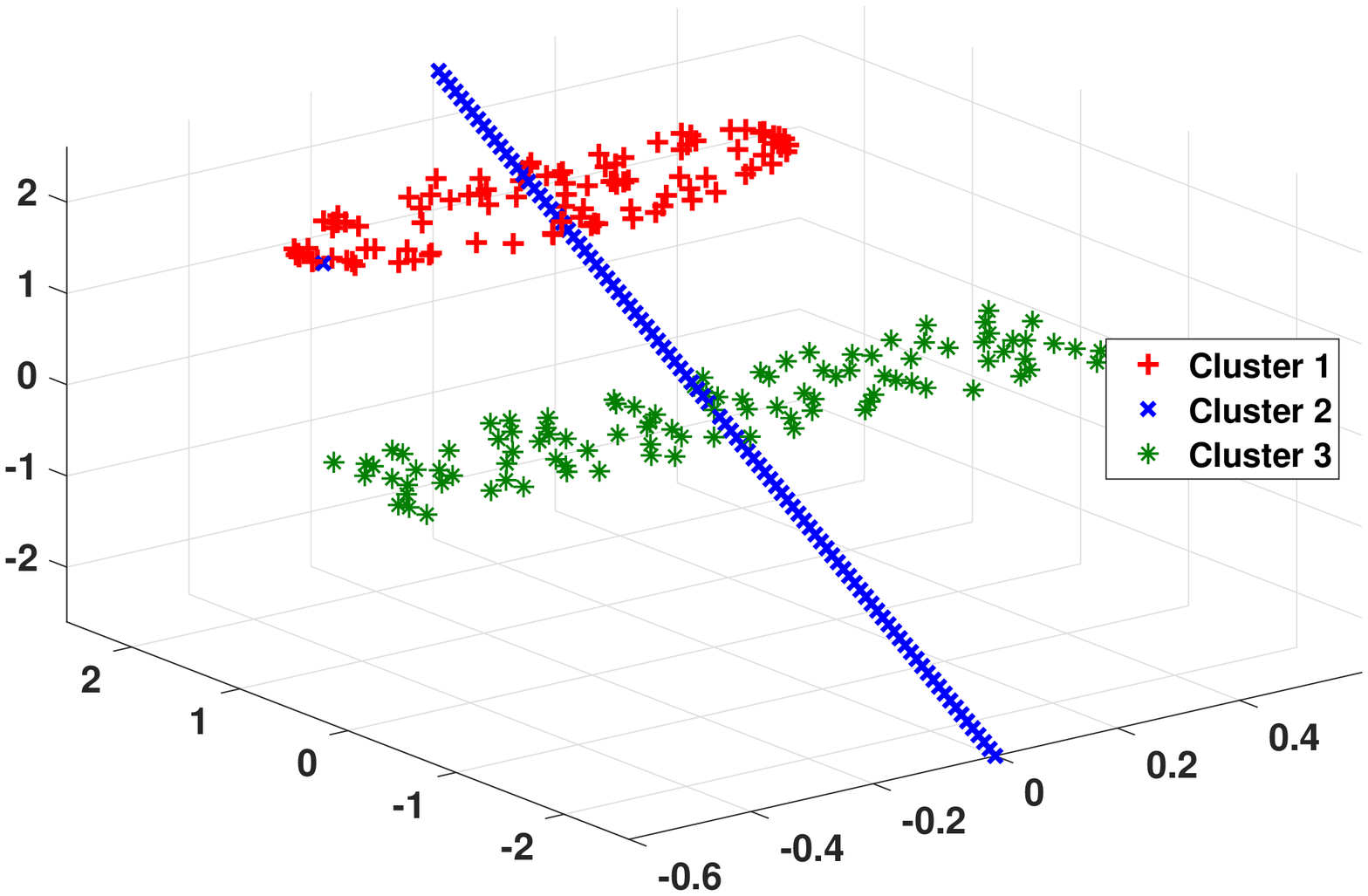}}
\subfigure[$k$FC]{\includegraphics[width=0.14\textheight]{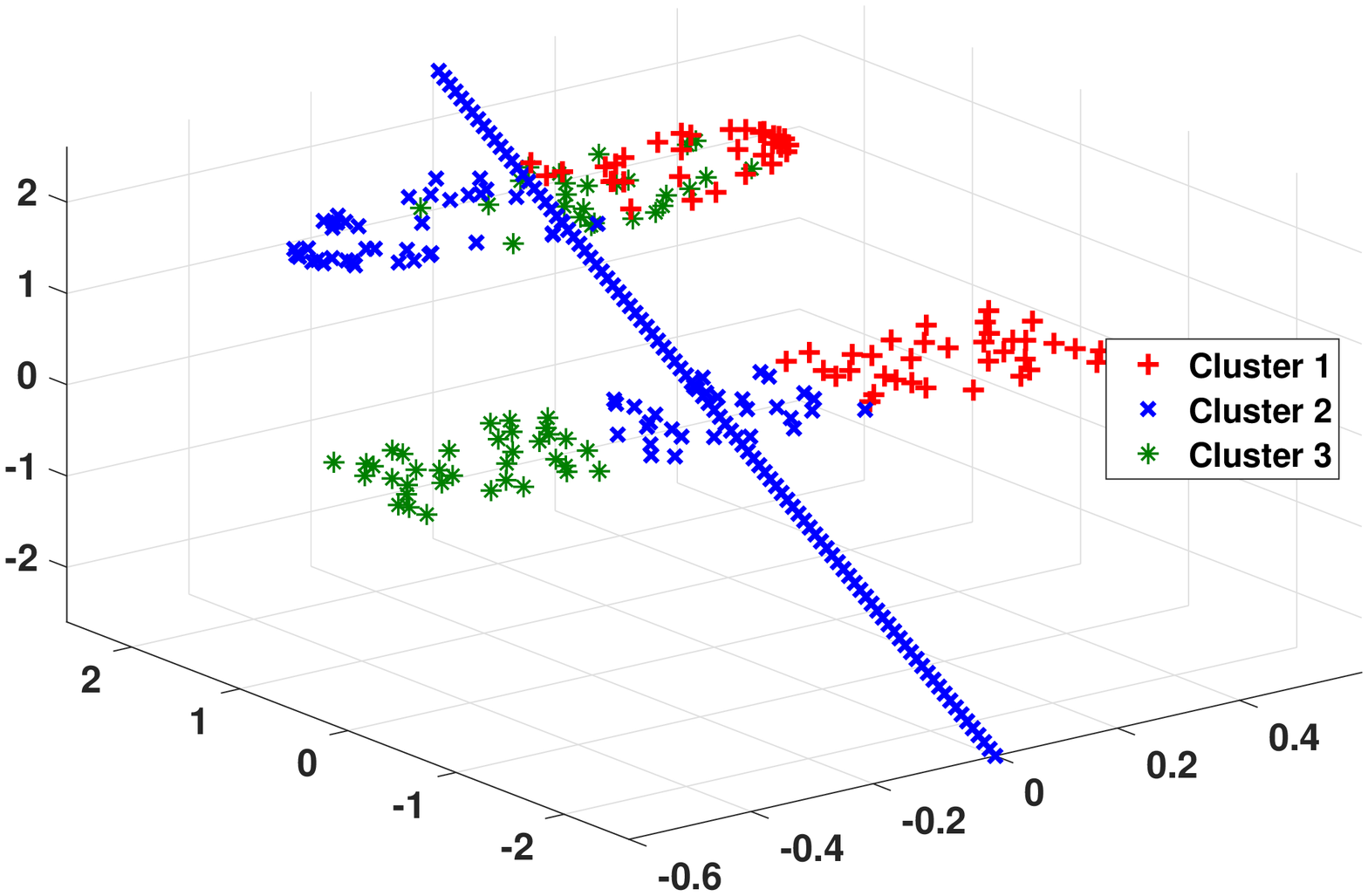}}
\subfigure[L$k$FC]{\includegraphics[width=0.14\textheight]{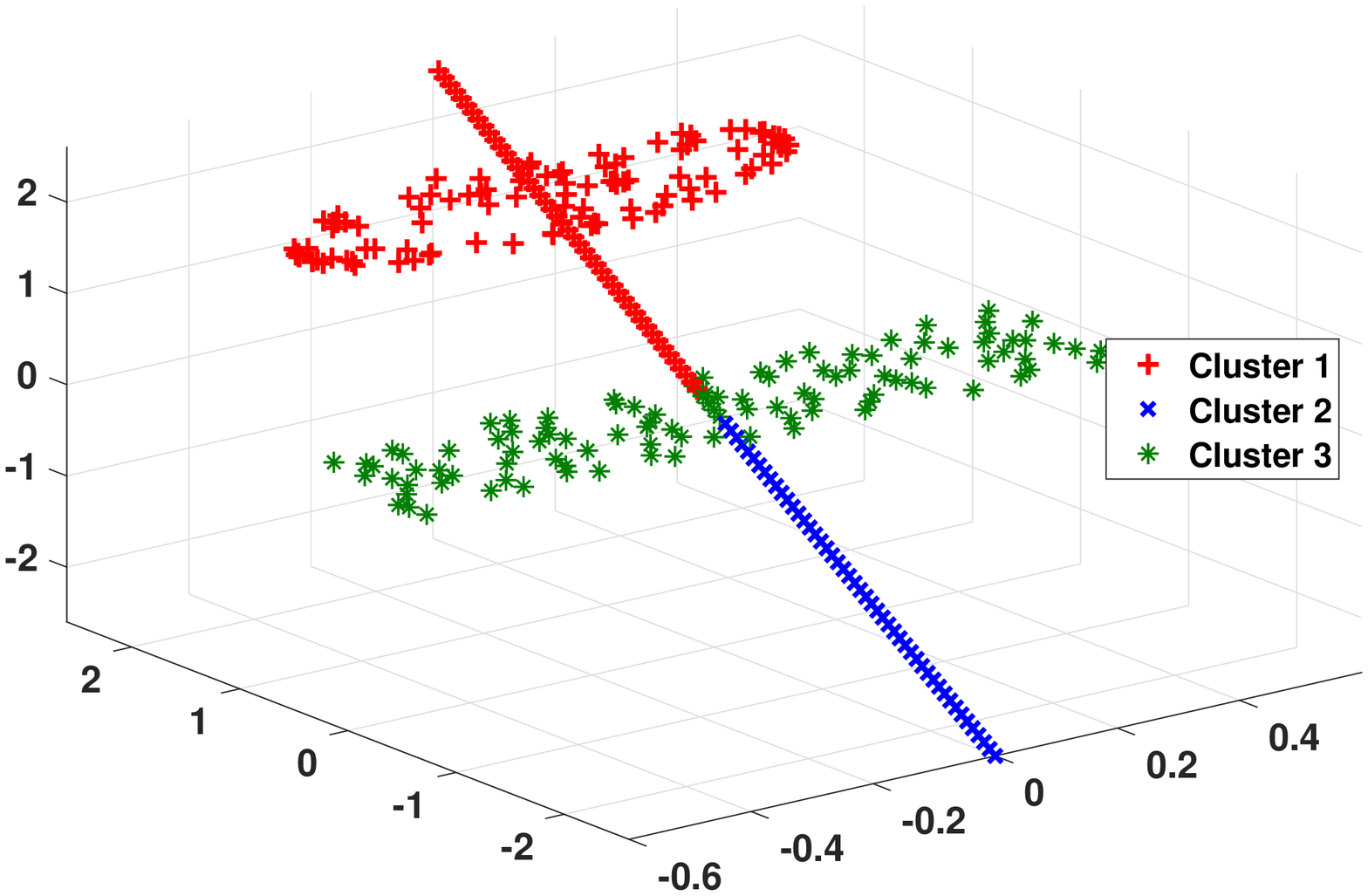}}
\subfigure[MFPC]{\includegraphics[width=0.14\textheight]{LPS_Our.eps}}
\caption{Clustering results of the state-of-the-art methods on the ``LPE'' dataset which includes a line, a plane and an ellipsoid in $R^3$, where the plane and ellipsoid intersect with the line.}\label{FigLPE}
\end{figure*}

\begin{figure*}[htbp]
\centering
\subfigure[Data]{\includegraphics[width=0.14\textheight]{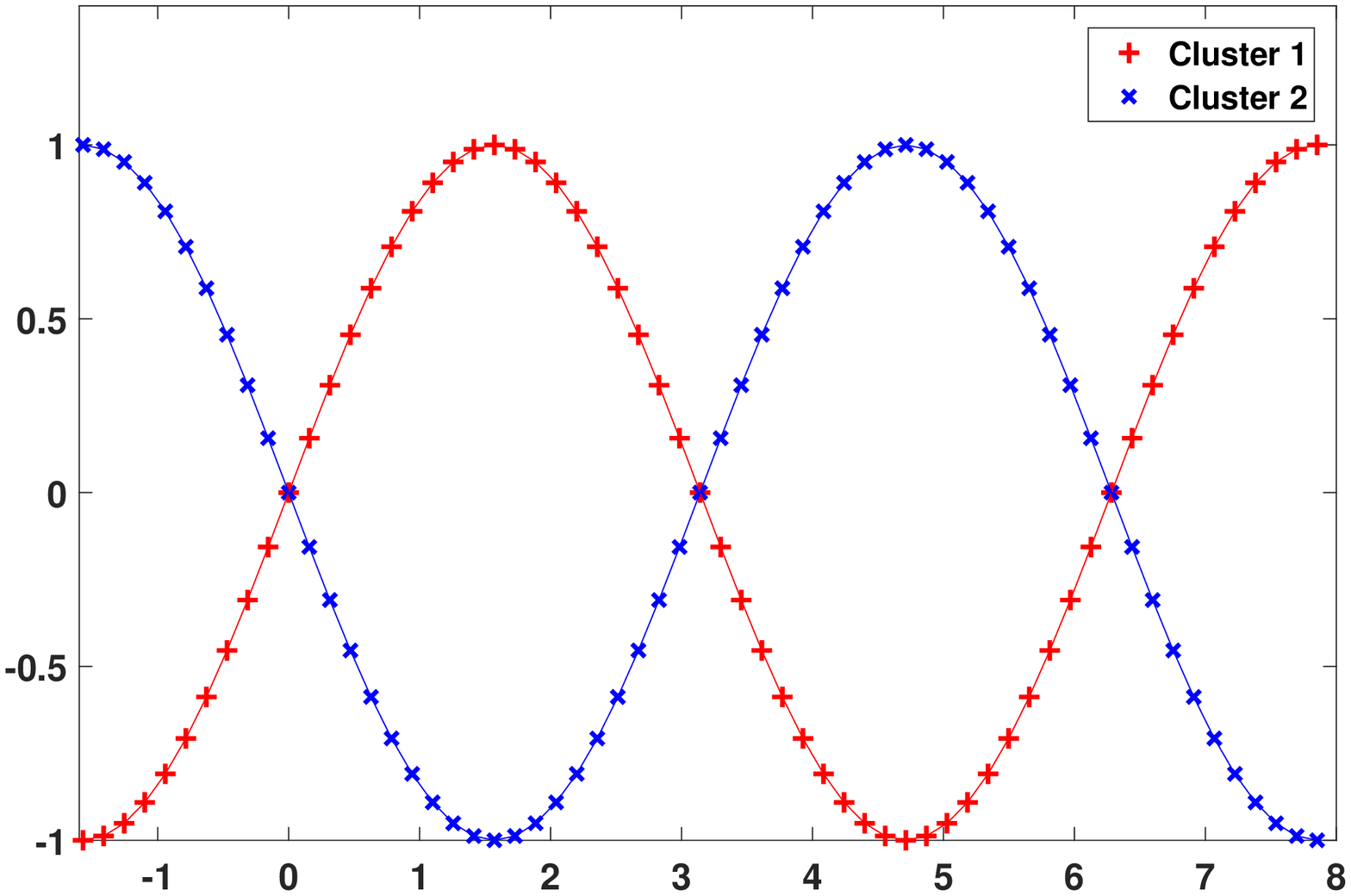}}
\subfigure[$k$means]{\includegraphics[width=0.14\textheight]{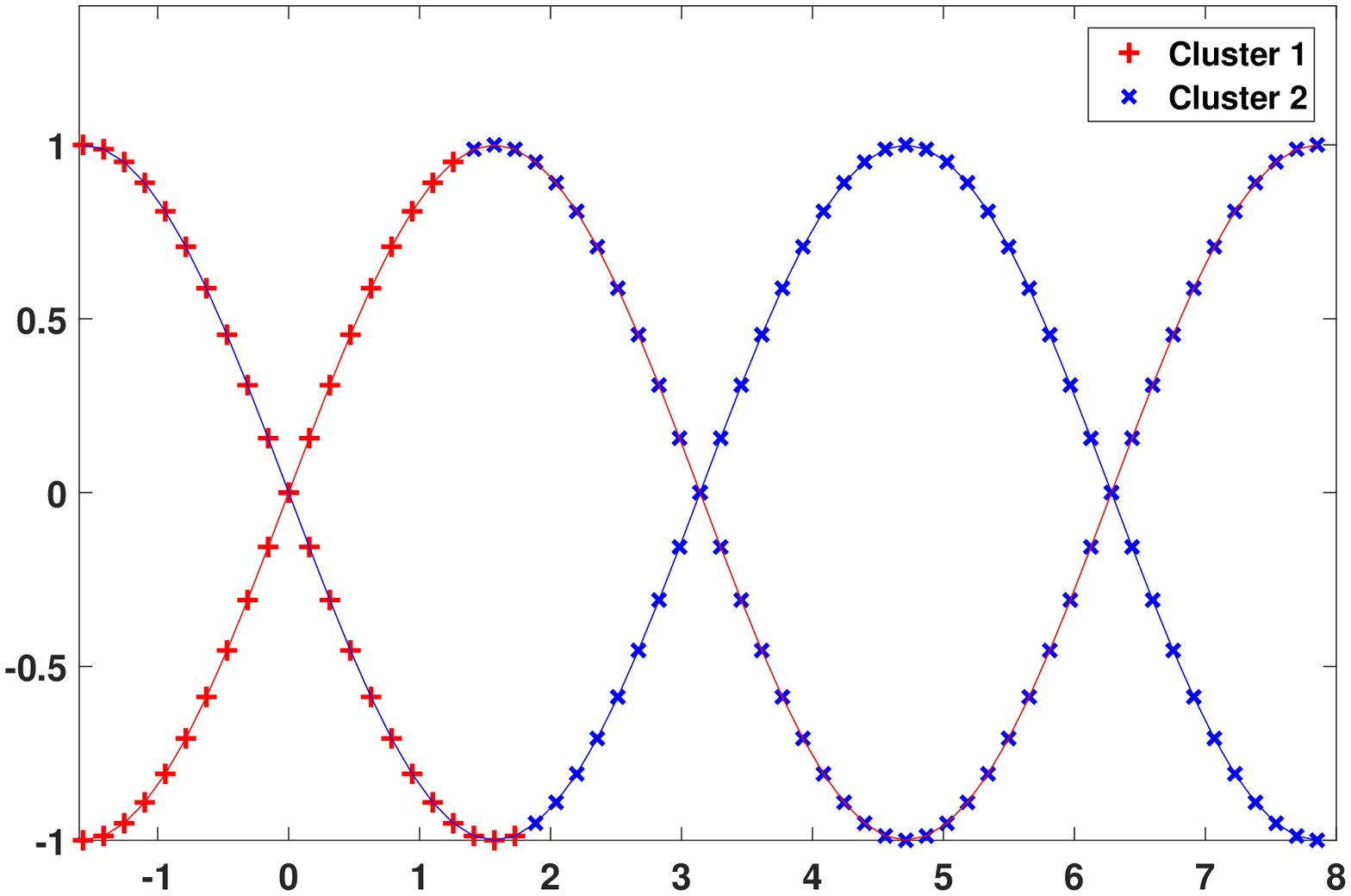}}
\subfigure[SMMC]{\includegraphics[width=0.14\textheight]{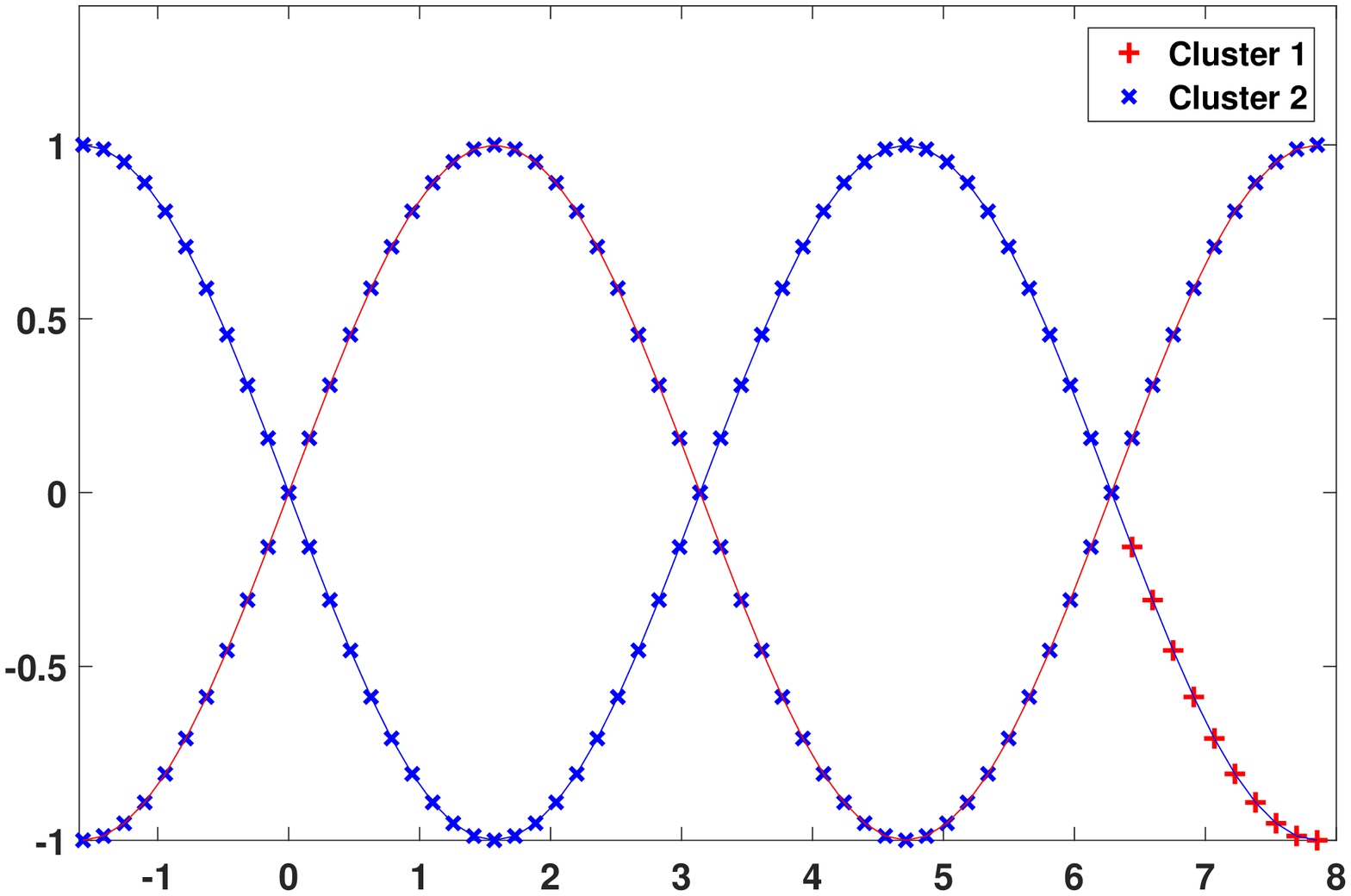}}
\subfigure[$k$PC]{\includegraphics[width=0.14\textheight]{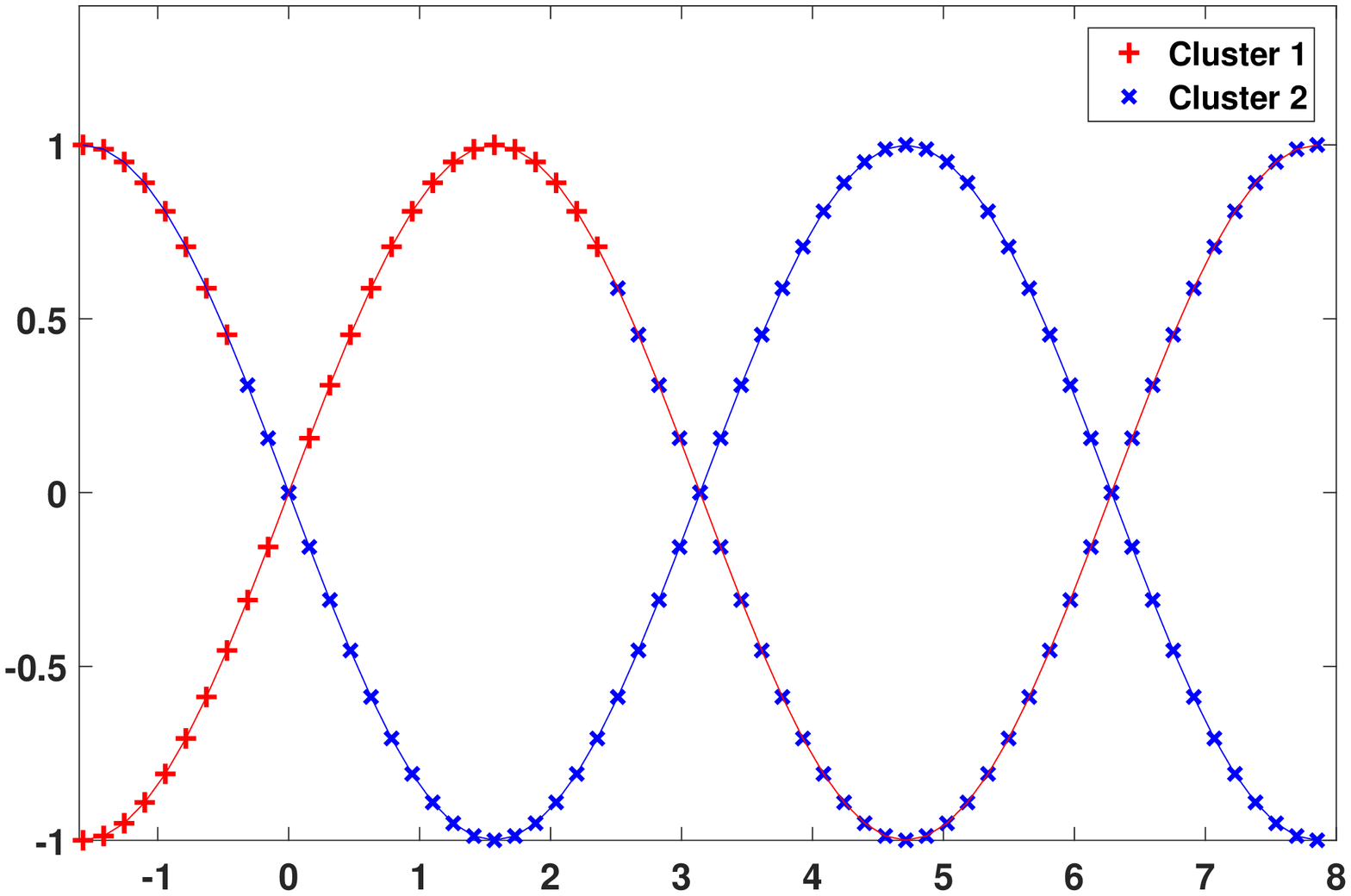}}
\subfigure[$k$PPC]{\includegraphics[width=0.14\textheight]{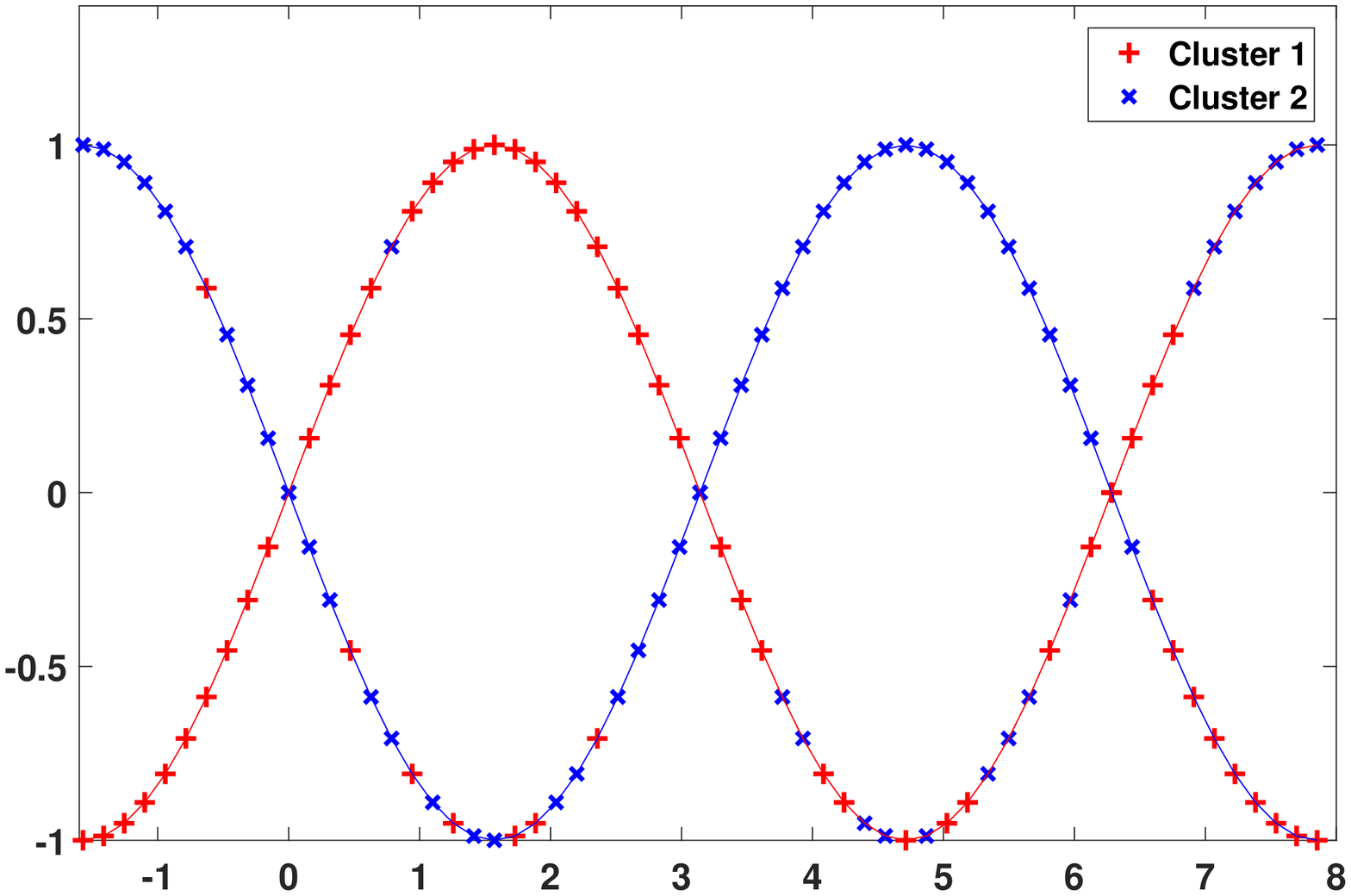}}
\subfigure[L$k$PPC]{\includegraphics[width=0.14\textheight]{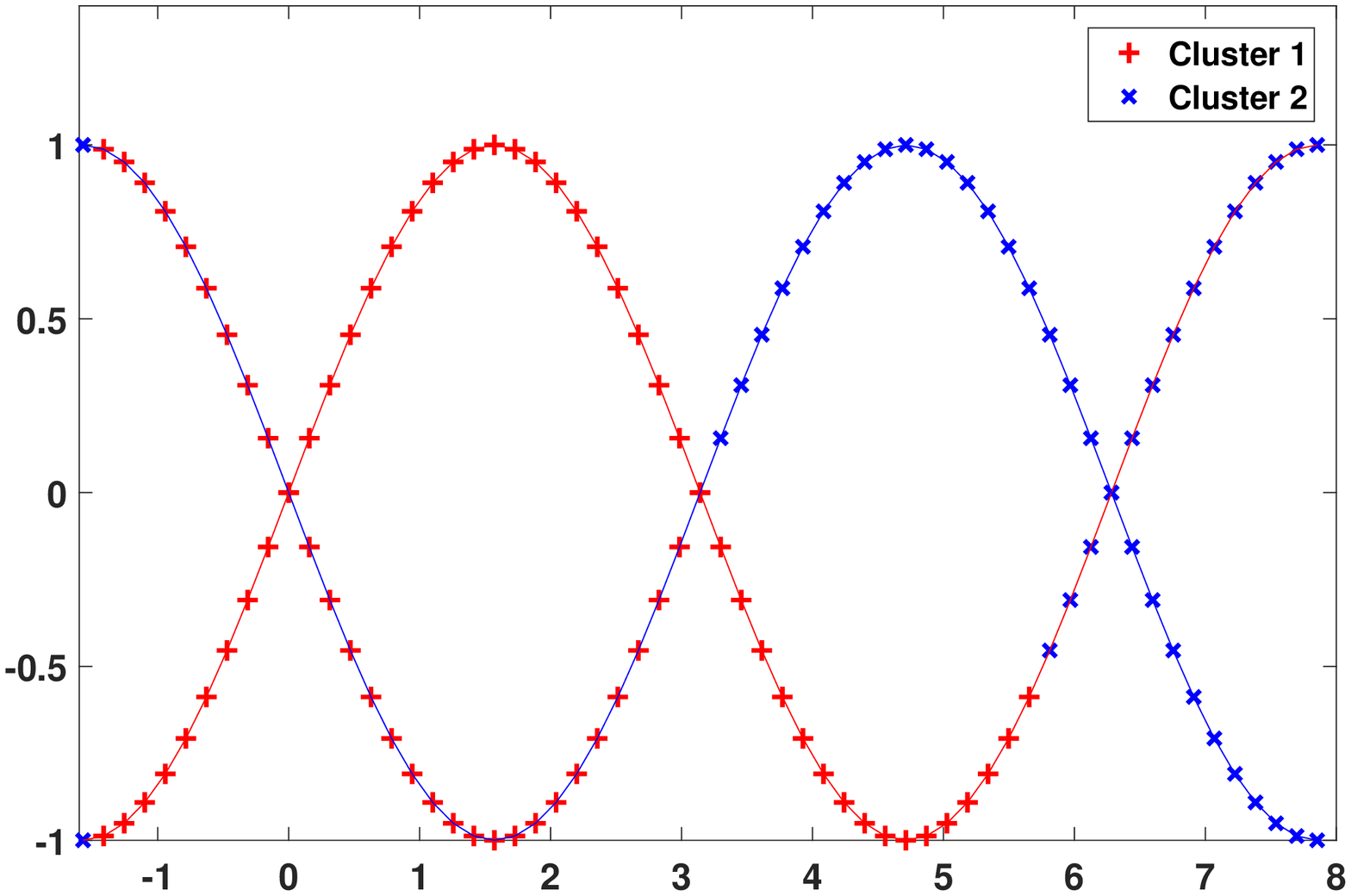}}
\subfigure[TWSVC]{\includegraphics[width=0.14\textheight]{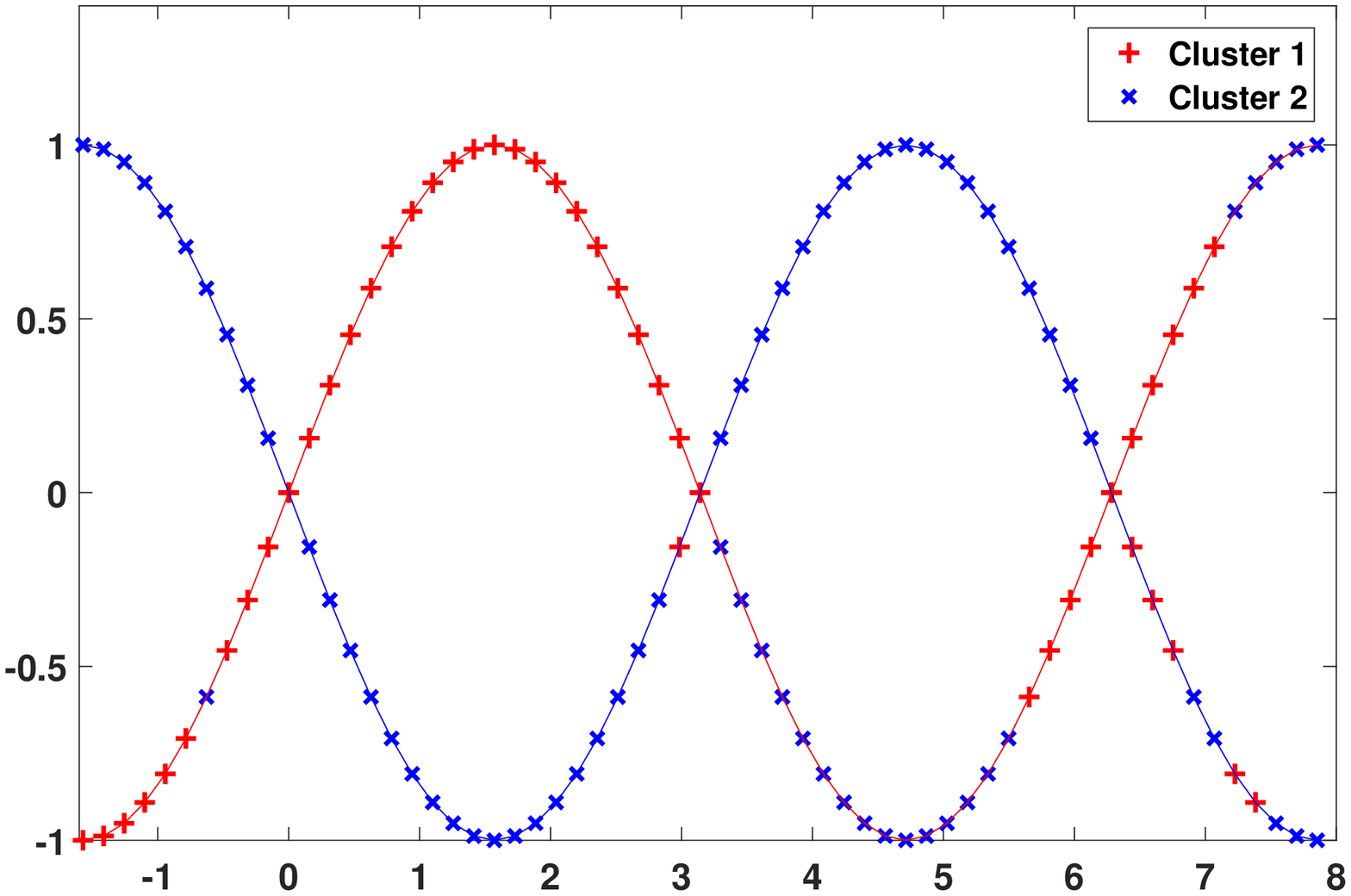}}
\subfigure[$k$FC]{\includegraphics[width=0.14\textheight]{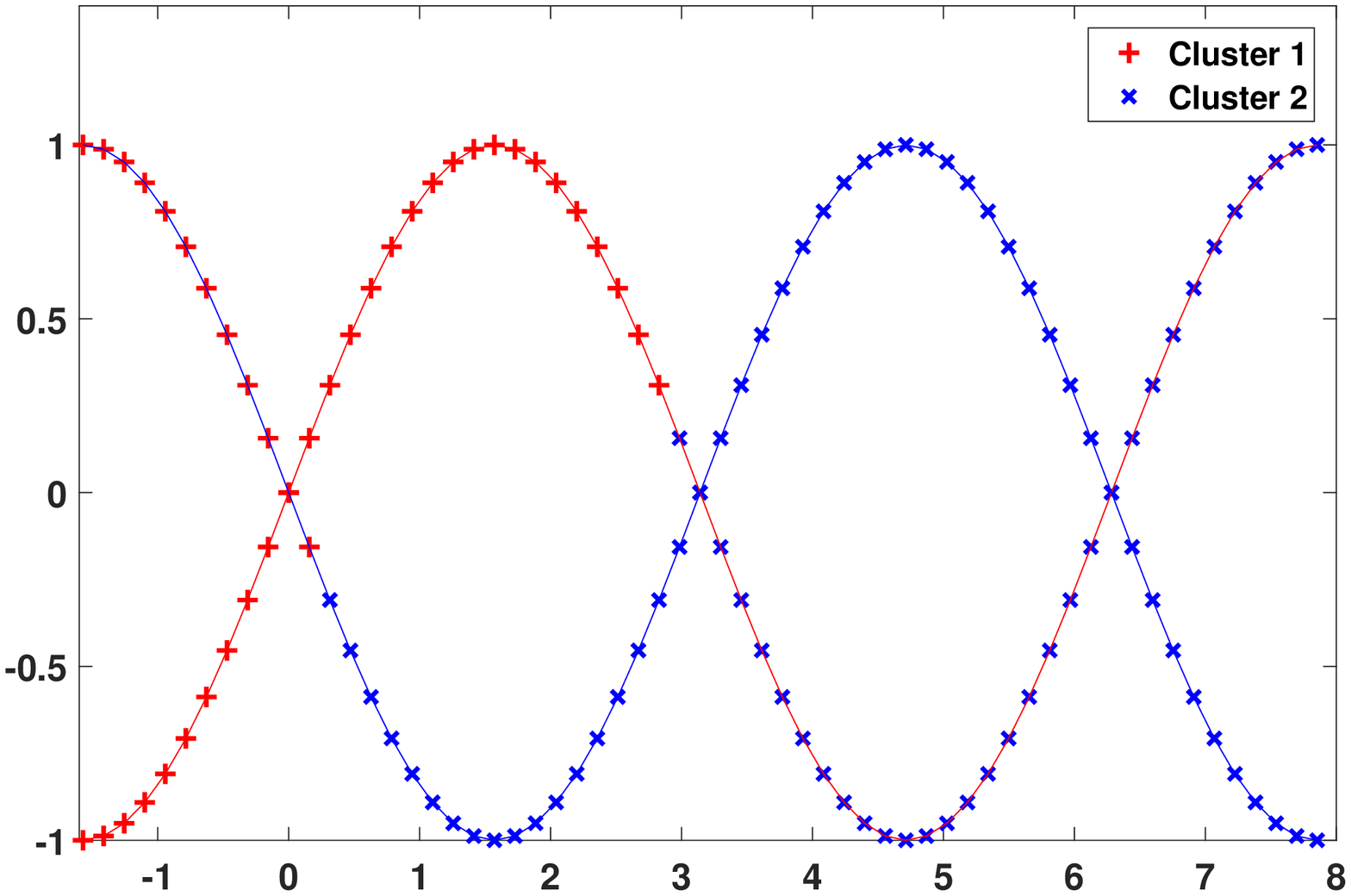}}
\subfigure[L$k$FC]{\includegraphics[width=0.14\textheight]{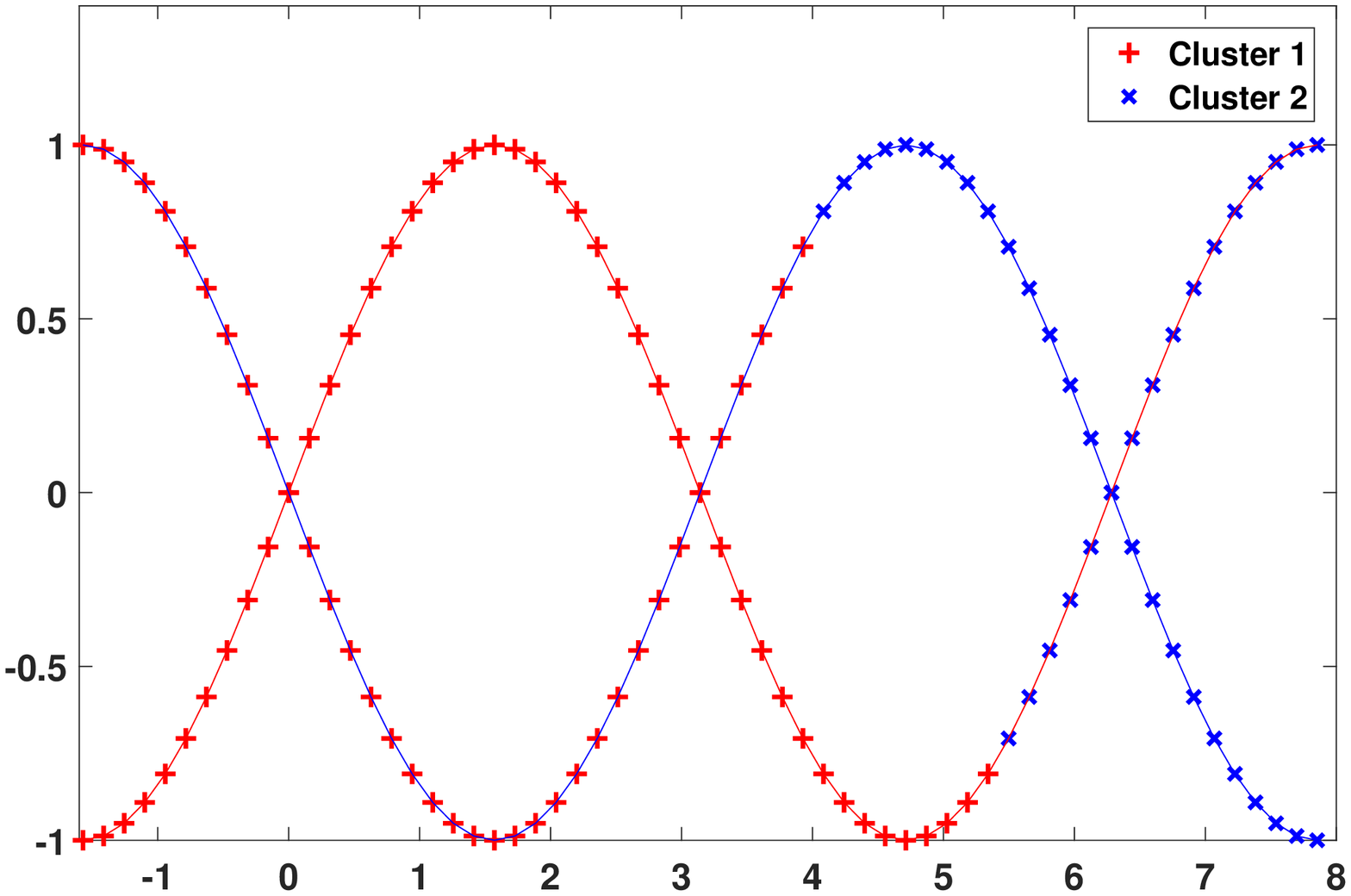}}
\subfigure[MFPC]{\includegraphics[width=0.14\textheight]{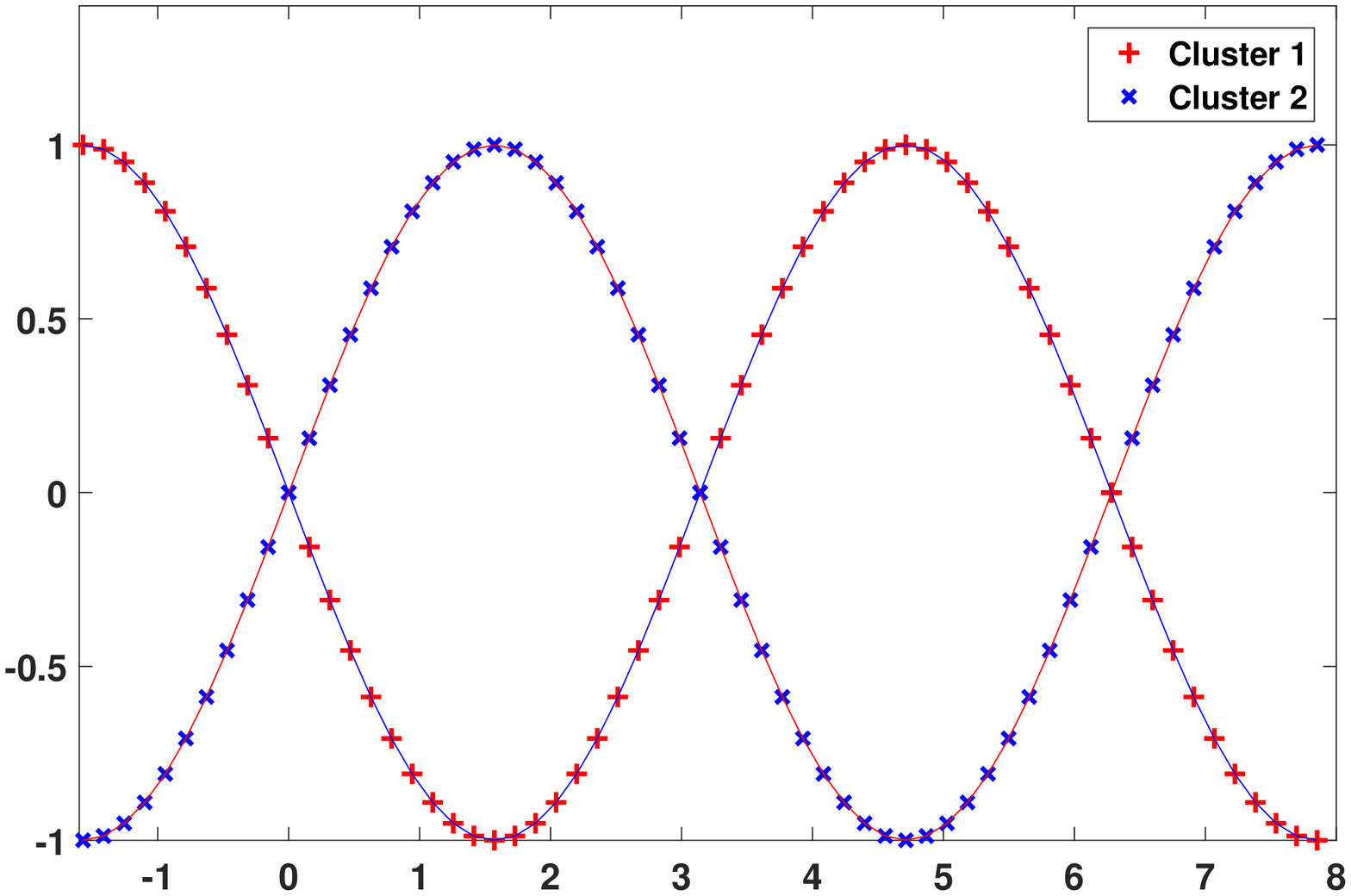}}
\caption{Clustering results of the state-of-the-art methods on the ``Sine2'' dataset which includes two sine curves in $R^2$, where the two curves intersect with each other.}\label{FigSin2}
\end{figure*}

\begin{figure*}[htbp]
\centering
\subfigure[Data]{\includegraphics[width=0.14\textheight]{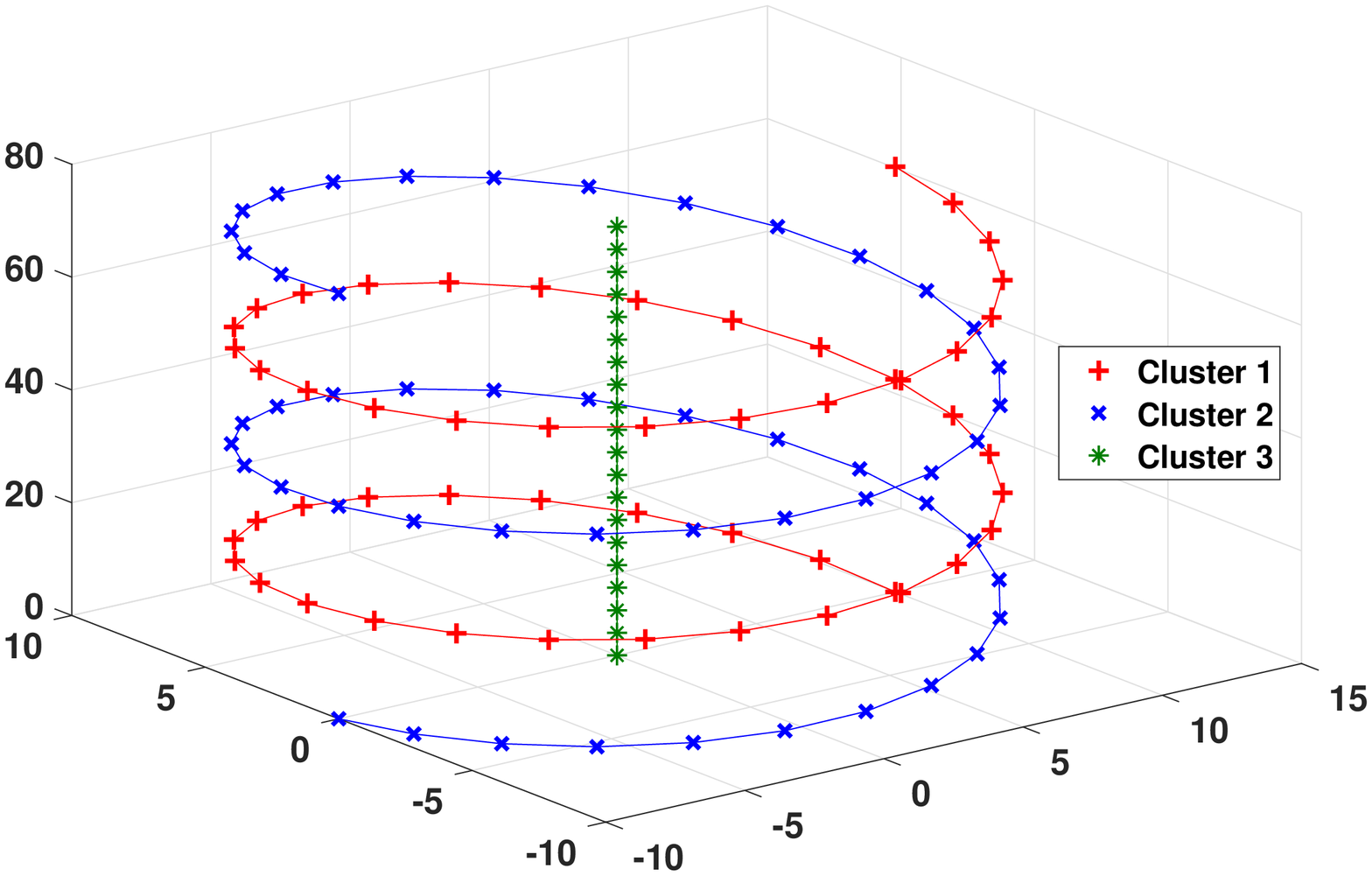}}
\subfigure[$k$means]{\includegraphics[width=0.14\textheight]{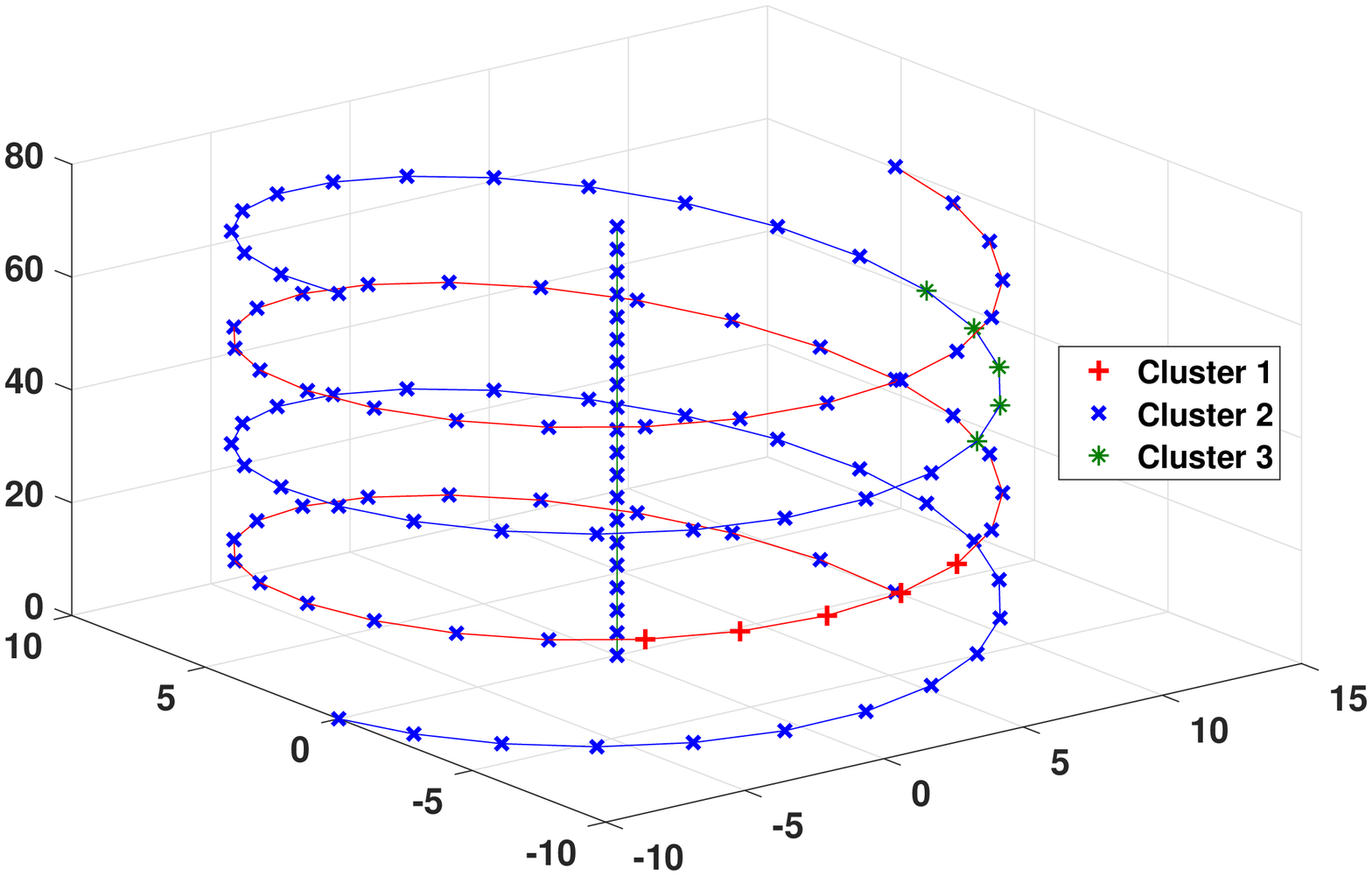}}
\subfigure[SMMC]{\includegraphics[width=0.14\textheight]{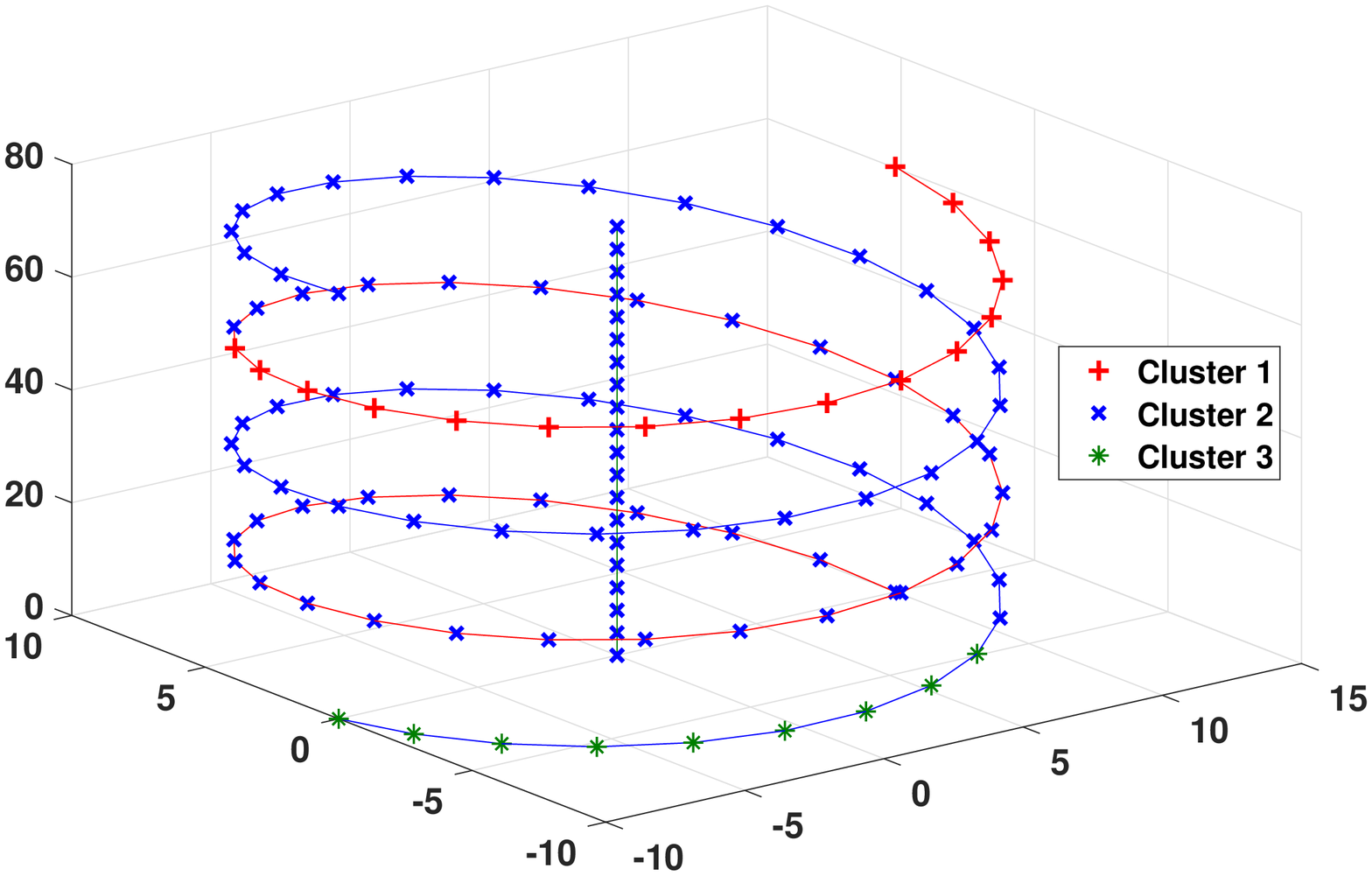}}
\subfigure[$k$PC]{\includegraphics[width=0.14\textheight]{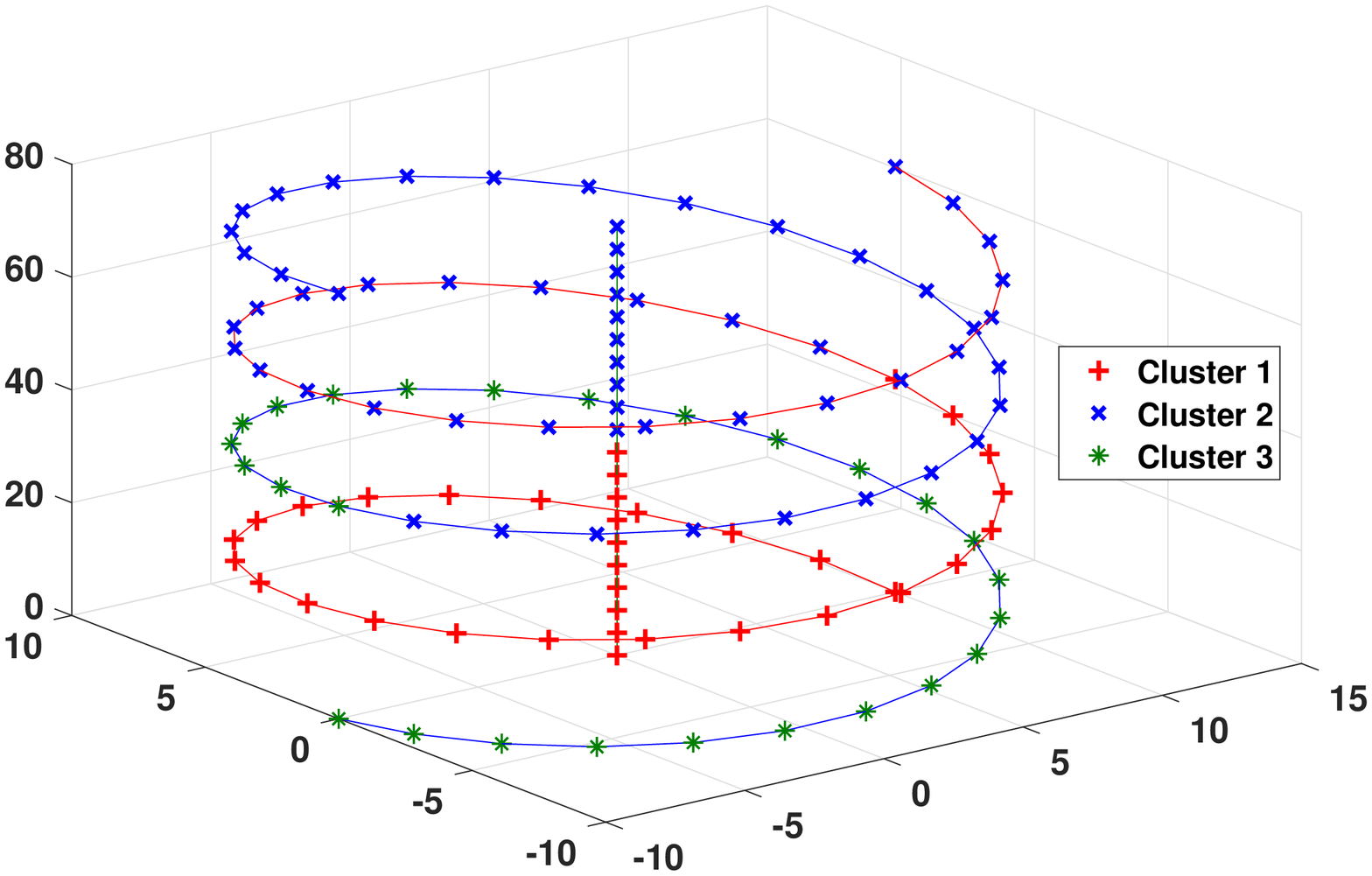}}
\subfigure[$k$PPC]{\includegraphics[width=0.14\textheight]{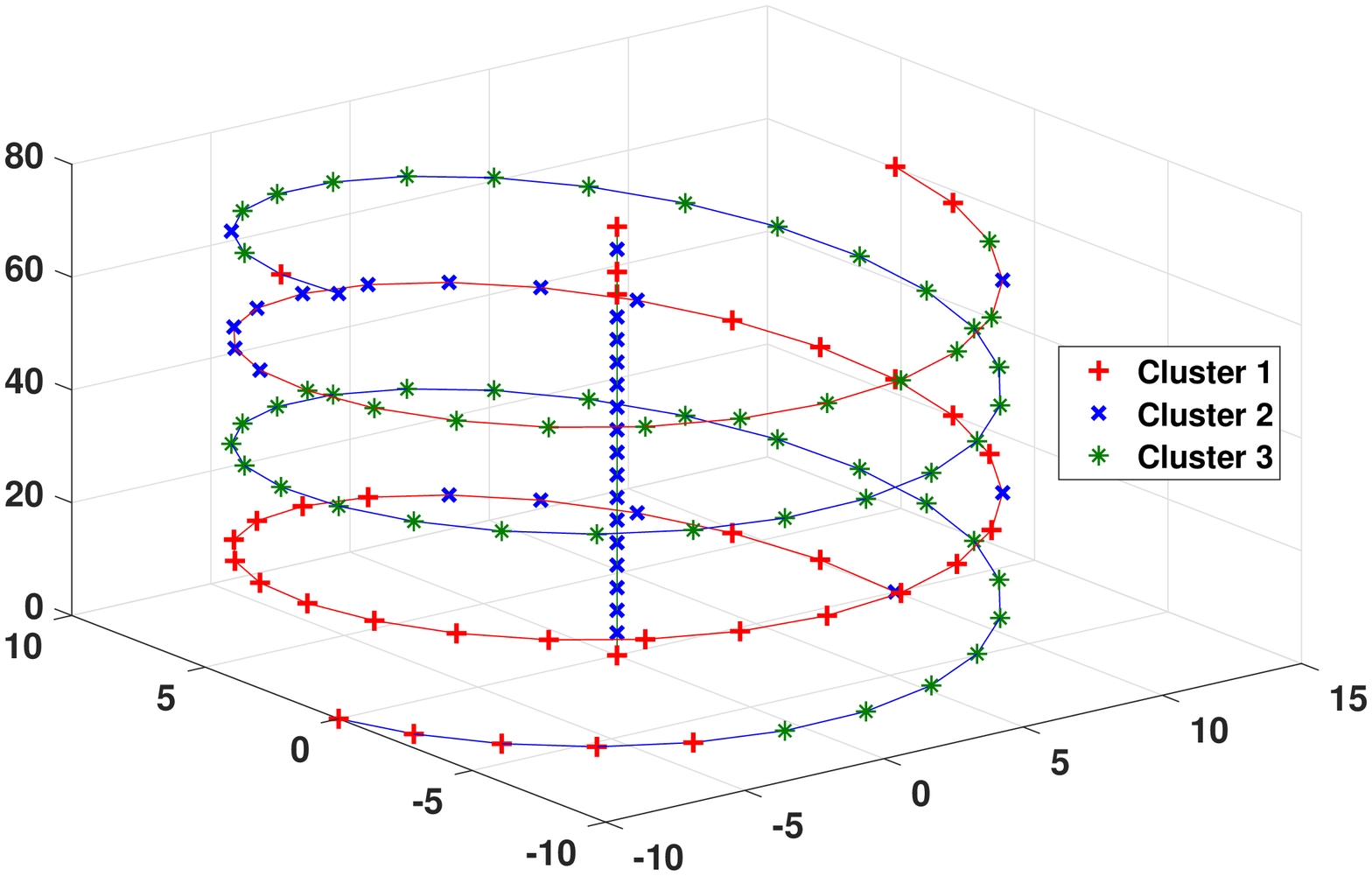}}
\subfigure[L$k$PPC]{\includegraphics[width=0.14\textheight]{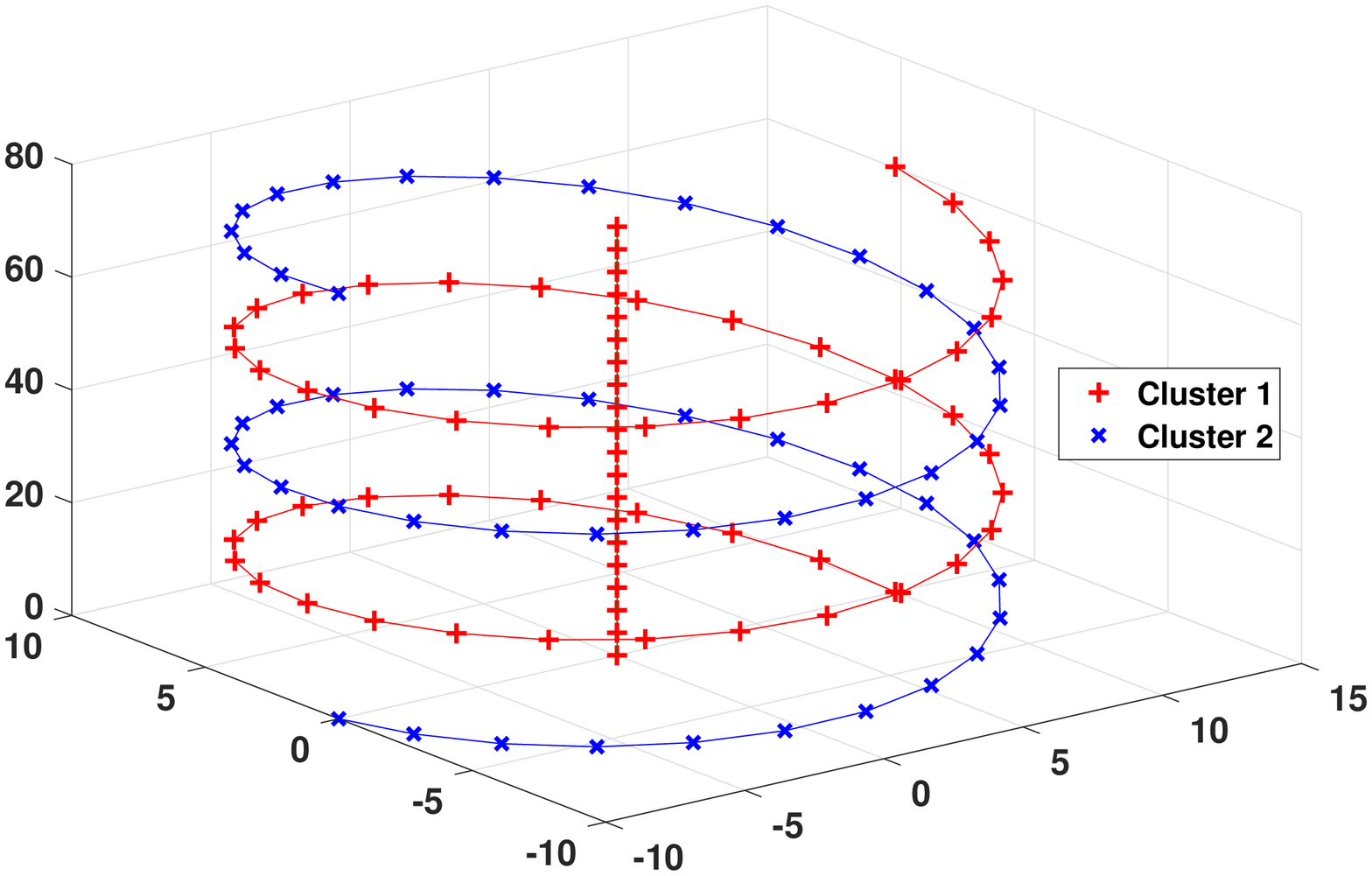}}
\subfigure[TWSVC]{\includegraphics[width=0.14\textheight]{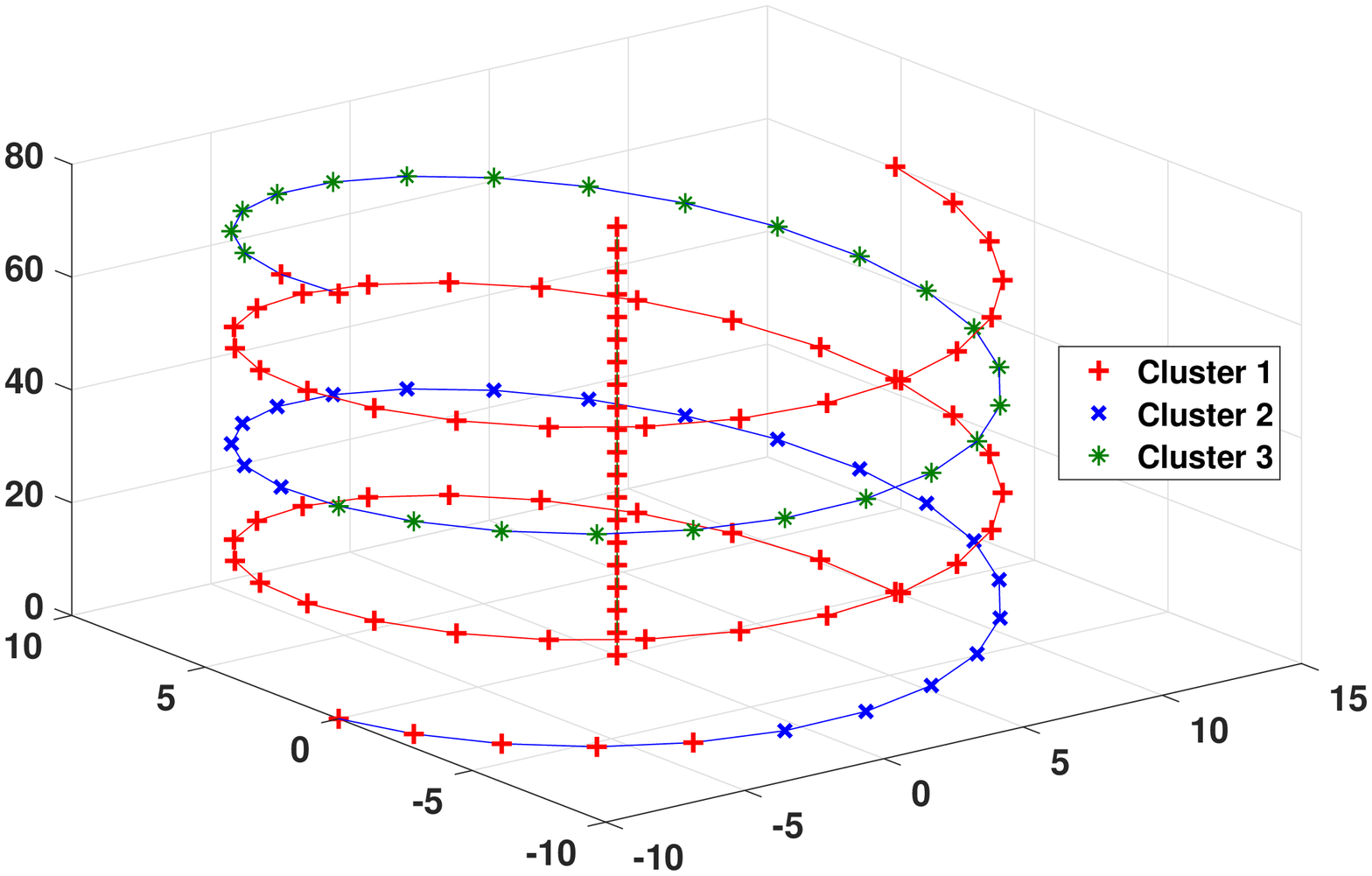}}
\subfigure[$k$FC]{\includegraphics[width=0.14\textheight]{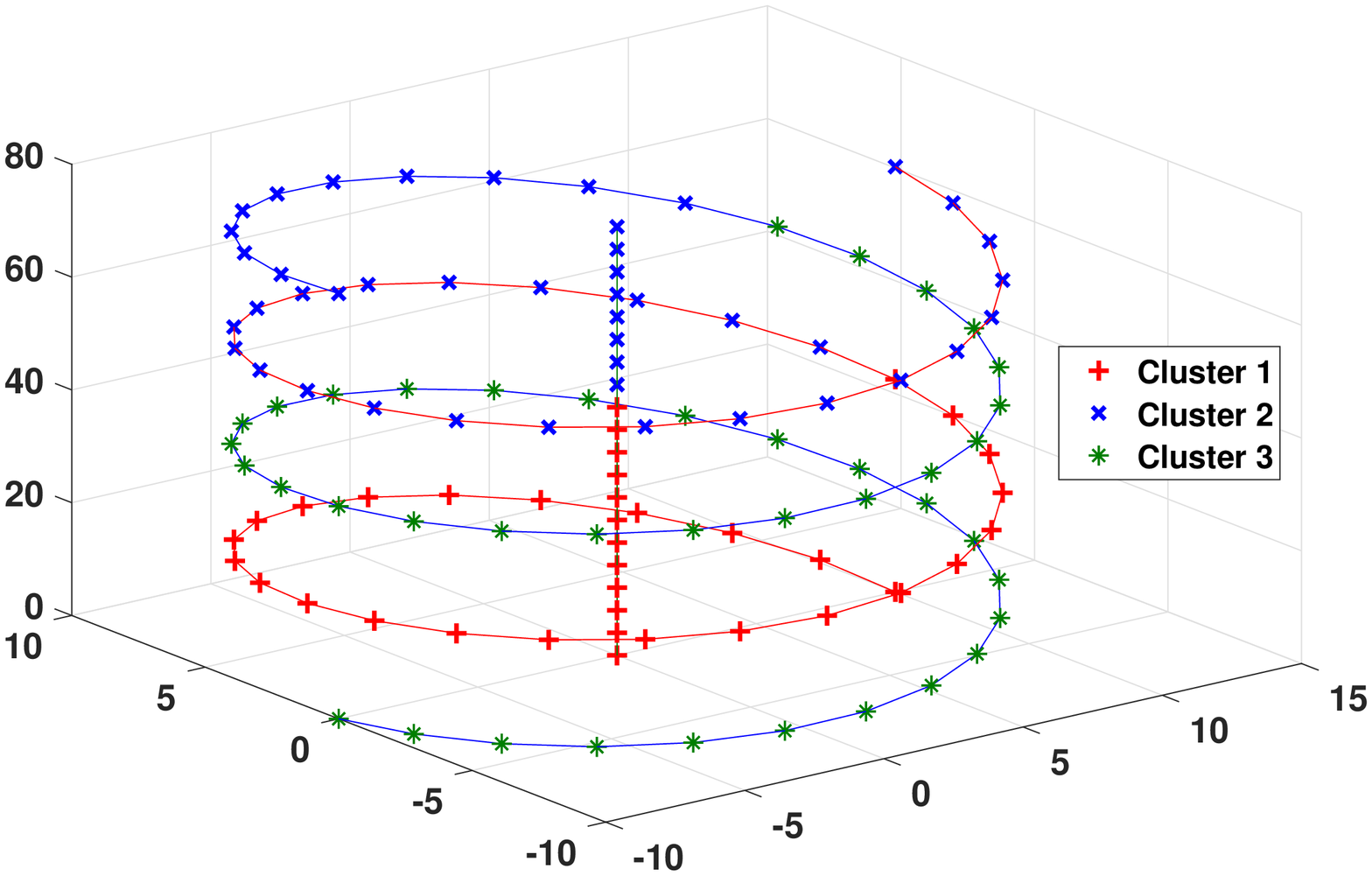}}
\subfigure[L$k$FC]{\includegraphics[width=0.14\textheight]{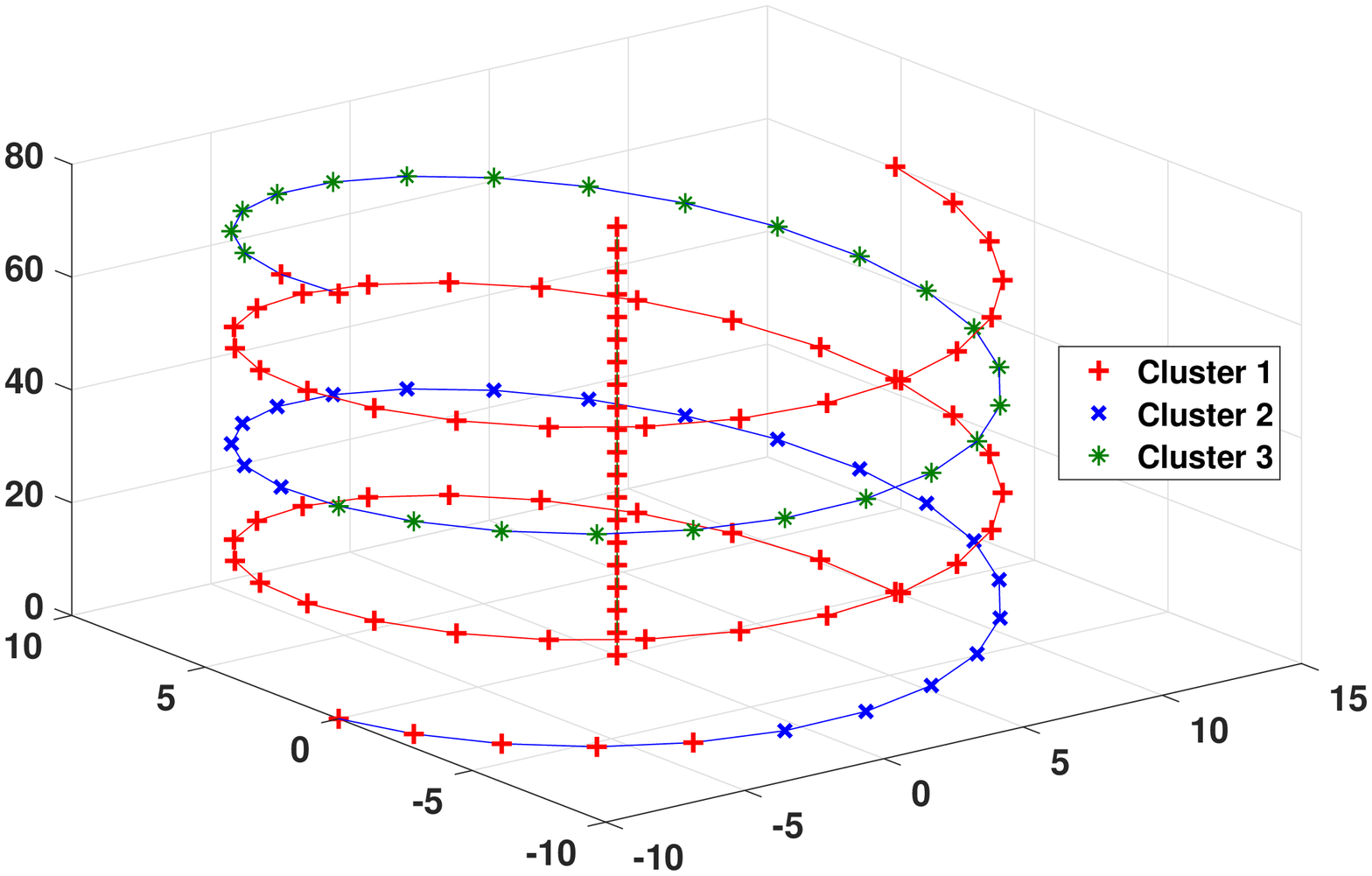}}
\subfigure[MFPC]{\includegraphics[width=0.14\textheight]{DNA.eps}}
\caption{Clustering results of the state-of-the-art methods on the ``Spiral'' dataset which includes two curves and a line without any intersections in $R^3$.}\label{FigDNA}
\end{figure*}

\section{Experimental results}
In this section, we analyze the performance of our MFPC compared with $k$means \cite{Kmeans2}, SMMC \cite{MSMMC}, $k$PC \cite{Kplane}, $k$PPC \cite{PPC}, L$k$PPC \cite{LKPPC}, TWSVC \cite{TWSVC}, $k$FC \cite{Kflat} and L$k$FC \cite{Lkflat} on some synthetic and benchmark datasets. All the methods were implemented by MATLAB2017 on a PC with an Intel Core Duo Processor (double 4.2 GHz) with 16GB RAM.
In the experiments, the adjusted rand index (ARI$\in[-1,1]$) and normalized mutual information (NMI$\in[0,1]$) \cite{ARI,NMI} were hired to measure their performance. The tradeoff parameters if needed in these methods were selected from $\{2^i|i=-8,-7,\ldots,7\}$. For nonlinear case, Gaussian kernel \cite{Kernel2} $K(x_1,x_2)=\exp\{-\mu||x_1-x_2||^2\}$ was used and its parameter $\mu$ was selected from $\{2^i |i = -10,-9,\ldots,5\}$. In our MFPC, if no specific instructions, $p$ (i.e., the number of columns in $W_i$) was selected from $1$ to $\min(n-1,10)$ for linear case, and it was selected from $1$ to $2$ for nonlinear case. For practical convenience, the synthetic datasets and the corresponding MFPC Matlab codes have been uploaded upon \url{http://www.optimal-group.org/Resources/Code/MFPC.html}.

\begin{table*}
\begin{center}
\caption{Clustering performance on four synthetic datasets}\label{ArtTab}
\begin{tabular}{lllllllllll}\\\hline
Data&Criterion&$k$means&SMMC&$k$PC&$k$PPC&L$k$PPC&TWSVC&$k$FC&L$k$FC&MFPC\\\hline
 Haws$^\dagger$&ARI& 0.5104$\pm$0.0755&0.8278$\pm$0.2386& 0.2141  &0.2424  &0.6738  &0.5980  &0.2141  &0.6233&$\mathbf{1.0000}$\\
 323$\times$3&NMI&0.5322$\pm$0.0637&0.8291$\pm$0.2288& 0.2367&0.2803& 0.6529&0.5721&0.2367&0.6285&$\mathbf{1.0000}$\\\hline
 LPE$^\dagger$&ARI&0.5215$\pm$0.0103&0.6237$\pm$0.1360&0.0437&0.2344&0.6228& 0.9800&0.2006&0.6235&$\mathbf{1.0000}$\\
 300$\times$3&NMI&0.5802$\pm$0.0068&0.7100$\pm$0.1167& 0.0560&0.2721&0.6905&0.9660&0.2489&0.6968&$\mathbf{1.0000}$\\\hline
Sine2$^\ddagger$&ARI&  0.0096$\pm$0.0165&0.0150$\pm$0.0349  & 0.0716  &0.1610  & 0.0615 &0.2523  & 0.0708&0.0145&$\mathbf{0.9033  }$\\
122$\times$2&NMI& 0.0528$\pm$0.0659&0.0758$\pm$0.0742 &0.0803&0.1250&0.0549&0.2086&0.0669&0.0184&$\mathbf{0.8355}$\\\hline
Spiral$^\ddagger$& ARI  &0.0560$\pm$0.0932& 0.0397$\pm$0.0495   &0.1908  & 0.3278 & 0.7249 &0.3656&0.3876 &0.3656&$\mathbf{ 1.0000 }$\\
122$\times$3&NMI&0.1002$\pm$0.1100&0.1508$\pm$0.0853&0.3349 &0.3216&0.8140&0.5058&0.4514&0.5058 &$\mathbf{1.0000}$\\\hline
\end{tabular}
\end{center}
~~~~~~~~~~~~$^\dagger$ linear formation; $^\ddagger$ nonlinear formation.
\end{table*}

\subsection{Synthetic datasets}
First, we tested these methods with linear formations on the ``Haws'' dataset which includes three manifolds (two spheres and a line), and the samples distribute uniformly on these manifolds. The clustering results were shown in Fig. \ref{Tanghulu1}. Many methods keep the samples from the spheres and part of the line into a cluster due to the intersections, e.g., $k$means, SMMC, L$k$PPC, TWSVC and L$k$FC. Other methods including $k$PC, $k$PPC and $k$FC separate the spheres into different clusters. However, our MFPC keeps the samples into three clusters from three manifolds exactly. Then, we ran these methods with linear formations on another ``LPE'' dataset which includes a line, a plane and an ellipsoid, where the plane and ellipsoid intersected with the line. Fig. \ref{FigLPE} shows the dataset and the clustering results of these methods. It can be seen from Fig. \ref{FigLPE} that $k$means, kPC, SMMC, L$k$PPC and L$k$FC assign the samples from the line into different clusters. Though $k$PPC and $k$FC assign the samples from the line into one cluster, they assign the samples from other two manifolds into three different clusters. Among these methods, TWSVC and our MFPC can handle this cross-manifold dataset by assigning the samples from different manifolds into different clusters. As shown in Figs. \ref{Tanghulu1} and \ref{FigLPE}, the $k$means, spectral-based SMMC, and other previous flat-type clustering methods cannot handle the linear cross-manifold problem. To further investigate the ability to handle cross-manifold problem, we tested these methods on a nonlinear cross-manifold ``Sine2'' dataset (shown in Fig. \ref{FigSin2}), where the samples were from two sine functions and they intersected with each other. These methods were implemented in $16$ high dimensional feature spaces generated by Gaussian kernel, and the best results by each method were selected and reported in Fig. \ref{FigSin2}. It is obvious that our MFPC assign the samples from differen sine curves into different clusters exactly, while other methods mix the samples from the two curves in a cluster. Thus, these methods cannot handle this nonlinear cross-manifold problem except our MFPC. The above tests illustrate the ability of our MFPC to handle some cross-manifold problems. In the following, we tested these methods on a complicate dataset ``Spiral'' without any intersections, which includes three manifolds: two curves and a line in $\mathbb{R}^3$. Fig. \ref{FigDNA} illustrate the dataset and the clustering results by these methods. It can be seen that our MFPC surpasses other methods on this dataset much more. Further, the clustering performance on the four synthetic datasets ``Haws'', ``LPE'', ``Sine2'' and ``Spiral'' was measured by ARI and NMI in Table \ref{ArtTab}. Thereinto, $k$means and SMMC were implemented repeatedly $20$ times and the average measurements and the standard deviations were reported, while other methods obtain stable performance with the nearest neighbor graph (NNG) \cite{TWSVC} initialization. Obviously, our MFPC outperforms other methods by both ARI and NMI from Table \ref{ArtTab}.

\begin{table*}
\centering
\caption{Performance of the state-of-the-art clustering methods with linear formations on benchmark datasets}
\begin{tabular}{llccccccccc}\\\hline
Data&Criterion&$k$means&SMMC&$k$PC&$k$PPC&L$k$PPC&TWSVC&$k$FC&L$k$FC&MFPC\\\hline
Australian&ARI&0.0033$\pm$0.0007  &0.0038$\pm$0.0000 &-0.0032 &0.0000 &0.0022 &0.0090 &0.0424 &0.0022&$\mathbf{0.2275}$\\
690$\times$14&NMI& 0.0317$\pm$0.0043 &0.0344$\pm$0.0000 &0.0032 &0.0143 &0.0255 &0.0298 &0.0272 &0.0255 &$\mathbf{0.2404}$\\\hline
Car&ARI&  0.0839$\pm$0.0620&0.0348$\pm$0.0636 &0.0429 &0.1377 &0.1684 &0.0765 &0.0997 &0.2000 &$\mathbf{0.2283}$\\
1728$\times$6&NMI& 0.1663$\pm$0.0805 &0.1103$\pm$0.1053 &0.0892 &0.1951 &$\mathbf{0.3173}$ &0.1876 &0.1483 &0.2831 &0.2964\\\hline
Dna&ARI& 0.2756$\pm$0.3066 &0.5128$\pm$0.3117 &0.4889 &0.3868 &$\mathbf{0.9386}$ &0.4889 &0.5296 &$\mathbf{0.9386}$ &$\mathbf{0.9386}$\\
2000$\times$180&NMI& 0.3673$\pm$0.3061 &0.5584$\pm$.2869 &0.5872 &0.4285 &$\mathbf{0.9171}$ &0.5872 &0.7024 &$\mathbf{0.9171}$ &$\mathbf{0.9171}$\\\hline
Echocardiogram&ARI&0.3797$\pm$0.1340  &0.5216$\pm$0.0378 &0.0250 &0.0884 &0.4780 &0.0159 &0.4557 &0.4571 &$\mathbf{0.7355}$\\
131$\times$10&NMI& 0.3298$\pm$0.1197 & 0.4875$\pm$0.0371& 0.0058&0.0375 & 0.4131&0.0941 & 0.3968&0.3992 &$\mathbf{0.6714}$\\\hline
Ecoli&ARI& 0.4130$\pm$0.0384 &0.0000$\pm$0.0000 &0.0341 &0.0390 &0.6823 &0.6422 &0.4121 &0.6986 &$\mathbf{0.7288}$\\
336$\times$7&NMI& 0.5975$\pm$0.0245 &0.0000$\pm$0.0000 &0.1620 &0.2178 &0.6691 &0.5850 &0.5207 &0.6586 &$\mathbf{0.6804}$\\\hline
Glass&ARI& 0.2600$\pm$0.0217 &0.1767$\pm$0.0382 &0.2223 &0.0570 &0.2953 &0.2257 &$\mathbf{0.3056}$ &0.2446 &0.2993\\
214$\times$9&NMI& 0.4157$\pm$0.0377 &0.3234$\pm$0.0446 &0.3028 &0.1046 &0.4782 &0.3392 &0.4763 &0.4333 &$\mathbf{0.4797}$\\\hline
Hepatitis&ARI& 0.0254$\pm$0.0107 &-0.0015$\pm$0.0000 & -0.0519&$\mathbf{0.0532}$ &0.0198 &0.0159 & 0.0520& 0.0159&0.0496\\
155$\times$19&NMI& 0.0037$\pm$0.0012 &0.0000$\pm$0.0000 & 0.0103&0.0090 &0.0039 & 0.0039& 0.0081&0.0039 &$\mathbf{0.0728}$\\\hline
Housevotes&ARI& 0.5751$\pm$0.0036 & 0.5920$\pm$0.0000& 0.2738& 0.3680& 0.6208& 0.5167& 0.4521& 0.5779&$\mathbf{0.8323}$\\
435$\times$16&NMI& 0.4867$\pm$0.0048 &0.5055$\pm$0.0000 & 0.3422& 0.2949& 0.5558& 0.4552& 0.4257& 0.4905&$\mathbf{0.7415}$\\\hline
Ionosphere&ARI& 0.1584$\pm$0.0541 & 0.3430$\pm$0.0035& 0.2204& 0.0611& 0.1871& 0.0056& $\mathbf{0.4188}$& 0.2092&0.1873\\
351$\times$33&NMI& 0.1229$\pm$0.0341 & 0.2757$\pm$0.0041&0.1400 & 0.0330&0.1349 &0.0278 &$\mathbf{0.3144}$ & 0.2602&0.1308\\\hline
Iris&ARI& 0.7247$\pm$0.0072 &0.7172$\pm$0.0917 &0.2666 &0.1229 &0.9037 &0.8032 &0.8176 &0.7445 &$\mathbf{0.9603}$\\
150$\times$4&NMI& 0.7517$\pm$0.0084 &0.7688$\pm$0.0391 &0.2547 &0.1321 &0.8801 &0.8315 &0.8027 &0.7777 &$\mathbf{0.9488}$\\\hline
Pathbased&ARI& 0.4628$\pm$0.0013 &0.4342$\pm$0.0018 &0.2458 &0.4582 &$\mathbf{0.4825}$ &0.4419 &0.2458 &0.1890 &0.4648\\
300$\times$2&NMI& 0.5482$\pm$0.0009 &0.5248$\pm$0.0017 &0.3018 &0.5445 &$\mathbf{0.5588}$ &0.5091 &0.3018 &0.2312 &0.5429\\\hline
Seeds&ARI& 0.7146$\pm$0.0039 &0.6264$\pm$0.0000 &0.4315 &0.2084 &0.7566 &0.3029 &0.4410 &0.7166 &$\mathbf{0.8889}$\\
210$\times$7&NMI&0.7033$\pm$0.0091  &0.6411$\pm$0.0000 &0.5169 &0.2006 &0.7243 &0.4256 &0.5297 &0.6949 &$\mathbf{0.8486}$\\\hline
Sonar&ARI& 0.0065$\pm$0.0047 &0.0042$\pm$0.0030 &-0.0040 &-0.0003 &0.0287 &0.0087 &0.0287 &0.0190 &$\mathbf{0.0580}$\\
208$\times$60&NMI& 0.0091$\pm$0.0035 &0.0065$\pm$0.0015 &0.0001 &0.0039 &0.0655 &0.0078 &0.0219 &0.0156 &$\mathbf{0.0912}$\\\hline
Soybean&ARI& 0.9367$\pm$0.2001 &0.5207$\pm$0.3308 &0.8335 &$\mathbf{1.0000}$ &$\mathbf{1.0000}$ &$\mathbf{1.0000}$ & $\mathbf{1.0000}$&$\mathbf{1.0000}$ &$\mathbf{1.0000}$\\
47$\times$35&NMI& 0.9413$\pm$0.1858 &0.5623$\pm$0.3020 &0.7857 &$\mathbf{1.0000}$ &$\mathbf{1.0000}$ &$\mathbf{1.0000}$ &$\mathbf{1.0000}$ &$\mathbf{1.0000}$ &$\mathbf{1.0000}$\\\hline
Spect&ARI& -0.1067$\pm$0.0000 &-0.1067$\pm$0.0000 &-0.0159 &0.0107 &0.0000 &-0.0159 &0.0000 &0.0000 &$\mathbf{0.0818}$\\
267$\times$44&NMI& 0.0898$\pm$0.0000 &0.0885$\pm$0.0010 &0.0147 &0.0104 &$\mathbf{0.1218}$ &0.0147 &0.0329 &0.0898 &0.0797\\\hline
Wine&ARI& 0.3634$\pm$0.0100 &0.3961$\pm$0.0016 &0.0387 &0.0446 &0.4330 &0.3505 &0.3474 &0.3694 &$\mathbf{0.5511}$\\
178$\times$13&NMI& 0.4269$\pm$0.0024 &0.3943$\pm$0.0002 &0.0838 &0.0523 &0.4772 &0.4958 &0.4357 &0.4429 &$\mathbf{0.6762}$\\\hline
Zoo&ARI& 0.6340$\pm$0.0775 &0.5669$\pm$0.0840 &0.2209 &0.5177 &0.7001 &0.6682 &0.7076 &0.8382 &$\mathbf{0.9388}$\\
101$\times$16&NMI& 0.7385$\pm$0.0339 &0.7340$\pm$0.0414 &0.5005 &0.5742 &0.7887 &0.7460 &0.8061 &0.8273 &$\mathbf{0.8937}$\\\hline
Rank&ARI&4.94  &5.29 &6.94 &6.11 &2.88 & 5.35&3.71 &3.71 &$\mathbf{1.47}$\\
&NMI& 4.76 &5.18 &6.82 &6.71 &2.53 &5.12 &4.35 &3.71 &$\mathbf{1.71}$\\\hline
\end{tabular}\label{LinearUCI}
\end{table*}

\begin{table*}
\centering
\caption{Performance of the state-of-the-art clustering methods with nonlinear formations on benchmark datasets}
\begin{tabular}{llccccccccc}\\\hline
Data&Criterion&$k$means&SMMC&$k$PC&$k$PPC&L$k$PPC&TWSVC&$k$FC&L$k$FC&MFPC\\\hline
Australian&ARI& 0.0003$\pm$0.0006 &0.0001$\pm$0.0001 &0.0068 &0.0329 &0.0327 &-0.0011 &0.0220 &0.0023 &$\mathbf{0.1146}$\\
690$\times$14&NMI& 0.0275$\pm$0.0148 &0.0004$\pm$0.0002 &0.0146 &0.0301 &0.0620 &0.0372 &0.0612 &0.0433 &$\mathbf{0.0850}$\\\hline
Car&ARI& 0.1900$\pm$0.0574 &0.2005$\pm$.0850 &0.1361 &0.1944 &0.2895 &0.2146 &0.2267 &0.2704 &$\mathbf{0.3062}$\\
1728$\times$6&NMI& 0.2517$\pm$0.0701 &0.2764$\pm$0.0770 &0.1977 &0.3364 &0.3745 &0.3556 &0.3745 & 0.3556 &$\mathbf{0.3956}$\\\hline
Dna&ARI& 0.2860$\pm$0.0599 &0.2530$\pm$0.0211 &0.5661 &0.4820 &0.5661 &0.5661 &0.5661 &0.5661 &$\mathbf{0.9695}$\\
2000$\times$180&NMI& 0.3311$\pm$.0595 &0.2597$\pm$.0271 &0.6361 &0.5314 &0.7024 &0.6361 &0.6600 &0.7024 &$\mathbf{0.9587}$\\\hline
Echocardiogram&ARI&0.4271$\pm$0.0136  &0.4118$\pm$0.0693 &0.0376 &0.1130 &0.4553 &0.0322 &0.1974 &0.4166 &$\mathbf{0.4855}$\\
131$\times$10&NMI& 0.3437$\pm$0.0101 & 0.3481$\pm$0.0428&0.1086 &0.0771 &0.3608 &0.1086 & 0.2443& $\mathbf{0.3731}$&0.3700\\\hline
Ecoli&ARI& 0.4271$\pm$0.0809 &- &0.5648 &0.1020 &0.7103 &0.7132 &0.6753 &0.7279 &$\mathbf{0.7527}$\\
336$\times$7&NMI& 0.5703$\pm$0.0192 &- &0.6182 &0.1902 &0.6729 &0.6785 &0.6428 &0.6812 &$\mathbf{0.7127}$\\\hline
Glass&ARI& 0.2572$\pm$0.0204 &- &0.2672 &0.0724 &0.2634 &0.2695 &0.2962 &0.2503 &$\mathbf{0.3047}$\\
214$\times$9&NMI& 0.4037$\pm$0.0491 &- &0.4349 &0.1043 &0.4378 &0.4592 &$\mathbf{0.4896}$ &0.4322 &0.4890\\\hline
Hepatitis&ARI& 0.0050$\pm$0.0229 & -0.0272$\pm$0.0000& 0.0872& 0.1361& 0.0718& 0.0362& 0.0000&0.0000 &$\mathbf{0.1671}$\\
155$\times$19&NMI& 0.0297$\pm$0.0305 &0.0101$\pm$0.0000 & 0.0213& 0.0728&$\mathbf{0.1104}$ &0.0728 &0.0382 & 0.0317&$\mathbf{0.1104}$\\\hline
Housevotes&ARI& 0.6014$\pm$0.0174 & 0.0012$\pm$0.0003& 0.5101& 0.5167& 0.5778& 0.8238& 0.6501&0.5778 &$\mathbf{0.8323}$\\
435$\times$16&NMI& 0.4816$\pm$0.0162 & 0.0054$\pm$0.0045& 0.4682&0.4728 &0.4794 & 0.7263&0.5602 & 0.4794&$\mathbf{0.7415}$\\\hline
Ionosphere&ARI& 0.2465$\pm$0.0000 & -0.0359$\pm$0.0000&0.1802 &0.1879 & 0.2890& 0.1802& 0.4087&0.2465 &$\mathbf{0.6844}$\\
351$\times$33&NMI& 0.2668$\pm$0.0000 & 0.0719$\pm$0.0000& 0.2412& 0.1866& 0.2922& 0.2412&0.3281 & 0.2668&$\mathbf{0.5728}$\\\hline
Iris&ARI& 0.7747$\pm$0.0373 &0.7734$\pm$0.0000 &0.8017 &0.0389 &0.8178 &0.8017 &0.9222 &0.8178 &$\mathbf{0.9410}$\\
150$\times$4&NMI& 0.8139$\pm$0.0000 &0.8139$\pm$0.0000 &0.7919 &0.0817 &0.8139 &0.7919 &0.9144 &0.8139 &$\mathbf{0.9192}$\\\hline
Pathbased&ARI& 0.9143$\pm$0.0049 &0.5548$\pm$0.1640 &0.5099 &0.0982 &0.9105 &0.5897 &0.9294 &0.9105 &$\mathbf{0.9798}$\\
300$\times$2&NMI&0.8847$\pm$0.0049  &0.6445$\pm$0.1284 &0.6298 &0.1160 &0.8809 &0.7036 &0.9045 &0.8809 &$\mathbf{0.9659}$\\\hline
Seeds&ARI& 0.7111$\pm$0.0168 &- &0.5223 &0.2899 &0.7400 &0.5879 &0.7329 &0.7005 &$\mathbf{0.7423}$\\
210$\times$7&NMI& 0.6954$\pm$0.0062 &- &0.6012 &0.2853 &0.7101 &0.6427 &0.7094 &0.6944 &$\mathbf{0.7158}$\\\hline
Sonar&ARI& 0.0076$\pm$0.0065 &0.0046$\pm$0.0000 &0.0324 &0.0532 &0.0444 &0.0445 &$\mathbf{0.0963}$ &0.0088 &0.0680\\
208$\times$60&NMI& 0.0516$\pm$0.0220 &0.0305$\pm$0.0000 &0.0679 &0.0679 &0.1181 &0.0755 &0.0800 &0.0679 &$\mathbf{0.1312}$\\\hline
Soybean&ARI& $\mathbf{1.0000}$$\pm$0.0000 &0.9149$\pm$0.0000 &$\mathbf{1.0000}$ &$\mathbf{1.0000}$ &$\mathbf{1.0000}$ &$\mathbf{1.0000}$ &$\mathbf{1.0000}$ &$\mathbf{1.0000}$ &$\mathbf{1.0000}$\\
47$\times$35&NMI& $\mathbf{1.0000}$$\pm$0.0000 &0.8711$\pm$0.0000 &$\mathbf{1.0000}$ &$\mathbf{1.0000}$ &$\mathbf{1.0000}$ &$\mathbf{1.0000}$ &$\mathbf{1.0000}$ &$\mathbf{1.0000}$ &$\mathbf{1.0000}$\\\hline
Spect&ARI& 0.2897$\pm$0.0128 &0.2879$\pm$0.0157 &0.1515 &0.1891 &0.2965 &0.1515 &0.1787 &0.2873 &$\mathbf{0.3316}$\\
267$\times$44&NMI& 0.1704$\pm$0.0000 &0.1704$\pm$0.0000 &0.1182 &0.1409 &0.1789 &0.0871 &0.1095 &0.1827 &$\mathbf{0.2411}$\\\hline
Wine&ARI&0.0665$\pm$0.0212  &- & 0.2144& 0.2261& 0.3797& 0.0361& 0.3179&0.0496 &$\mathbf{0.4083}$\\
178$\times$13&NMI& 0.1571$\pm$0.0246 &- &0.2466 &0.2765 &$\mathbf{0.4586}$ &0.0965 &0.3384 &0.1323 &0.4175\\\hline
Zoo&ARI&0.6481$\pm$0.0319  &0.4854$\pm$0.1507 &0.7130 &0.6841 &0.6951 &0.7130 &0.7130 &0.8013 &$\mathbf{0.8087}$\\
101$\times$16&NMI& 0.7377$\pm$0.0197 &0.6876$\pm$0.0000 &0.8166 &0.7502 &0.8120 &0.8166 &0.8166 &0.8166 &$\mathbf{0.8218}$\\\hline
Rank&ARI& 5.00 &6.88 &5.35 &5.29 &3.12 &4.88 &3.35 &4.35 &$\mathbf{1.06}$\\
&NMI&4.88  &6.53 &5.41 &5.47 &2.65 &4.24 &3.00 &3.53 &$\mathbf{1.18}$\\\hline
\end{tabular}\label{NonlinearUCI}
\\`-' throw errors from the probabilistic principal components analysis step in SMMC.
\end{table*}

During the above synthetic tests, it can be found that $k$means always assigns the samples close to each other into a cluster, because it hires points as the cluster centers. Thus, $k$means cannot handle more general cluster centers, e.g., lines and planes. The flat-type methods settle this issue by extending the cluster center from points to different flats. However, many flat-type methods are disordered by the cross-manifold structures from Figs. \ref{Tanghulu1}, \ref{FigLPE}, \ref{FigSin2} and \ref{FigDNA}. It is worth to notice that some flat-type methods may assign the samples from one manifold into a cluster on some cross-manifold datasets, e.g., L$k$PPC captures a sphere in Fig. \ref{Tanghulu1}, TWSVC works well in Fig. \ref{FigLPE}, and L$k$FC captures a plane in Fig. \ref{FigLPE}. This phenomenon indicates that the flat-type methods has the capacity to deal with linear cross-manifold clustering. In fact, our MFPC works well on the two linear cross-manifold datasets. Moreover, Fig. \ref{FigSin2} manifest the ability of MFPC to handle more complicated cross-manifold structures. Finally, MFPC keeps on top of the general manifold clustering from Fig. \ref{FigDNA}. In addition, we observe from Table \ref{ArtTab} that SMMC performs much better on ``Haws'' than other datasets, which indicates its limited adaptiveness. Besides, SMMC works unstably due to its large standard deviations in Table \ref{ArtTab}. In conclusion, our MFPC outperforms other methods with stable performance in the synthetic experiments.

\begin{figure*}[htbp]
\centering
\subfigure[Australian]{\includegraphics[width=0.18\textheight]{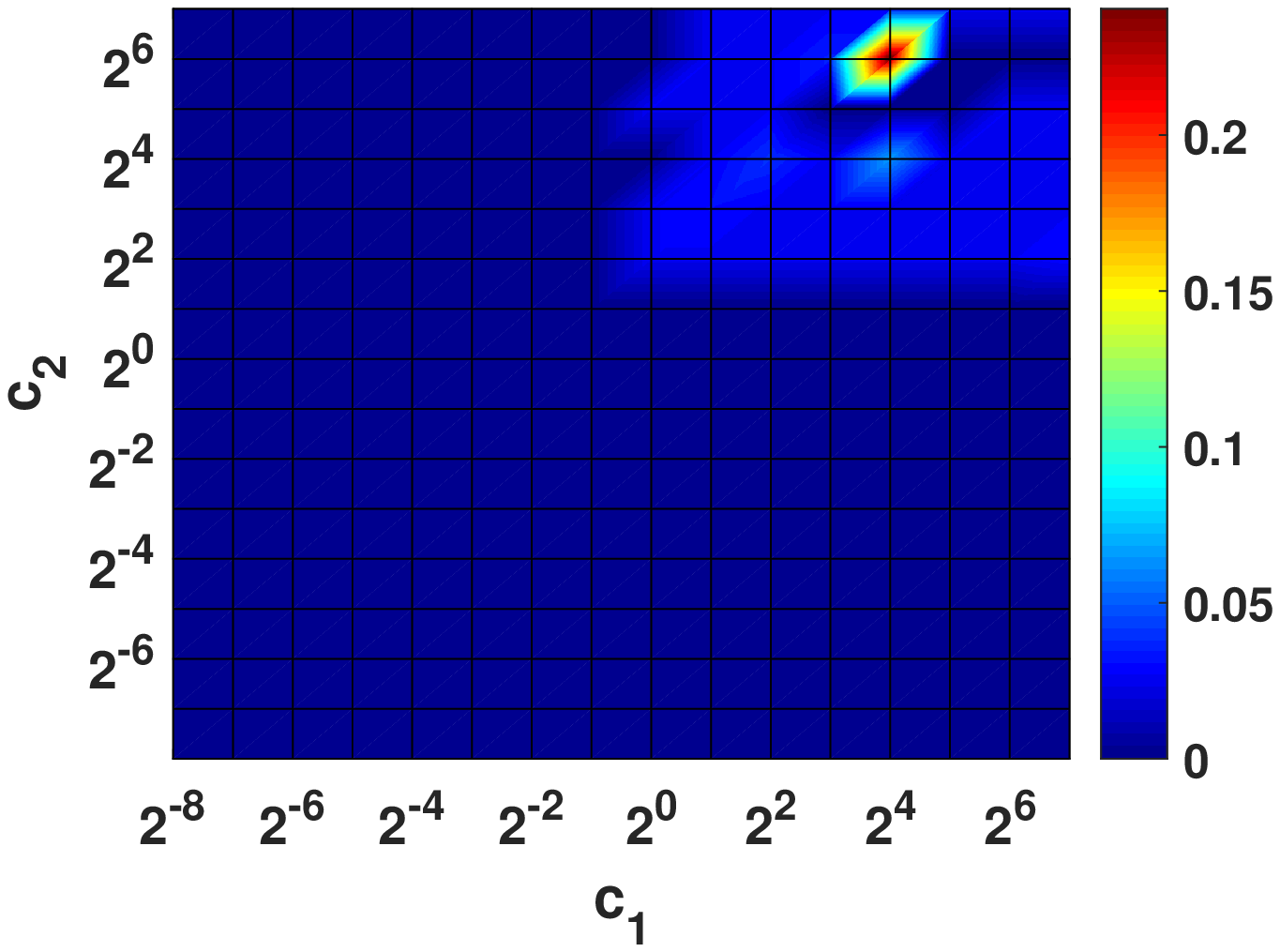}}
\subfigure[Dna]{\includegraphics[width=0.18\textheight]{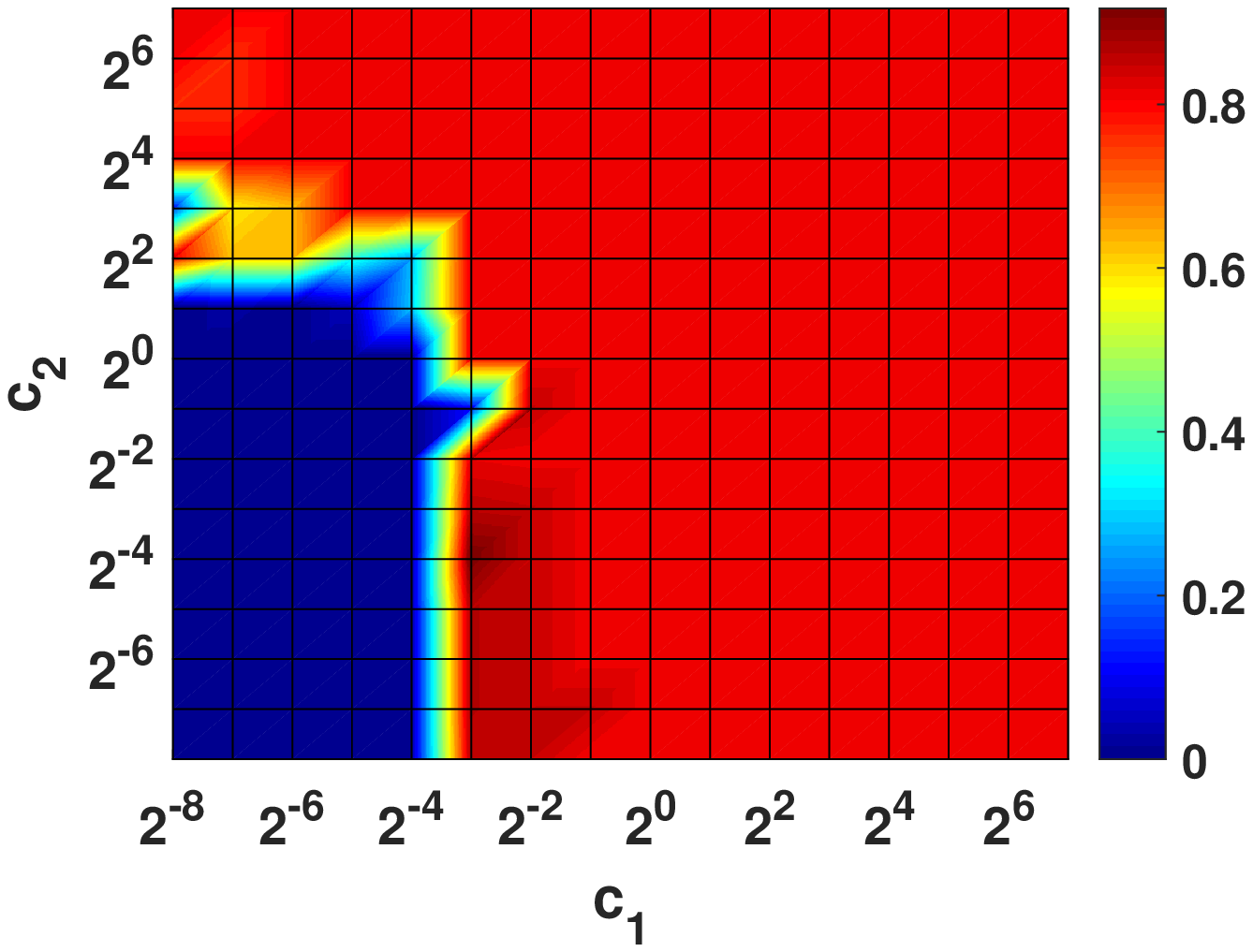}}
\subfigure[Echocardiogram]{\includegraphics[width=0.18\textheight]{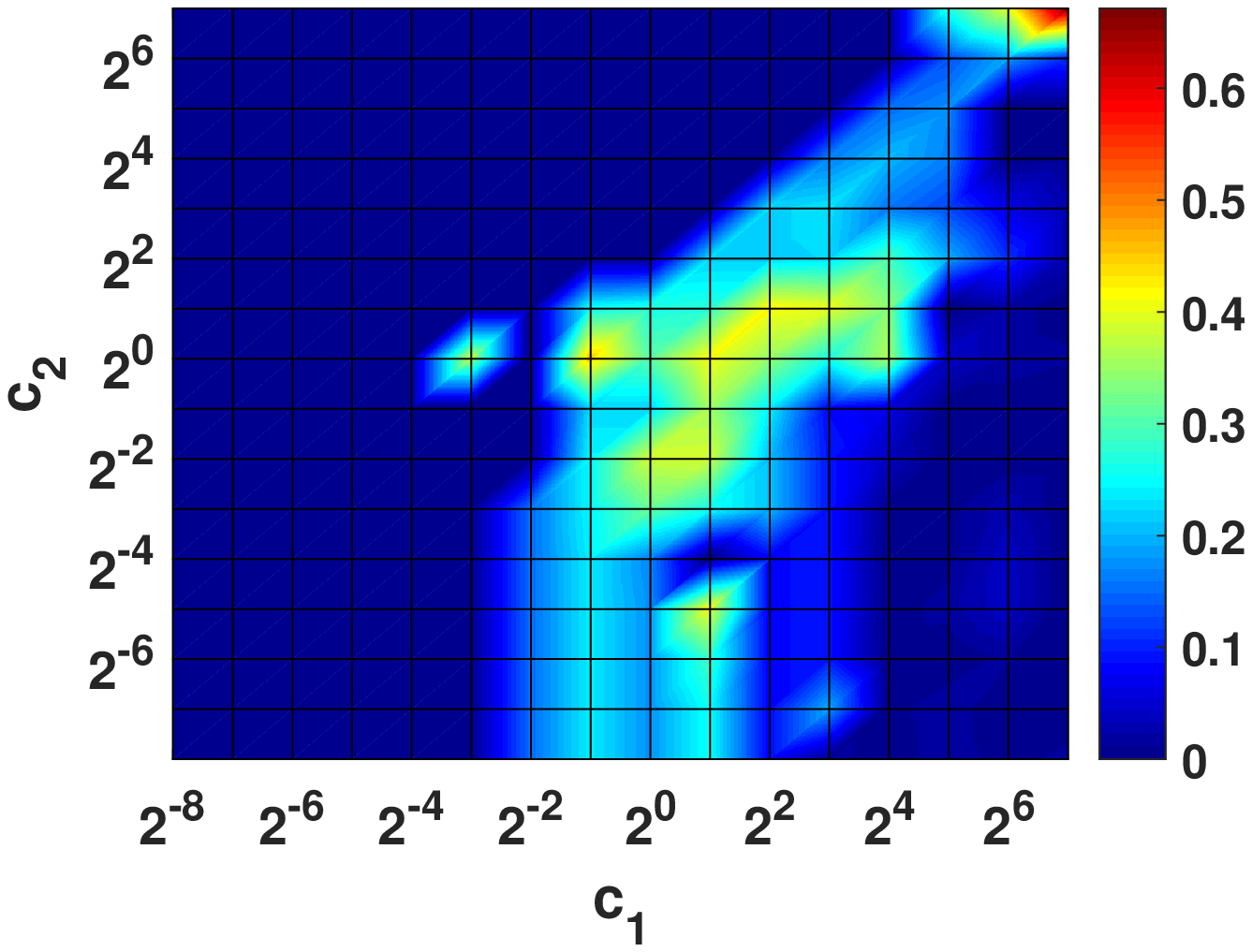}}
\subfigure[Ecoli]{\includegraphics[width=0.18\textheight]{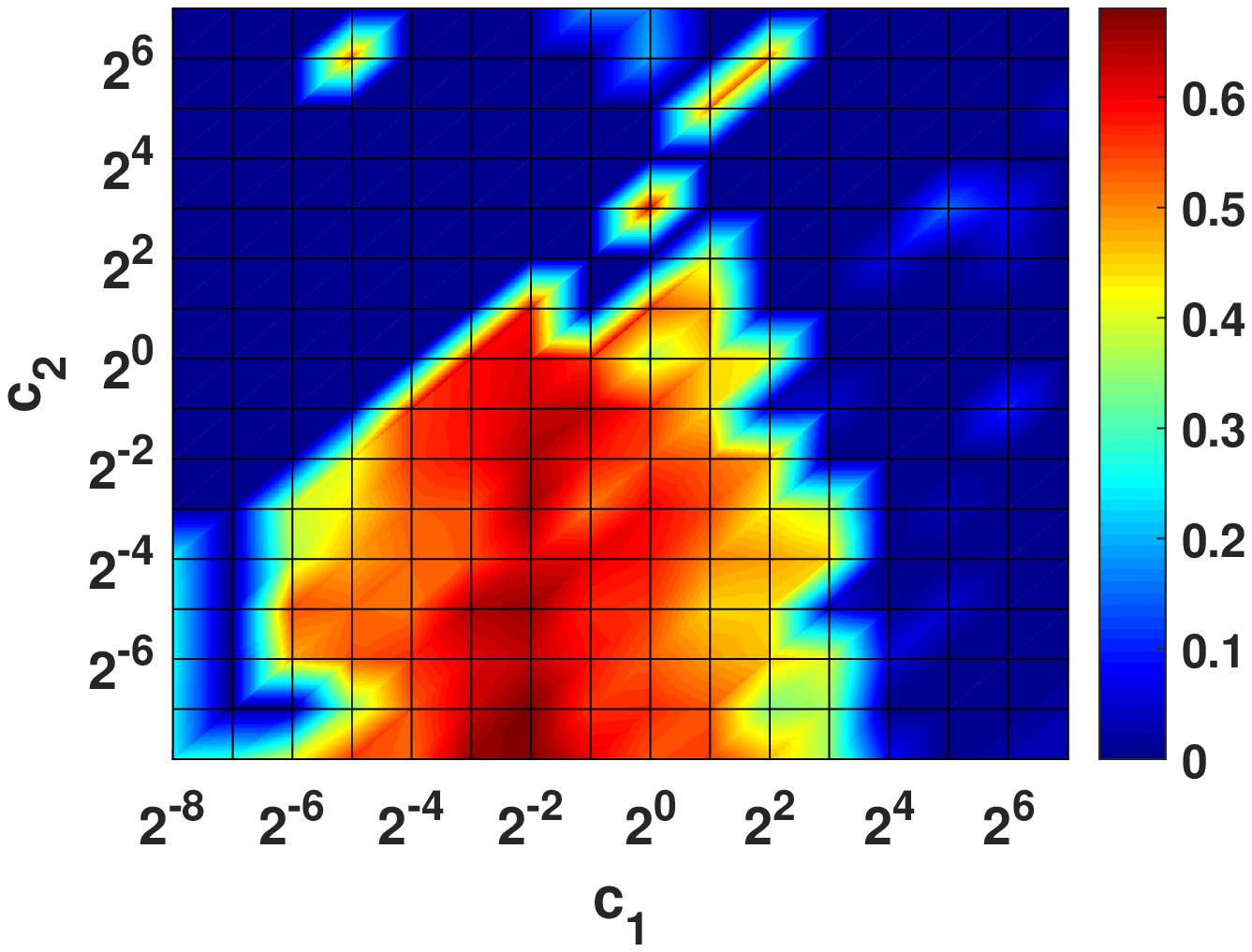}}
\subfigure[Housevotes]{\includegraphics[width=0.18\textheight]{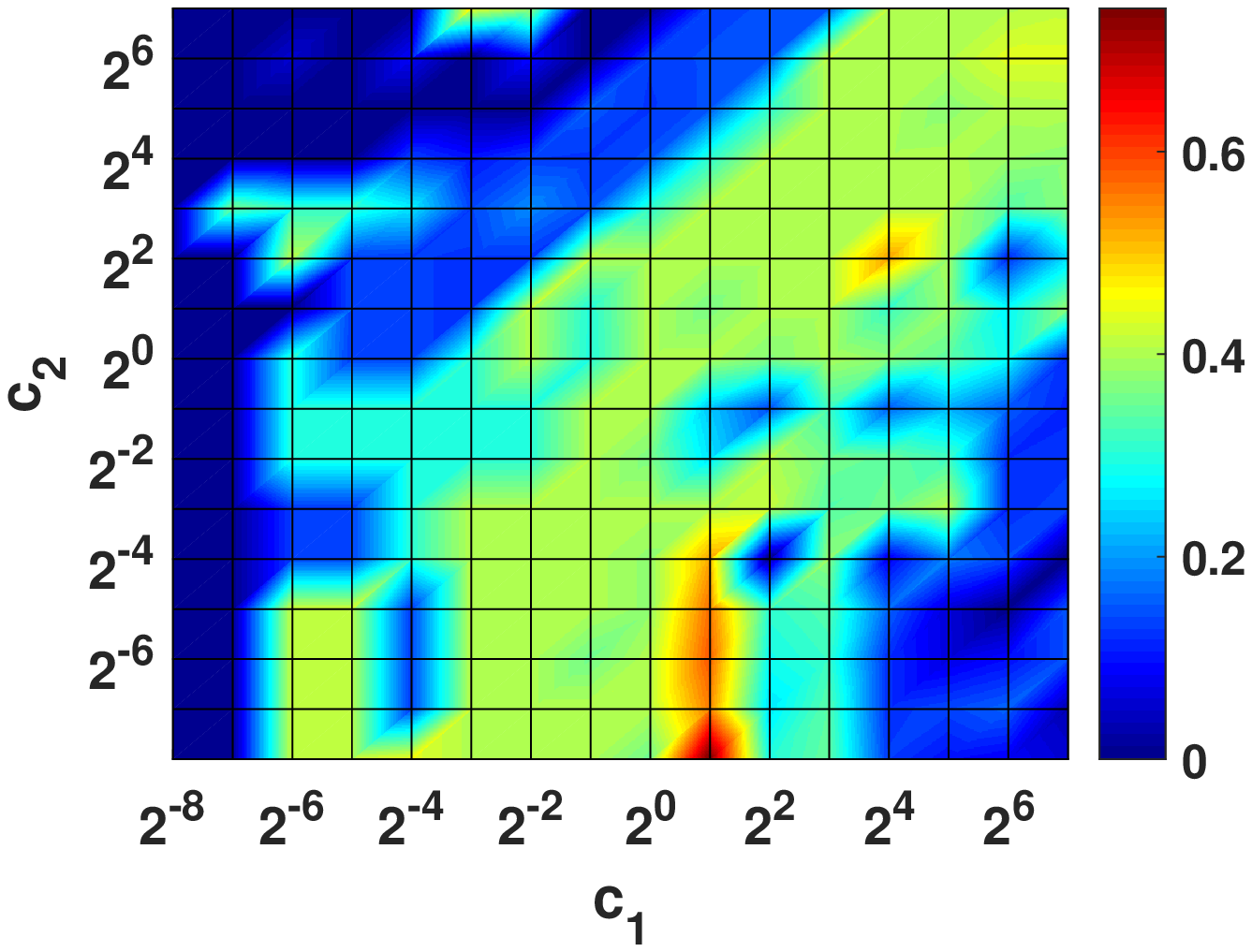}}
\subfigure[Iris]{\includegraphics[width=0.18\textheight]{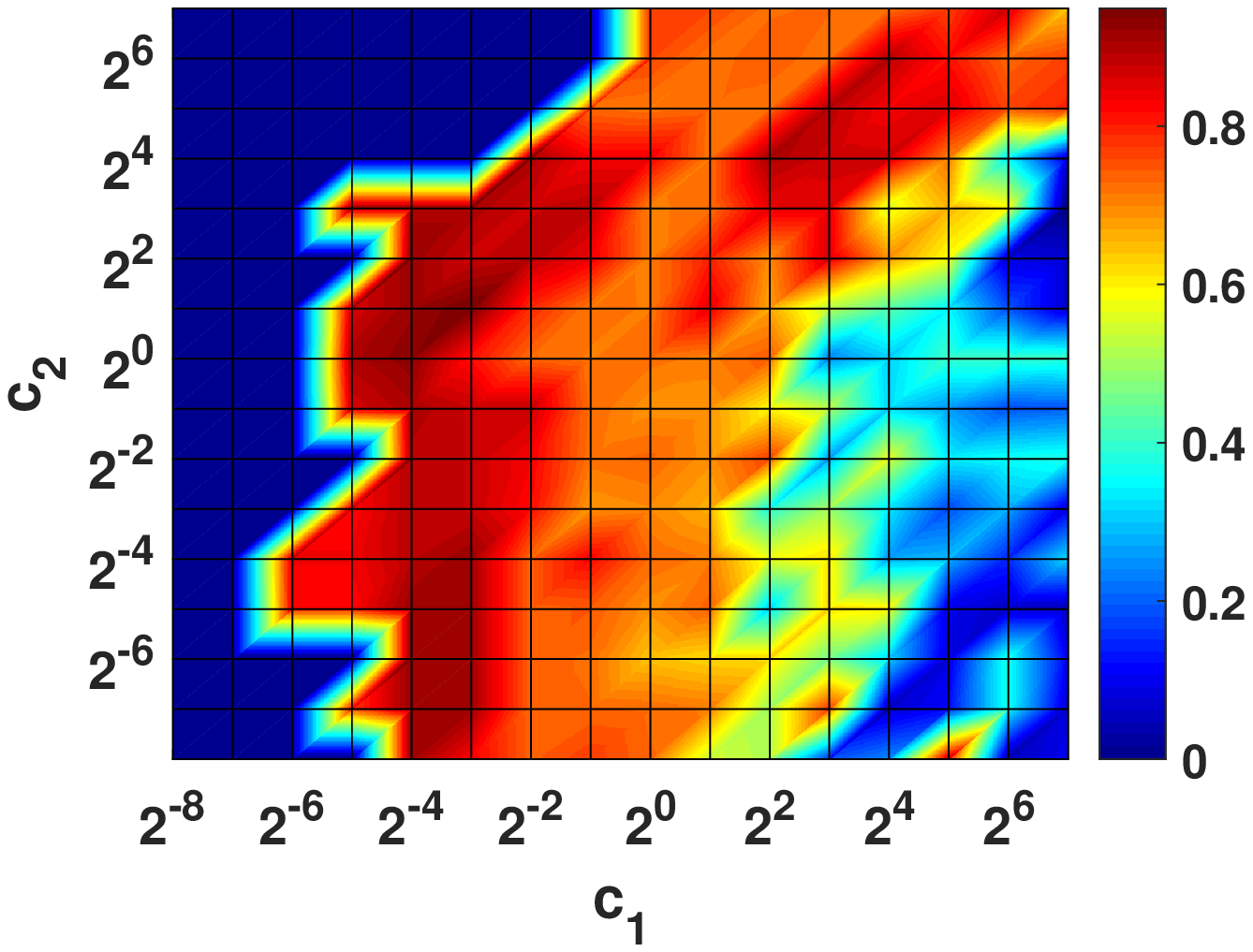}}
\subfigure[Seeds]{\includegraphics[width=0.18\textheight]{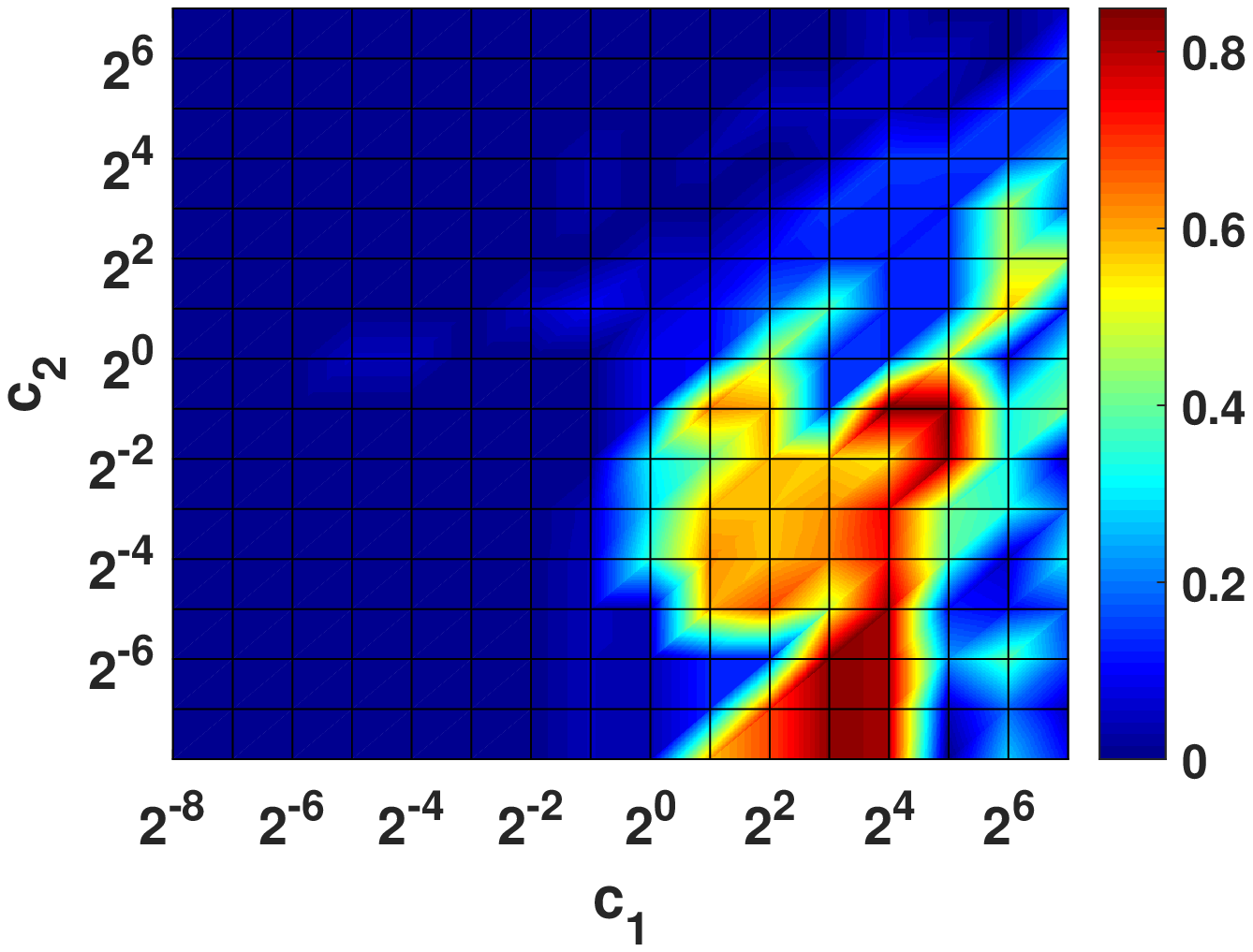}}
\subfigure[Wine]{\includegraphics[width=0.18\textheight]{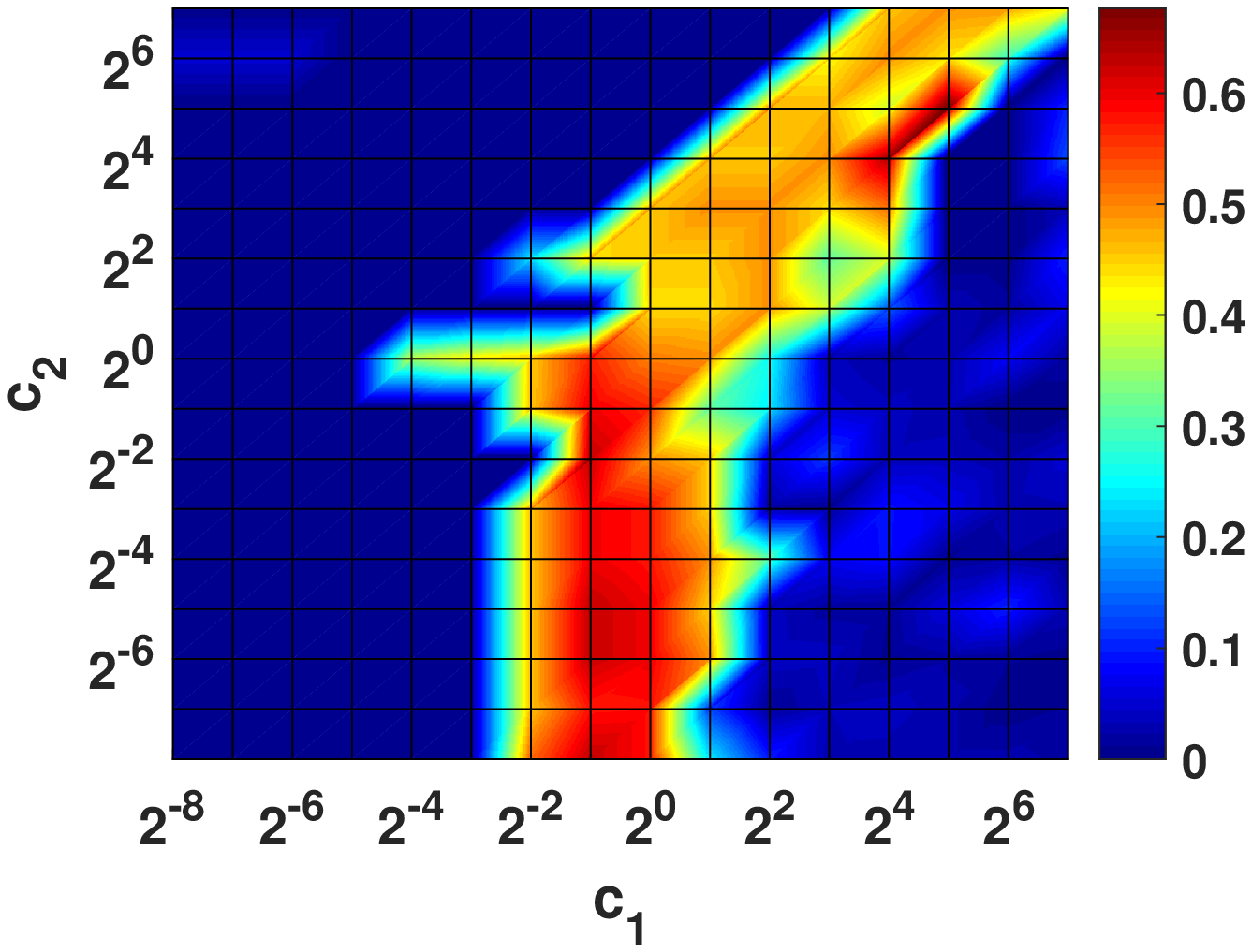}}
\caption{Influence of the trade-off parameters of MFPC with linear formation on some benchmark datasets, where the performance of each pair of $(c_1,c_2)$ is measured by NMI and denoted by color.}\label{FigLinearPara}
\end{figure*}

\begin{figure*}[htbp]
\centering
\subfigure[Australian]{\includegraphics[width=0.18\textheight]{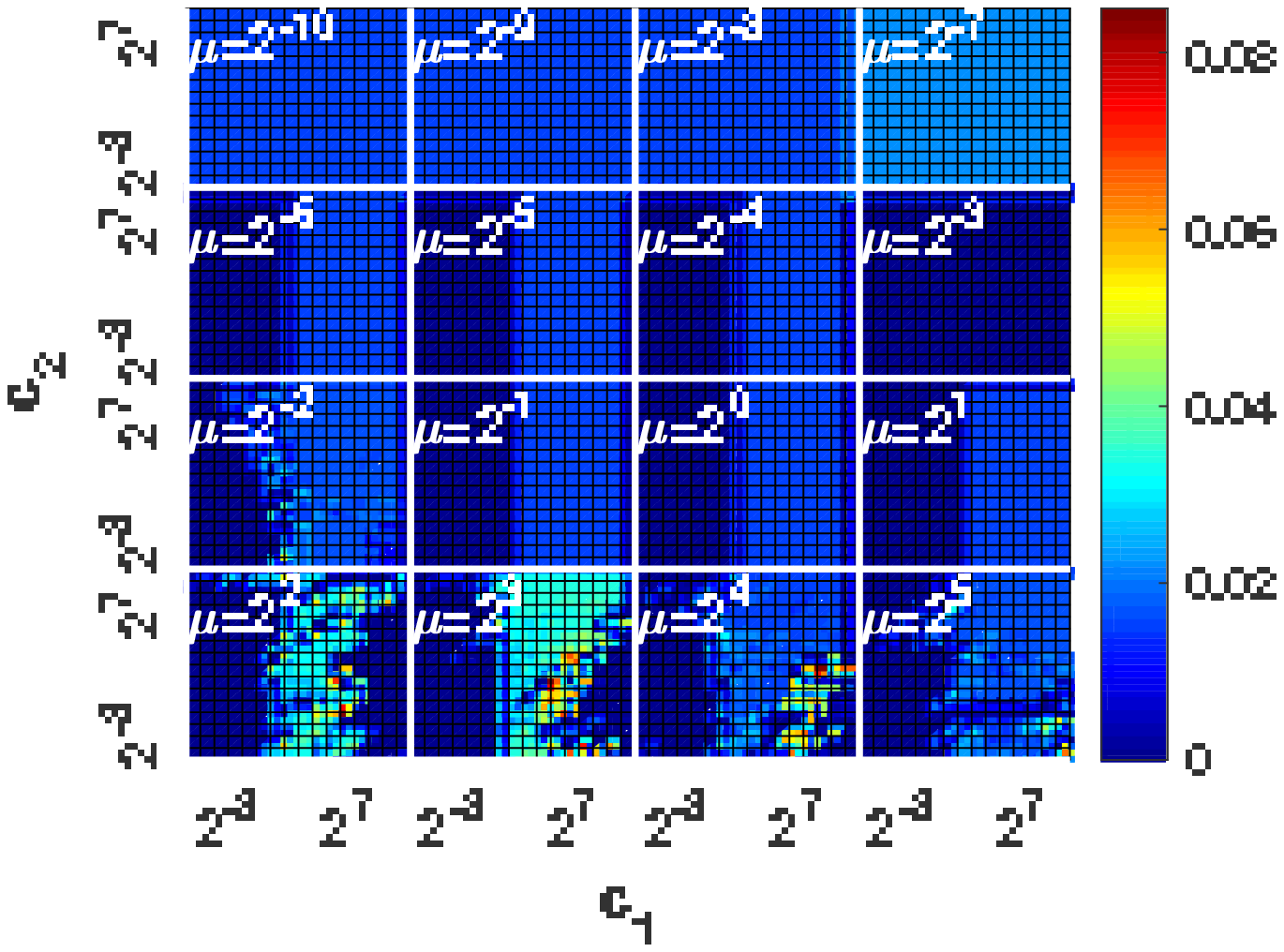}}
\subfigure[Dna]{\includegraphics[width=0.18\textheight]{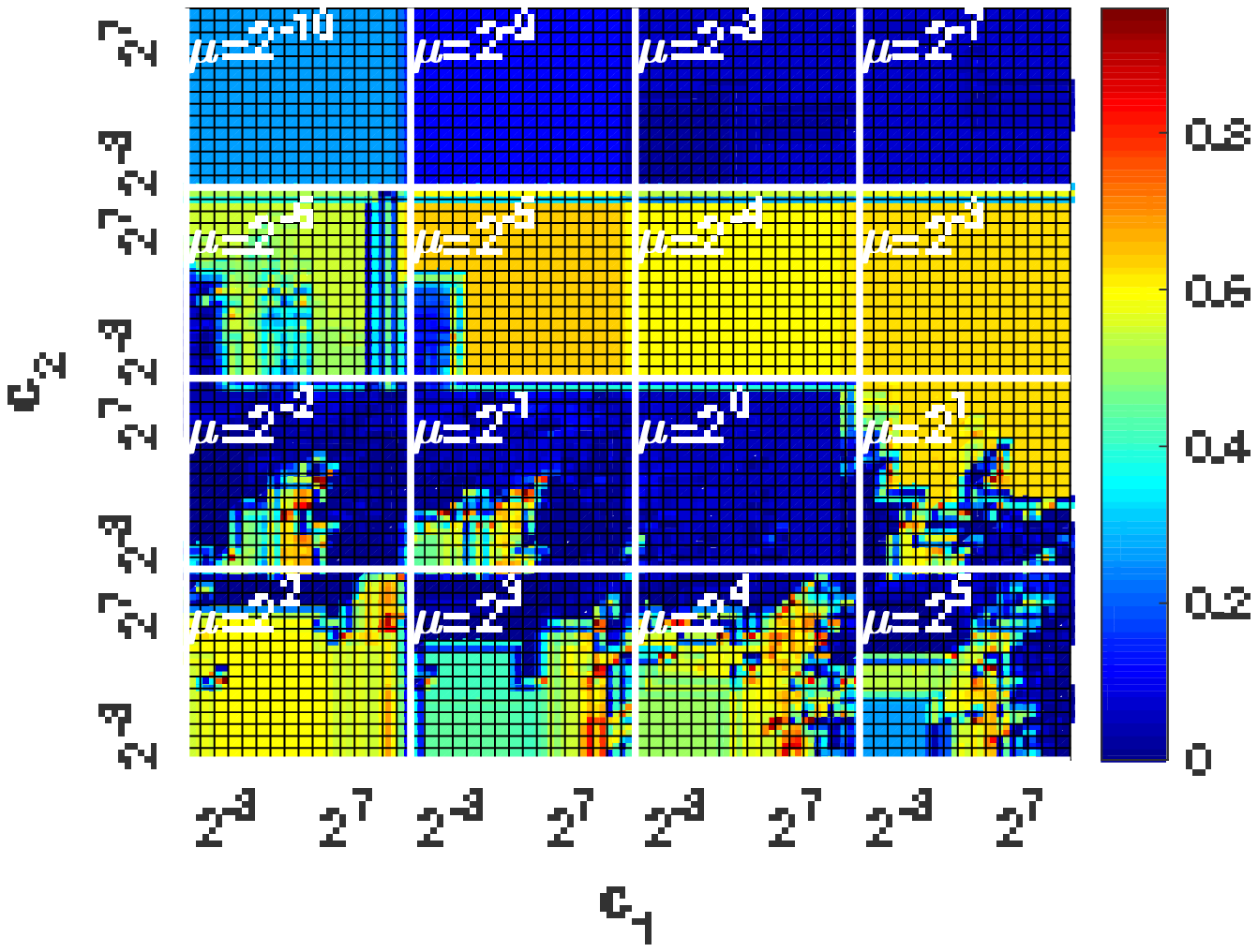}}
\subfigure[Echocardiogram]{\includegraphics[width=0.18\textheight]{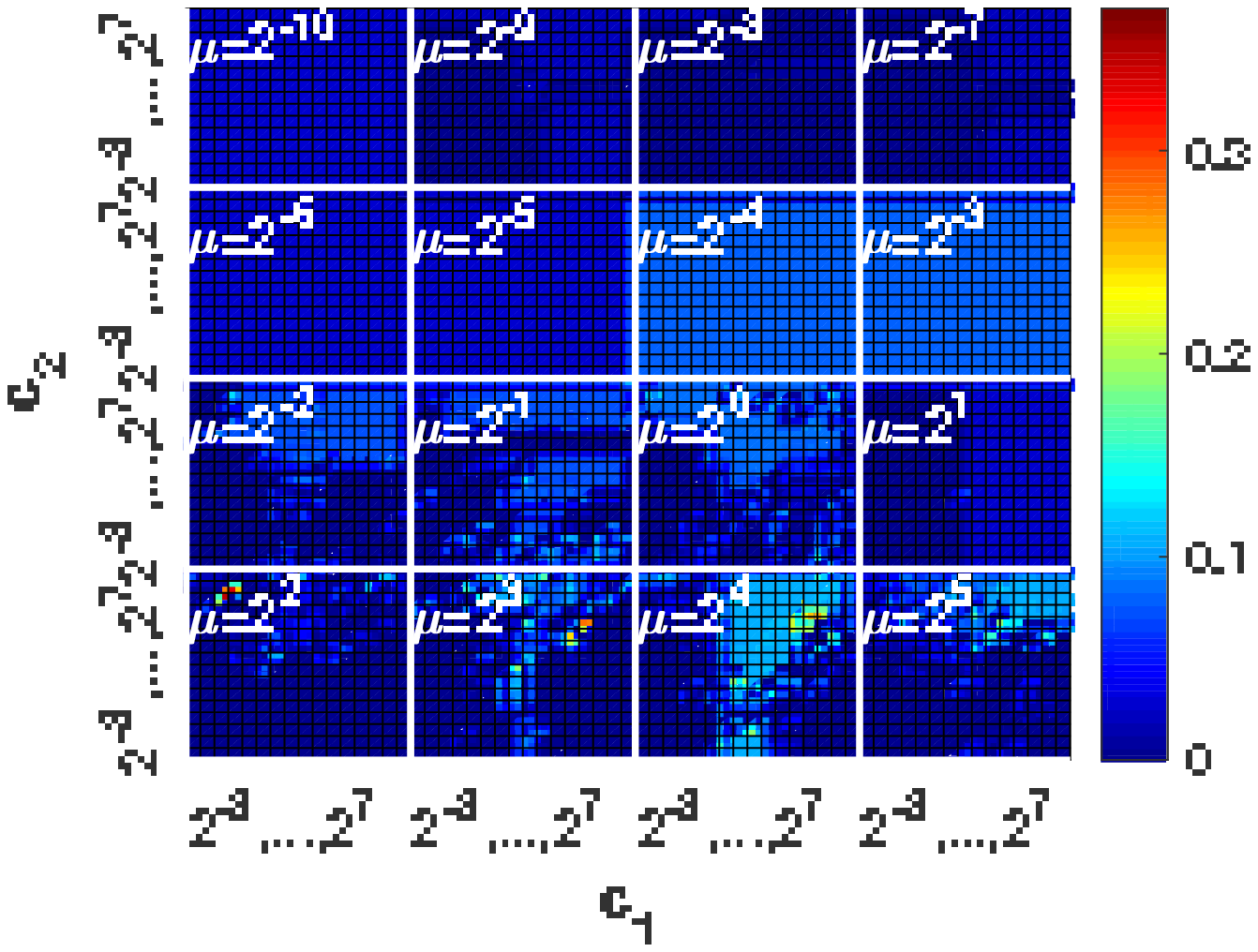}}
\subfigure[Ecoli]{\includegraphics[width=0.18\textheight]{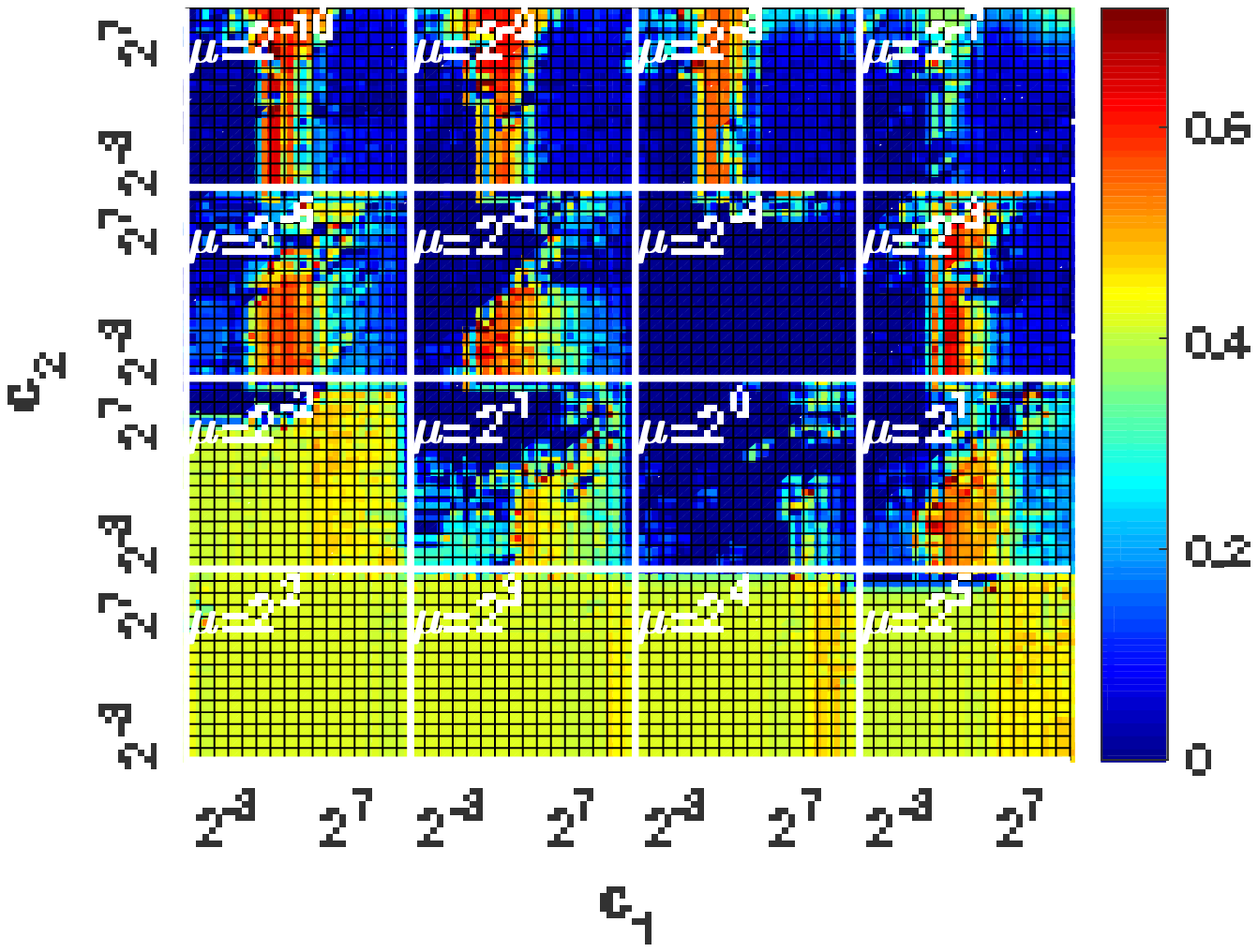}}
\subfigure[Housevotes]{\includegraphics[width=0.18\textheight]{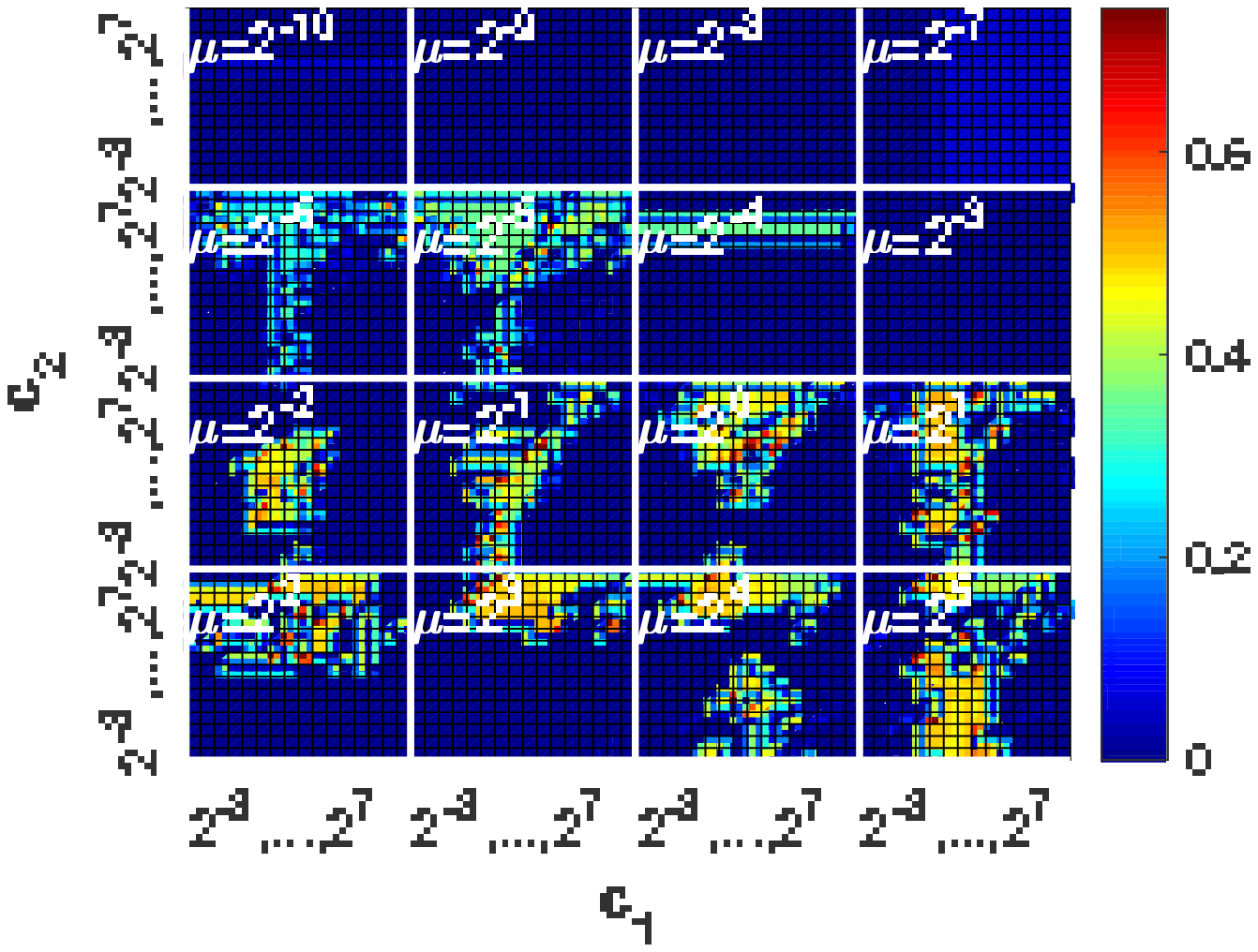}}
\subfigure[Iris]{\includegraphics[width=0.18\textheight]{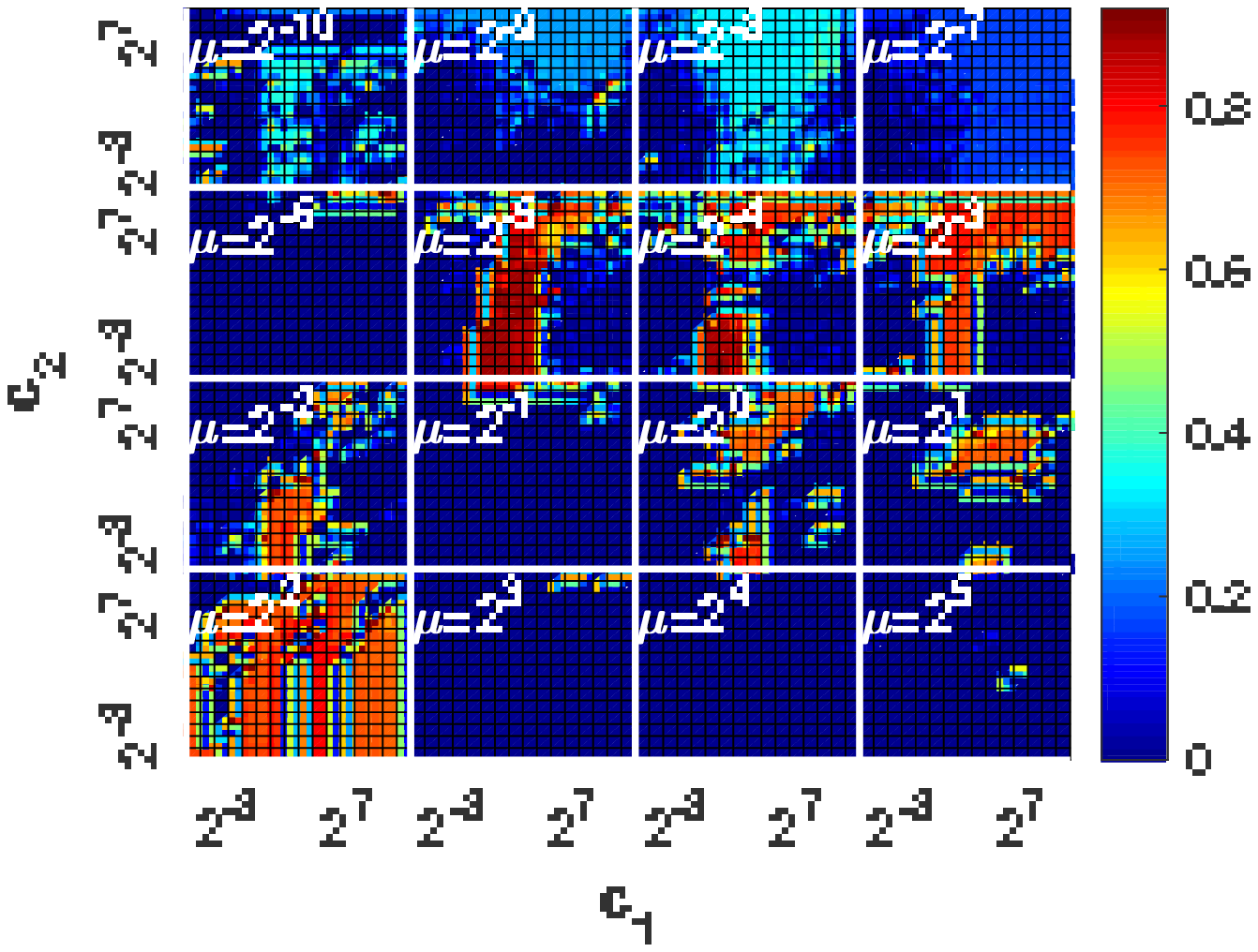}}
\subfigure[Seeds]{\includegraphics[width=0.18\textheight]{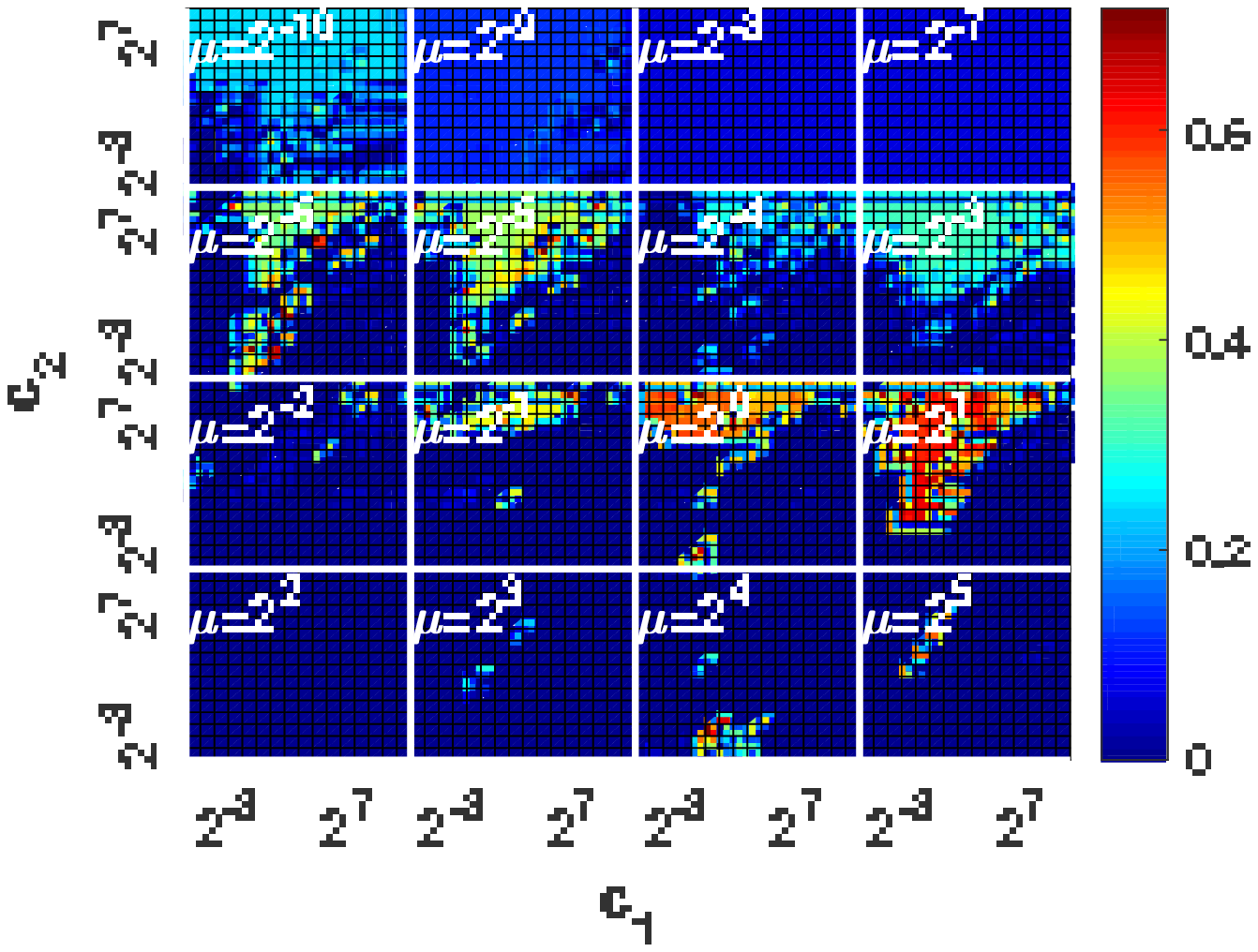}}
\subfigure[Wine]{\includegraphics[width=0.18\textheight]{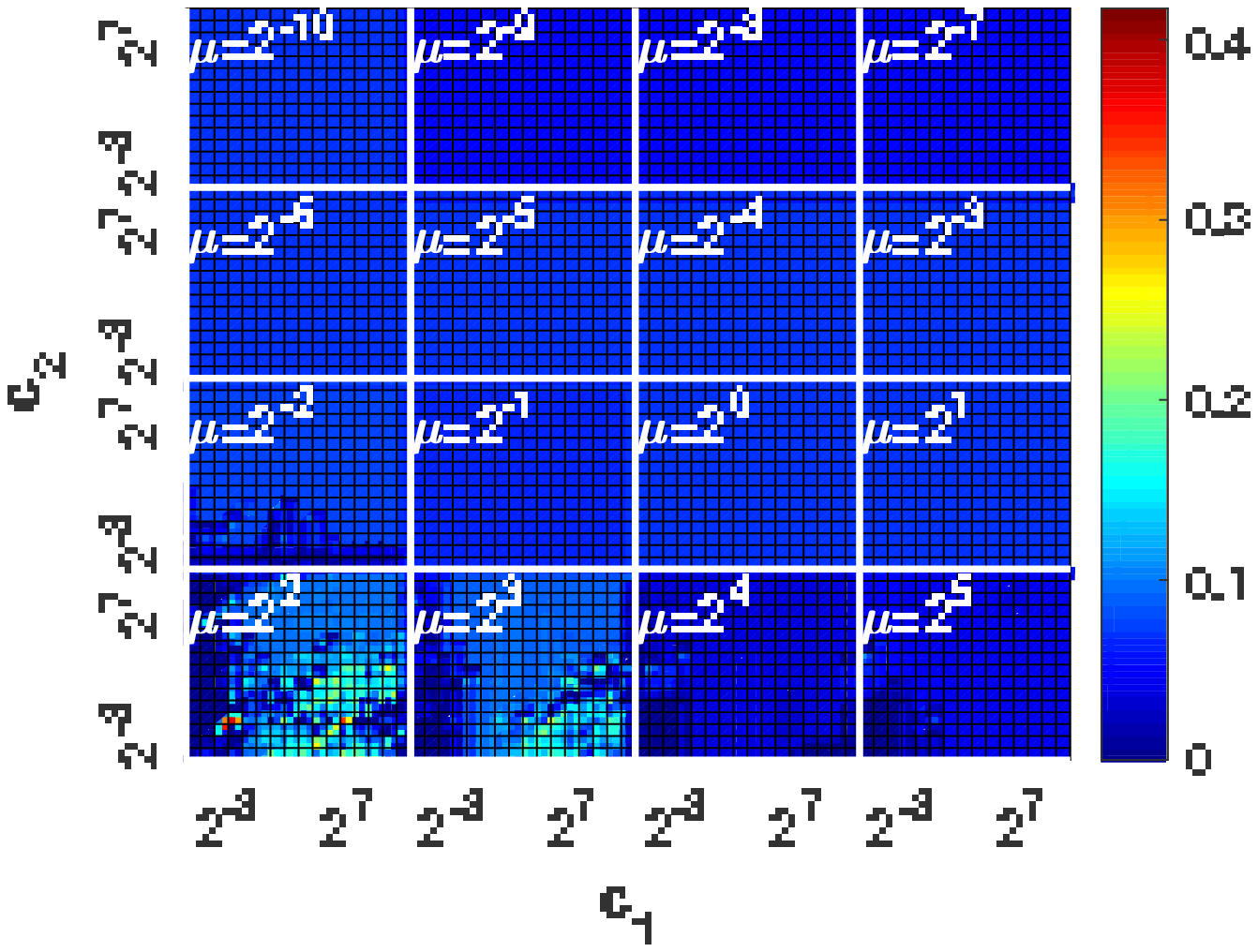}}
\caption{Influence of the trade-off parameters of MFPC with nonlinear formation on some benchmark datasets, where each figure includes $16$ subfigures corresponding to $16$ Gaussian parameters, and the performance of each pair of $(c_1,c_2)$ in the subfigures is measured by NMI and denoted by color. The four subfigures on the first row in each figure corresponds to $\mu\in\{2^{-10}, 2^{-9},2^{-8},2^{-7}\}$, and the next $12$ subfigures on the next three rows corresponds to $\mu\in\{2^{-6}, 2^{-5},\ldots,2^{5}$\}. }\label{FigNonlinearPara}
\end{figure*}

\subsection{Benchmark datasets}
The synthetic experiments have shown the effectiveness of our MFPC in manifold clustering. This subsection analyzed its performance on $17$ benchmark datasets \cite{UCI} compared with $k$means, SMMC and other flat-type methods. Thereinto, $k$means and SMMC were run $20$ times for their randomness, and the average measurements and standard deviations were recorded. The flat-type methods, including $k$PC, $k$PPC, L$k$PPC, TWSVC, $k$FC, L$k$FC and our MFPC, were run once with the NNG initialization, and their highest ARIs and NMIs on these datasets were recorded. All the results were reported in Tables \ref{LinearUCI} and \ref{NonlinearUCI} for linear and nonlinear formations, respectively. The highest ARI and NMI for each dataset were bold. From Tables \ref{LinearUCI} and \ref{NonlinearUCI}, it is obvious that our MFPC performs much better than other methods on most of the datasets, and it is comparable with the best one on the rest datasets. Additionally, some other phenomena are noticeable in these tables. First of all, ARI is consistent with NMI generally, i.e., a method obtains a higher ARI than another method often with a higher NMI concurrently, and vice versa, though ARI is based on label partition statistics and NMI is based on information theory. For simplicity, NMI is always hired in the following experiments. Secondly, we found that $k$means and SMMC were stable on some datasets, e.g., $k$means on ``Spect'' in Table \ref{LinearUCI} and SMMC on ``Hepatitis'' in Table \ref{NonlinearUCI} with standard deviation zeros. These two methods often provides different results with different initializations in theory. Thus, it is almost certain that $k$means and SMMC do their best to work on the datasets if they obtain deviation zeros in $20$ repeated tests. In contrast, the flat-type methods were implemented by the NNG initialization to perform stably. In this situation, a flat-type method would be always better than $k$means or SMMC on a dataset if its ARI/NMI is higher than the latter's average plus standard deviation. Furthermore, we cannot conclude that a flat-type method would be worse than $k$means or SMMC on a dataset if its measurement is lower than the latter's. Compared with Tables \ref{LinearUCI} and \ref{NonlinearUCI}, the performance of many methods was promoted by the kernel tricks, and the representative results were on ``Pathbased'' dataset. No method is more accurate than $60\%$ on this dataset in Table \ref{LinearUCI}, while many methods are more accurate than $90\%$ in Table \ref{NonlinearUCI}. Of course, these methods with nonlinear formations are sometimes worse than their linear formations, e.g., on ``Australian'' dataset. Hence, the kernel tricks can promote these methods, but an improper kernel may reduce their performance. Last but not least, for the methods we compared, there are a little datasets on which some of them outperform other methods, e.g., the flat-type methods on ``Soybean''. This indicates different type methods have their different applicable scopes, e.g., $k$means for point-based cluster centers and flat-type methods for plane-based cluster centers. However, our MFPC suits for many different cases in Tables \ref{LinearUCI} and \ref{NonlinearUCI} obviously, which implies that our MFPC has a larger applicable scope than other methods. If there is not any prior information, MFPC may be an admirable choice.

To evaluate the performance of the nine methods on the $17$ datasets, we ranked them with following strategy: for each dataset, the methods were ordered by the measurement, where the highest one received the ranking $1$ and the lowest one received the ranking $9$. The average rankings were reported at the last rows in Tables \ref{LinearUCI} and \ref{NonlinearUCI}. Among these methods, the original flat-type $k$FC is better that the plane-based $k$PC, because $k$FC can degenerate to $k$PC. After some improvements, L$k$PPC based on $k$PC exceeds $k$FC and L$k$FC. Obviously, our MFPC is on the first place among these methods with both linear and nonlinear formations.

In Fig. \ref{FigLinearPara}, we further reported the NMIs for each pair of parameters in our linear MFPC on eight benchmark datasets to show the influence of the parameters, where higher NMI corresponds to warmer color. Apparently, the subfigures in Fig. \ref{FigLinearPara} are different from each other. For instance, MFPC reach the only peak in Fig. \ref{FigLinearPara}(a), while there are many peaks at various pairs of $(c_1,c_2)$ in Fig. \ref{FigLinearPara}(f). Generally, the trade-off parameters $c_1$ and $c_2$ played the important roles in MFPC on these datasets, but ``Dna'' and ``Iris'' are two exceptions. On ``Dna'', MFPC is insensitive with $c_2$, i.e., MFPC can obtain a desirable result with an appropriate $c_1$ for any $c_2$. The same thing appears on ``Iris''. However, on the other six datasets, one should carefully select the parameters to achieve the best performance. Fig. \ref{FigNonlinearPara} illustrated the influence of the parameters in nonlinear MFPC. Each subfigure in Fig. \ref{FigNonlinearPara} were split into $16$ parts corresponding to $16$ Gaussian kernel parameters. Normally, the samples are mapped into various high dimensional feature spaces with different kernel parameters. Thus, the manifolds represented by the samples are transformed too. It can be seen that our MFPC often works well on a certain feature spaces on most of datasets. Compared with the parameters $c_1$ and $c_2$, the kernel parameter $\mu$ has significant effect on MFPC. Thus, an appropriate feature space, which actually improve the performance of nonlinear MFPC, has the precedence in parameter selection.

\begin{figure}[htbp]
\includegraphics[width=0.35\textheight]{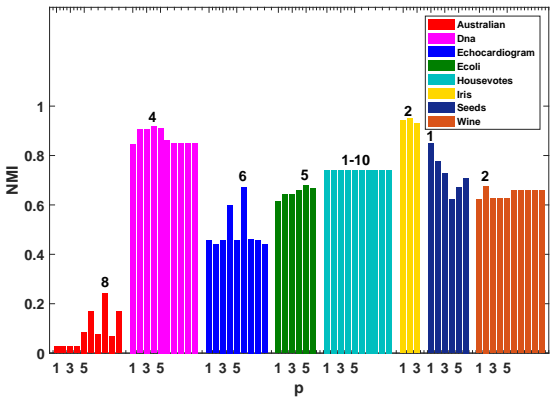}
\caption{Influence of flat dimension of MFPC on some benchmark datasets, where the number above the bar relates to the highest NMI among these bars for each dataset.}\label{FigDimension}
\end{figure}

Finally, we analyzed the influence of the flat dimension in our MFPC, where the flat dimension is controlled by parameter $p$. We ran MFPC on eight datasets with $p\in \{1,2,\ldots, \min(n-1,10)\}$, and the highest NMIs corresponding to different $p$ were reported in Fig. \ref{FigDimension}. It is clear that MFPC performs differently with different flat dimension generally. For each dataset, the number above the bar related to the highest NMI among these bars. The highest bar indicates the appropriate dimension of manifolds in the datasets. For instance, MFPC has the highest NMI with $p=6$ on ``Echocardiogram'', and thus we shall infer that there are some implicit manifolds with the dimension $n-p=4$. If MFPC obtains the same results with different $p$, e.g., on ``Housevotes'', there would be some implicit manifolds with much lower dimension due to flat with high dimension can degenerate to flat with low dimension. It should be pointed out that our MFPC regards the implicit manifolds as the flats with the same dimension. Therefore, a more reasonable way to capture the implicit manifolds is to hire flats with various dimensions, which we will consider in the future work.

\section{Conclusion}
A multiple flat projections clustering method (MFPC) for cross-manifold clustering has been proposed. It projects the given samples into multiple subspaces to discover the implicit manifolds. In MFPC, the samples on the same manifold would be distinguished from the others, though they may be separated by the cross structures. The non-convex matrix optimization problems in MFPC are decomposed into several non-convex vector optimization problems recursively, which are solved by a convergent iterative algorithm. Moreover, MFPC has been extended to nonlinear case via kernel tricks, and this nonlinear model can handle more complex cross-manifold clustering. The synthetic tests have shown that our MFPC has the ability to discover the implicit manifolds from cross-manifold data. Further, experimental results on the benchmark datasets have indicated
that our MFPC outperforms many other state-of-the-art clustering methods. For practical convenience, the synthetic datasets and the corresponding MFPC codes have been uploaded upon http://www.optimal-group.org/Resources/Code/MFPC.html. It is true that the computation cost of our MFPC is higher than other methods. Consequently, designing more efficient solvers and model selection methods are the future works.




\section{appendices}
\subsection{The proof of Theorem \uppercase\expandafter{\romannumeral3}.1}

\begin{proof}
Assume there is no relaxation term in the first restriction condition in \eqref{MSSVCmain}, and consider the following simple form
\begin{eqnarray}\label{MSSVCsample}
\begin{array}{l}
\underset{W_{i}}{\min}~~||W_{i}||_F^2\\
\hbox{s.t.}
~~~||W_{i}^{\top}(x_j-\bar{x}_i)||\geq 1,~j\in N\backslash{N}_i.
\end{array}
\end{eqnarray}
Suppose there exits the solution $W_i^*$ to problem \eqref{MSSVCsample}. The distance between the center $\bar{x}_i$ and every sample $x_j~(j\in N\backslash{N}_i)$ from other cluster in the $i$-th projection subspace can be expressed as
\begin{eqnarray}\label{MSSVCd_k}
\begin{array}{l}
d_j=||\sqrt{(W_i^{*\top}W_i^*)^{-1}}W_i^{*\top}(x_j-\bar{x}_i)||,
\end{array}
\end{eqnarray}
where the square root of a matrix is such a matrix whose elements are the square roots of the elements from the previous matrix.
Then, the distance between the center of the $i$-th cluster and the closest point in other clusters in the projection subspace can be expressed as
\begin{eqnarray}\label{MSSVCd_D}
\begin{array}{l}
d_{\min}=\underset{x_j}\min||\sqrt{(W_i^{*\top}W_i^*)^{-1}}W_i^{*\top}(x_j-\bar{x}_i)||\\
~~~\geq\min(\frac{1}{||w_{i,1}^*||},\frac{1}{||w_{i,2}^*||},\ldots,\frac{1}{||w_{i,p}^*||})||W_i^{*\top}(x_j-\bar{x}_i)||\\
~~~\geq\min(\frac{1}{||w_{i,1}^*||},\frac{1}{||w_{i,2}^*||},\ldots,\frac{1}{||w_{i,p}^*||}).
\end{array}
\end{eqnarray}
Therefore, maximizing $\min(\frac{1}{||w_{i,1}^*||},\frac{1}{||w_{i,2}^*||},\ldots,\frac{1}{||w_{i,p}^*||})$, which is equal to minimize $\min(||w_{i,1}^*||,||w_{i,2}^*||,\ldots,||w_{i,p}^*||)$, will result in maximizing $d_{\min}$. Note that minimizing $||W_i||_F^2$ in \eqref{MSSVCmain} includes minimizing $\min(||w_{i,1}^*||,||w_{i,2}^*||,\ldots,||w_{i,p}^*||)$, and thus the conclusion holds.
\end{proof}

\subsection{The proof of Theorem \uppercase\expandafter{\romannumeral3}.3}
\begin{proof}
For the $l$-th iteration, note that $\tilde{w}_{i,l}=w_{i,l}/||w_{i,l}||$. Thus, we have
\begin{equation}\label{1}
\begin{array}{l}
w_{i,l}^{\top} x_{j,{l+1}}=w_{i,l}^{\top} x_{j,l}-w_{i,l}^{\top}(\tilde{w}_{i,l}\tilde{w}_{i,l}^{\top})x_{j,l}=0
\end{array}
\end{equation}
i.e., $w_{i,l}$ is orthogonal with the projected samples $x_{j,l+1}$ (for all $j\in N$).
On the other hand, the regularization term in problem \eqref{MSSVCW} is obviously a strictly monotonical increasing real-value function on $[0,\infty)$. From the representer theorem \cite{Bernhard2001A}, $w_{i,l+1}$ obtained by \eqref{MSSVCW} is represented linearly by the projected samples $x_{j,l+1}$ (for all $j\in N$). Thus, $w_{i,l}$ is orthogonal with $w_{i,l+1}$.

Moreover, $w_{i,l}$ is orthogonal with $x_{j,l+2}$ (for all $j\in N$) because $x_{j,l+2}$ (for all $j\in N$) is generated linearly by $w_{i,l+1}$ and $x_{j,l+1}$. By the representer theorem again, we can get that $w_{i,l}$, $w_{i,l+1}$ and $w_{i,l+2}$ are orthogonal to each other. The above orthogonality can be established sequentially from $l=1$ to $l=p$.
\end{proof}

\section*{Acknowledgment}
This work is supported in part by National Natural Science Foundation of
China (Nos. 61966024, 11926349, 61866010 and 11871183), in part by Program for Young Talents of Science and Technology in Universities of Inner Mongolia Autonomous Region (No. NJYT-19-B01), in part by Natural Science Foundation of Inner Mongolia Autonomous Region (Nos. 2019BS01009, 2019MS06008), in part by Scientific Research Foundation of Hainan University (No. kyqd(sk)1804).

\ifCLASSOPTIONcaptionsoff
  \newpage
\fi



%

%
\bibliographystyle{IEEEtran}
\bibliography{FBib}
\clearpage






\end{document}